\documentclass[journal,onecolumn]{IEEEtran}

\usepackage{color}
\usepackage{hyperref}
\usepackage{cite}

\ifCLASSINFOpdf
   \usepackage[pdftex]{graphicx}
\else
  \usepackage[dvips]{graphicx}
\fi

\usepackage{amsmath}
\usepackage{amsfonts}

\usepackage{algorithm,algorithmicx}
\usepackage{algpseudocode}

\usepackage{array}
\usepackage{multirow}
\newcolumntype{C}{>{\centering\arraybackslash}p{5em}}
\newcolumntype{d}{>{\centering\arraybackslash}p{3em}}

\ifCLASSOPTIONcompsoc
  \usepackage[caption=false,font=normalsize,labelfont=sf,textfont=sf]{subfig}
\else
  \usepackage[caption=false,font=footnotesize]{subfig}
\fi

\usepackage{dblfloatfix}

\begin{document}
%
\title{Unsupervised Classification in Hyperspectral Imagery with Nonlocal Total Variation  and Primal-Dual Hybrid Gradient Algorithm}
%
%
%

\author{Wei Zhu, Victoria Chayes, Alexandre Tiard, Stephanie Sanchez, Devin Dahlberg,\\ 
	Andrea L. Bertozzi,  Stanley Osher, Dominique Zosso, and Da Kuang%
\thanks{This work was supported by NSF grants DMS-1118971, DMS-1045536,  DMS-1417674, and ONR grant N00014-16-1-2119}%
\thanks{Wei Zhu, Da Kuang, Andrea L. Bertozzi, Stanley Osher, Dominique Zosso, and Stephanie Sanchez are with the Department of Mathematics at University of California, Los Angeles. Email:\{weizhu731, dakuang, bertozzi, sjo, zosso\}@math.ucla.edu, stephanie2000@g.ucla.edu}%
\thanks{Victoria Chayes is with Bard College. Email: vminervachayes@gmail.com}%
\thanks{Alexandre Tiard is with ENSE3, Grenoble Institute of Technology. Email: alexandretiard@gmail.com}%
\thanks{Devin Dahlberg is with University of California, San Diego. Email: dahlbergdevin@gmail.com}}

\maketitle

\begin{abstract}
In this paper, a graph-based nonlocal total variation method (NLTV) is proposed for unsupervised classification of hyperspectral images (HSI). The variational problem is solved by the primal-dual hybrid gradient (PDHG) algorithm. By squaring the labeling function and using a stable simplex clustering routine, an unsupervised clustering method with random initialization can be implemented. The effectiveness of this proposed algorithm is illustrated on both synthetic and real-world HSI, and numerical results show that the proposed algorithm outperforms other standard unsupervised clustering methods such as spherical K-means, nonnegative matrix factorization (NMF), and the graph-based Merriman-Bence-Osher (MBO) scheme.
\end{abstract}


\section{Introduction}

\IEEEPARstart{H}{yperspectral} imagery (HSI) is an important domain in the field of remote sensing with numerous applications in agriculture, environmental science, mineralogy, and surveillance \cite{chang2003hyperspectral}. Hyperspectral sensors capture information of intensity of reflection at different wavelengths, from the infrared to ultraviolet. They take measurements 10-30nm apart, and up to 200 layers for a single image. Each pixel has a unique spectral signature, which can be used to differentiate objects that cannot be distinguished based on visible spectra, for example: invisible gas plumes, oil or chemical spills over water, or healthy from unhealthy crops.

The majority of HSI classification methods are either \textit{unmixing} methods or \textit{clustering} methods. Unmixing methods  extract the information of the constitutive materials (the \textit{endmembers}) and the abundance map\cite{unmixingoverview,unmixinggillis,unmixingjia,kuang}. Clustering methods do not extract endmembers; instead, they return the spectral signatures of the centroids of the clusters. Each centroid is the mean of the signatures of all the pixels in a cluster. However, when it is assumed that most of the pixels are dominated mostly by one endmember, i.e. in the absence of partial volume effects \cite{partialvolumeeffect}, which is  usually the case for high-resolution HSI, these two types of methods are expected to give similar results \cite{kuang}. The proposed nonlocal total variation (NLTV) method for HSI classification in this paper is a clustering method.

Much work has been carried out in the literature in both the unmixing and the clustering categories. HSI unmixing models can be characterized as linear or nonlinear. In a linear unmixing model (LUM), each pixel is approximated by a linear combination of the endmembers. When the linear coefficients are constrained to be nonnegative, it is equivalent to  nonnegative matrix factorization (NMF), and good unsupervised classification results have been achieved in \cite{unmixingjia, unmixinggillis, kuang} using either NMF or hierarchical rank-2 NMF (H2NMF). Despite the simplicity of LUM, the assumption of a linear mixture of materials has been shown to be physically inaccurate in certain situations\cite{nonlinearunmixing}. Researchers are starting to expand aggressively into the much more complicated nonlinear unmixing realm \cite{nonlinearunmixingoverview},  where nonlinear effects such as atmospheric scattering are explicitly modeled. However, most of the work that has been done for nonlinear unmixing so far is supervised in the sense that prior knowledge of the endmember signatures is required \cite{unmixingoverview}. Discriminative machine learning methods such as support vector machine (SVM) \cite{SVM2004,SVMkernel,SVM2008} and relevance vector machine (RVM) \cite{RVM,RVM2008,RVM2011} based approaches have also been applied to hyperspectral images, but they are also supervised methods since a training set is needed to learn the classifiers.

On the contrary, graph-based clustering methods implicitly model the nonlinear mixture of the endmembers. This type of method is built upon a weight matrix that encodes the similarity between the pixels, which is typically a sparse matrix constructed using the distances between the spectral signatures. Graph-cut problems for graph segmentation have been well-studied in the literature \cite{gc1,gc2,gc3,gc4}. In 2012, Bertozzi and Flenner proposed a diffuse interface model on graphs with applications to classification of high dimensional data\cite{GL}. This idea has been combined with the Merriman-Bence-Osher (MBO) scheme \cite{MBO} and applied to multi-class graph segmentation \cite{multiclass_graph1,multiclass_graph2} and HSI classification \cite{huiyi_plume,ekaterina_plume}. The method in \cite{GL} minimizes a graph version of the Ginzburg-Landau (GL) functional, which consists of the Dirichlet energy of the labeling function and a double-well potential, and uses Nystr\"{o}m extension to speed up the calculation of the eigenvectors for inverting the graph Laplacian. This graph-based method performed well compared to other algorithms in the detection of chemical plumes in hyperspectral video sequences\cite{huiyi_plume,ekaterina_plume}. However, the GL functional is non-convex due to its double-well term, which may cause the algorithm to get stuck in local minima. This issue can be circumvented by running the algorithm multiple times with different initial conditions and hand-picking the best result.

The two methods proposed in this paper are unsupervised graph-based clustering techniques. Instead of minimizing the GL functional, which has been proved to converge to the total variation (TV) semi-norm, this work proposes to minimize the NLTV semi-norm of the labeling functions $\|\nabla_w u_l\|_{L^1}$ directly. A detailed explanation of the nonlocal operator $\nabla_w$ and the labeling function $u_l$ will be provided in Section \ref{sec:background} and Section \ref{sec:model}. The $L^1$ regularized convex optimization problem is solved by the primal-dual hybrid gradient (PDHG) algorithm, which avoids the need to invert the graph Laplacian. We also introduce the novel idea of the quadratic model and a stable simplex clustering technique, which ensures that anomalies converge to their own clusters and makes random endmember initialization possible in the proposed algorithm. The direct usage of the NLTV semi-norm makes the proposed clustering methods more accurate than other methods when evaluated quantitatively on HSI with ground-truth labels, and the quadratic model with stable simplex clustering is a completely new addition to the field of HSI classification.


This paper is organized as follows: in Section \ref{sec:background} background is provided on total variation and nonlocal operators. Two NLTV models (linear and quadratic) and a stable simplex clustering method are presented in Section \ref{sec:model}. Section \ref{sec:pd} provides a detailed explanation on the application of the PDHG algorithm to solving the convex optimization problems in the linear and quadratic models. Section \ref{sec:results} presents the numerical results and a sensitivity analysis on the key model parameters. Section \ref{sec:conclusion} presents the conclusions. 

\section{Total Variation and Nonlocal Operators}
\label{sec:background}
Total variation (TV) method was introduced by Rudin et al in 1992 \cite{ROFpaper} and has been applied to various image processing tasks \cite{tvapplication}. Its advantage is that one can preserve the edges in the image when minimizing $\|\nabla u\|_{L^1}$ (TV semi-norm). The total variation model is:
\begin{equation*}
 \min_u E(u)=\|\nabla u \|_{L^1} + \lambda S(u).
\end{equation*}
The parameter $\lambda$ can be adjusted to give higher priority to the TV-regularizing term, or the data fidelity term $S(u)$.

Despite its huge success in image processing, the total variation method is still a local method. More specifically,  the gradient of a pixel is calculated using its immediate adjacent pixels. It is known that local image processing techniques fail to produce satisfactory results when the image has repetitive structures, or intrinsically related objects in the image are not spatially connected. To address this problem, Buades et al proposed a nonlocal means method based on patch distances for image denoising \cite{nlmean}. Gilboa and Osher \cite{oshernonlocal} later formalized a systematic framework for nonlocal image processing. Nonlocal image processing produces much better results because theoretically any pixel in the image can interact with any other, which better preserves texture and fine details.

In HSI classification, clusters can have elements that are not spatially connected. Thus it is necessary to develop a nonlocal method of gradient calculation. We provide a review of nonlocal operators in the rest of this section. Note that the model is continuous, and the weights are not necessarily symmetric \cite{retinex}.

Let $\Omega$ be a region in $\mathbb{R}^n$, and $u: \Omega \to \mathbb{R}$ be a real function. In the model for HSI classification, $\Omega$ is the  domain of the pixels, and $u : \Omega \to [0,1]$ is the labeling function of a cluster. The larger the value of $u(x)$, the more likely that pixel $x$ would be classified in that cluster.  The nonlocal derivative is: \begin{equation*} \frac{\partial u}{\partial y}(x):=\frac{u(y)-u(x)}{d(x,y)},\hspace{10pt} \text{for all } x, y \in \Omega, \end{equation*} where $d$ is a positive distance between $x$ and $y$.  In the context of hyperspectral images, $d(x,y)$ provides a way to measure the similarity between pixels $x$ and $y$. Smaller $d(x,y)$ implies more resemblance between these two pixels.  The nonlocal weight is defined as $w(x,y)=d^{-2}(x,y)$.

The nonlocal gradient $\nabla_{w} u$ for $u \in L^{2}(\Omega)$ can be defined as the collection of all partial derivatives, which is a function from $\Omega$ to $L^{2}(\Omega)$, i.e. $\nabla_w u \in L^2(\Omega, L^2(\Omega))$:
\begin{equation*} \nabla_w u(x)(y)=\frac{\partial u}{\partial y}(x) = \sqrt{w(x,y)}(u(y)-u(x)). \end{equation*} The standard $L^2$ inner products on Hilbert spaces $L^2(\Omega)$ and $ L^2(\Omega, L^2(\Omega))$ are used in the definition. More specifically, for $u_1, u_2 \in L^2(\Omega)$ and $v_1, v_2 \in L^2(\Omega,L^2(\Omega))$,
\begin{align*}
  \langle u_1, u_2\rangle&:=\int_\Omega u_1(x)u_2(x)dx,\\
  \langle v_1, v_2\rangle&:=\int_\Omega \int_\Omega v_1(x)(y)v_2(x)(y)dydx.
\end{align*}
The nonlocal divergence div$_w$ is defined as the negative adjoint of the nonlocal gradient:
\begin{equation*}
	\text{div}_wv(x):=\int_\Omega \sqrt{w(x,y)}v(x)(y)-\sqrt{w(y,x)}v(y)(x)dy.
\end{equation*}
At last, a standard $L^1$ and $L^\infty$ norm is defined on the space $L^2(\Omega,L^2(\Omega))$:
\begin{equation*}
	\| v \|_{L^1}:=\int_\Omega \| v(x) \|_{L^2} dx=\int_\Omega \left| \int_\Omega \left|v(x)(y)\right|^2dy\right|^{\frac{1}{2}}dx,
\end{equation*}
\begin{equation*}
	\| v \|_{L^\infty}:= \sup_{x}\| v(x) \|_{L^2}.
\end{equation*}

\section{Two NLTV Models for Unsupervised  HSI classification}
\label{sec:model}
In this section, two NLTV models are explained for unsupervised classification of HSI. The linear model runs faster in each iteration, but it requires a more accurate centroid initialization. The quadratic model runs slower in each iteration, but it is more robust with respect to the centroid initialization. Moreover, the quadratic model converges faster if the initialization is not ideal.
\subsection{Linear Model}
\label{sec:linearmodel}
We extend the idea from \cite{primaldual} to formulate a linear model for classification on HSI. The linear model seeks to minimize:
\begin{align}
\nonumber
E_1(u) & = \| \nabla_w u\|_{L^1}+ \langle u, f \rangle \\
&=\sum_{l=1}^k\|\nabla _w u_l \|_{L^1}+\sum_{l=1}^{k}\int u_{l}(x)f_{l}(x)dx,
\label{e1}
\end{align}
where $u=(u_1,u_2,\ldots,u_k):\Omega \rightarrow \mathbb{K}^k$ is the labeling function, $k$ is the number of clusters, $\mathbb{K}^k= \{ (x_1, x_2, \ldots, x_k)|\sum_{i=1}^k x_i=1, x_i \ge 0\}$ is the unit simplex in $\mathbb{R}^k$, and $\nabla_w u=(\nabla_w u_1, \ldots, \nabla _w u_k)$ such that $\| \nabla_w u \|_{L^1}=\sum_{l=1}^k\|\nabla_w u_l \|_{L^1}$.  $f_{l}(x)$ is the error function defined as $f_{l}(x)=\frac{\lambda}{2}\left| g(x)-c_{l}\right|_{\mu}^{2}$, where $g(x)$ and $c_l$ are the spectral signatures of pixel $x$ and the $l$-th centroid, which is initially either picked randomly from the HSI or generated by any fast unsupervised centroid extraction algorithm (e.g. H2NMF,  K-means.) The distance in the  definition of $f_l(x)$ is a linear combination of cosine distance and Euclidean distance:
\begin{equation*}
\left|g(x)-c_l\right|_{\mu}=1-\frac{\langle g(x),c_l \rangle}{\| g(x)\|_2\| c_l\|_2}+\mu \| g(x)-c_l\| _2, \quad \mu \ge 0.
\end{equation*}
In HSI processing, the cosine distance is generally used because it is more robust to atmospheric interference and topographical features \cite{distance}. The reason why the Euclidean distance is also used is that sometimes different classes have very similar spectral angles, but vastly different spectral amplitudes (e.g. ``dirt" and ``road" in the Urban dataset, which is illustrated in Section \ref{sec:results}.) This is called the linear model since the power of the labeling function $u_l$ in (\ref{e1}) is one.

The intuition of the model is as follows: In order to minimize the fidelity term $\sum_{l=1}^k\int u_l(x)f_l(x)$, a small $u_l(x)$ is required if $f_l(x)$ is large, while no such requirement is needed if $f_l(x)$ is relatively small. This combined with the fact that $\left(u_1(x),\ldots,u_l(x)\right)$ lies on a unit simplex implies that $u_l(x)$ would be the largest term if pixel $x$ is mostly similar to the $l$-th centroid $c_l$. Meanwhile, the NLTV regularizing term $\sum_{l=1}^k\|\nabla _w u_l \|_{L^1}$ ensures that pixels  similar to each other tend to have analogous values of $u$. Therefore a classification of pixel $x$ can be obtained by choosing the index $l$ that has the largest value $u_l(x)$.

Now we discuss how to discretize (\ref{e1}) for numerical implementation.

\subsubsection{Weight Matrix}
Following the idea from \cite{oshernonlocal},  the patch distance is defined as:\begin{equation*}d_{\sigma}(x,y)=\int_\Omega G_{\sigma}(t)\left|g(x+t)-g(y+t)\right|^2dt,\end{equation*} where $G_{\sigma}$ is a Gaussian of standard deviation $\sigma$. To build a sparse weight matrix, we take a patch $P_i$ around every pixel $i$, and truncate the weight matrix by constructing a $k$-d tree \cite{nn} and searching the $m$ nearest neighbors of $P_i$. $k$-d tree is a space-partitioning data structrue that can significantly reduce the time cost of nearest neighbor search \cite{kdtree}. We employ a randomized and approximate version of this algorithm \cite{flann} implemented in the open source VLFeat package \footnote{\url{http://www.vlfeat.org}}. The weight is binarized by setting all nonzero entries to one. In the experiments, patches of size $3\times 3$ are used, and $m$ is set to 10. Note that unlike RGB image processing, the patch size for HSI does not have to be very large. The reason is that while low dimensional RGB images require spatial context to identify pixels, high dimensional hyperspectral images already encode enough information for each pixel in the spectral dimension. Of course, a larger patch size  that is consistent with the spatial resolution of the HSI will still be preferable when significant noise is present.

\subsubsection{The Labeling Function and the Nonlocal Operators}
The labeling function, $u=(u_1,u_2,\ldots,u_k)$, is discretized as a matrix of size $r\times k$, where $r$ is the number of pixels in the hyperspectral image, and $(u_l)_j$ is the $l$-th labeling function at $j$-th pixel; $(\nabla_wu_l)_{i,j}=\sqrt{w_{i,j}}((u_l)_j-(u_l)_i)$ is the nonlocal gradient of $u_l$; $(\text{div}_wv)_i=\sum_j\sqrt{w_{i,j}}v_{i,j}-\sqrt{w_{j,i}}v_{j,i}$ is the divergence of $v$ at $i$-th pixel; and the discrete $L^1$ and $L^\infty$ norm of $\nabla_wu_l$ are defined as: $\| \nabla_wu_l \|_{L^1}=\sum_i\left(\sum_j(\nabla_wu_l)_{i,j}^2\right)^{\frac{1}{2}}$, and $\| \nabla_wu_l \|_{L^\infty}=\max_i\left(\sum_j(\nabla_wu_l)_{i,j}^2\right)^{\frac{1}{2}}$.

The next issue to address is how to minimize (\ref{e1}) efficiently. The convexity of the energy functional $E_1$ allows us to consider using convex optimization methods. First-order primal-dual algorithms have been successfully used in image processing with $L^1$ type regularizers \cite{primaldual, primaldualchan, primaldualchan2, esser}. We use the primal-dual hybrid gradient (PDHG) algorithm. The main advantage is that no matrix inversion is involved in the iterations, as opposed to general graph Laplacian methods. The most expensive part of the computation comes from sparse matrix multiplications, which are still inexpensive due to the fact that only $m=10$ nonzero elements are kept in each row of the nonlocal weight matrix.

We then address centroid updates and stopping criteria for the linear model. The concept of centroid updates is not uncommon; in fact, the standard K-means algorithm consists of two steps: first, it assigns each point to a cluster whose mean yields the least within-cluster sum of squares, then it re-calculates the means from the centroids, and terminates when assignments no longer change\cite{k-means}. Especially for data-based methods, re-calculating the centroid is essential for making the algorithm less sensitive to initial conditions and more likely to find the ``true" clusters.

After solving (\ref{e1}) using the PDHG algorithm, the output $u$ will be thresholded to $u_{hard}$. More specifically, for every $i\in \{1,2,\ldots,r\}$, the largest element among $\left((u_1)_i,(u_2)_i,\cdots,(u_k)_i\right)$ is set to 1, while the others are set to 0, and we claim the $i$-th pixel belongs to that particular cluster. Then the $l$-th centroid is updated by taking the mean of all the pixels in that cluster.  The process is repeated until the difference between two consecutive $u_{hard}$ drops below a certain threshold. The pseudocode for the proposed linear model on HSI is listed in Algorithm \ref{pdlinear}.

\begin{algorithm}[!t]
  \caption{Linear Model}
  \begin{algorithmic}[1]
    \State Initialization of centroids: Choose $(c_l)_{l=1}^k$ (randomized or generated by unsupervised centroid extraction algorithms). 
    \State Initialization of parameters: Choose $\tau,\sigma>0$ satisfying $\sigma \tau \| \nabla_w \|^2 \le 1$, $\theta = 1$
    \State Initial iterate: Set $u^0 \in \mathbb{R}^{r\times k}$ and $p^0 \in \mathbb{R}^{(r\times r)\times k}$ randomly, set $\bar{u}^0=u^0$, $u_{hard}=threshold(u^0)$
    \While{not converge}
      \State Minimize energy $E_1$ using PDHG algorithm
      \State $u_{hard}=threshold(u)$
      \State Update $(c_l)_{l=1}^k$
    \EndWhile
  \end{algorithmic}
\label{pdlinear}
\end{algorithm}

Before ending the discussion of the proposed linear model, we point out its connection to the piecewise constant Mumford-Shah model for multi-class graph segmentation \cite{MS}. Assume that  the domain $\Omega$ of the HSI is segmented by a contour $\Phi$ into $k$ disjoint regions, $\Omega=\cup_{l=1}^k \Omega_l$. The piecewise constant Mumford-Shah energy is defined as:
\begin{equation}
	E_{MS}(\Phi, \{c_l\}_{l=1}^k)=\left| \Phi \right| + \lambda \sum_{l=1}^k \int_{\Omega_l}\left|g(x)-c_l\right|^2 dx,
\label{eq:MS}
\end{equation}
where $\left|\Phi\right|$ is the length of the contour. To illustrate the connection between (\ref{e1}) and (\ref{eq:MS}), consider the ``local" version of (\ref{e1}), which essentially replaces the NLTV regularizer $\|\nabla_w u_l\|_{L^1}$ with its local counterpart :
\begin{equation}
E_1^{\text{loc}}(u) =\sum_{l=1}^k\|\nabla u_l \|_{L^1}+\sum_{l=1}^{k}\int u_{l}(x)f_{l}(x)dx.
\label{e1local}
\end{equation}
Assume that the labeling function $u_l$ is the characteristic function of $\Omega_l$. Then $\int u_{l}(x)f_{l}(x)dx$ is equal to $\int_{\Omega_l}\left|g(x)-c_l\right|^2 dx$ up to a multiplicative constant. Moreover, the total variation of a characteristic function of a region equals the length of its boundary, and hence $\left|\Phi\right|=\sum_{l=1}^k\|\nabla u_l \|_{L^1}$. So the linear model (\ref{e1}) can be viewed as a nonlocal convex-relaxed version of Mumford-Shah model. We  also note that the linear energy (\ref{e1}) has been studied in \cite{huiyi_plume}. But in their work, the authors used a graph-based MBO method to minimize (\ref{e1}) instead of the PDHG algorithm, and the difference of the numerical performances can be seen in Section \ref{sec:results}.

\subsection{Quadratic Model}
\subsubsection{Intuition}\label{intuition}
The aforementioned linear model performs very well when the centroids are initialized by accurate centroid extraction algorithms. As shown in Section \ref{sec:results}, the linear model can have a significant boost to the accuracy of other algorithms if the centroid extraction algorithm is reasonable, without sacrificing speed. However, if centroids are not extracted accurately, or if random  initialization is used, the segmenting results are no longer reliable, and the algorithm takes far more iterations to converge to a stable classification.

To reduce the times of centroid updates and merge similar clusters automatically and simultaneously,  the following quadratic model is proposed:
\begin{equation}
E_2(u)=\sum_{l=1}^k\|\nabla_w u_l \|_{L^1}+ \sum_{l=1}^{k}\int u_{l}^2(x)f_{l}(x)dx \label{e2}.
\end{equation}

Similar as before, $u=(u_1,u_2,\ldots,u_k):\Omega \rightarrow \mathbb{K}^k$ is the labeling function, $k$ is the number of clusters, $\mathbb{K}^k$ is the unit simplex in $\mathbb{R}^k$, and $f_{l}(x)$ is the error function.

Note that the only difference between (\ref{e1}) and (\ref{e2}) is that the power of the labeling function $u_l$ here is two. The intuition for this is as follows:

\begin{figure}[!t]
\centering
\begin{tabular}{cc}
  \includegraphics[width=1.3in]{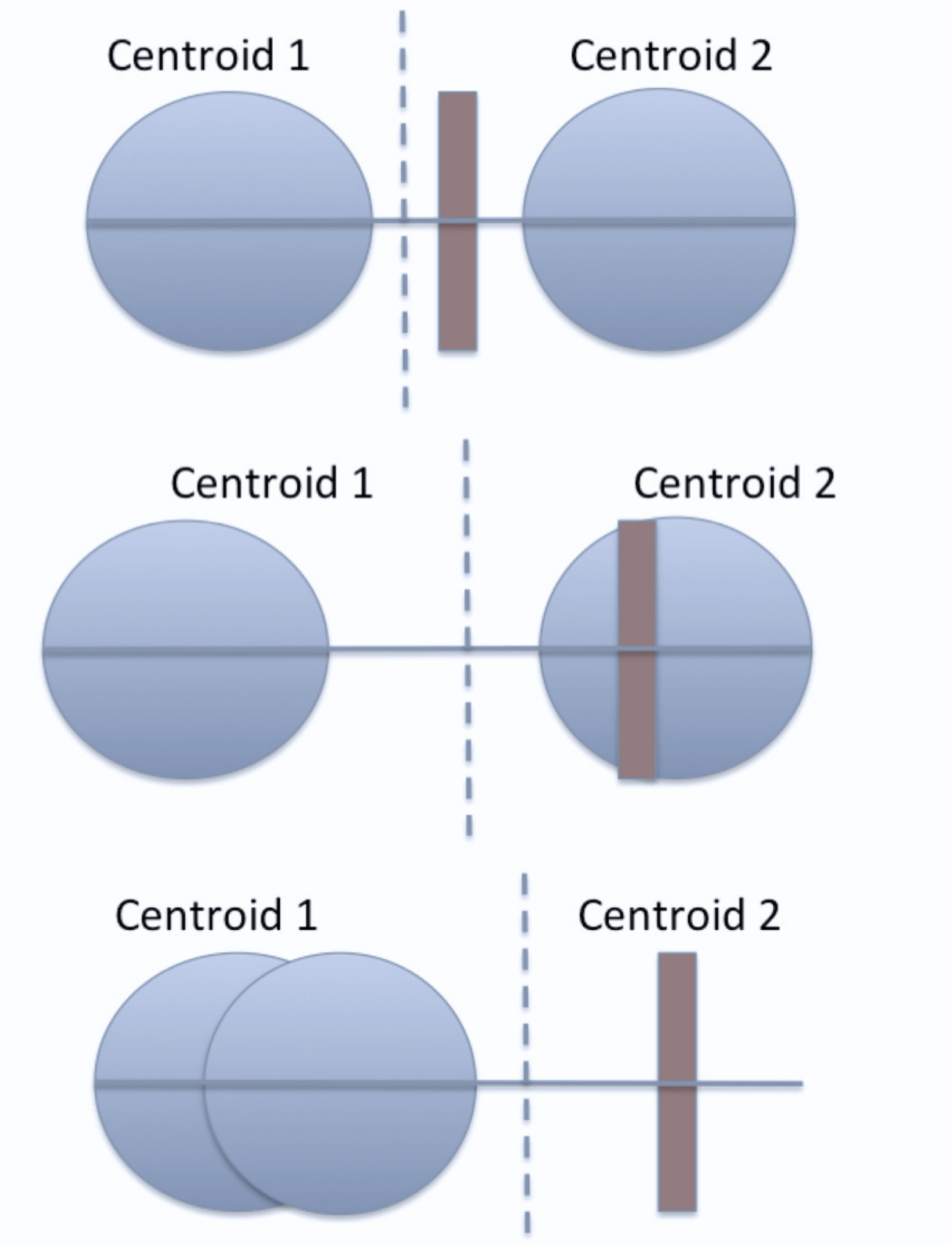}  &
  \includegraphics[width=1.7in]{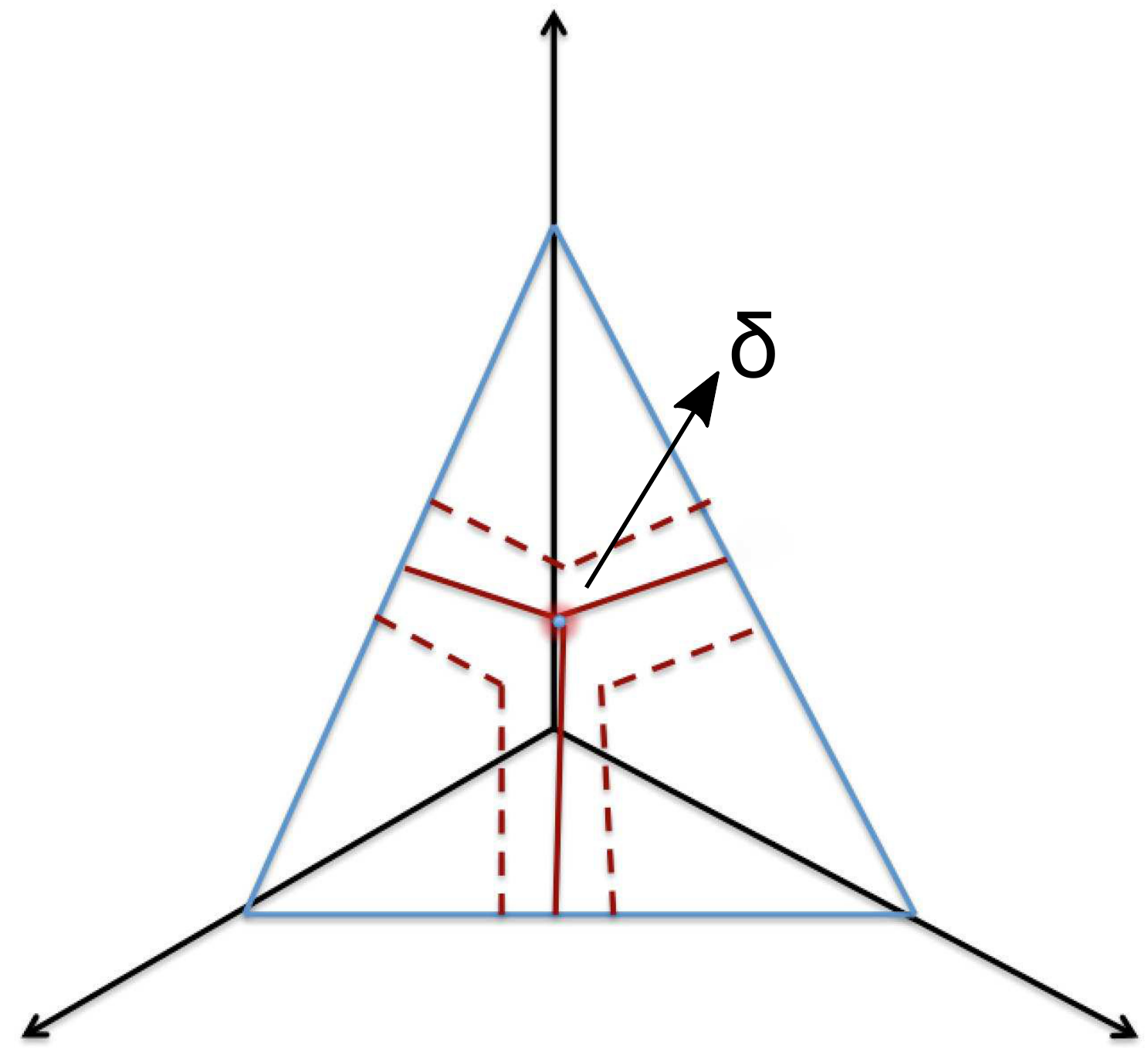}
\end{tabular}
\caption[caption]{The first figure shows the ``pushing'' mechanism of the quadratic model. The horizontal line represents the unit simplex in $\mathbb{R}^2$. Signatures from cluster $A_1$ are colored blue, and signatures from cluster $A_2$ are colored brown. The vertical dashed bar is generated by a stable simplex clustering method, and it thresholds the points on the simplex into two categories. \\\hspace{\textwidth}The second figure shows the stable simplex clustering. Every grid point $\delta$ on the simplex generates a simplex clustering. We want to choose a $\delta$ such that there are very few data points falling into the ``Y-shaped region".}
\label{pushing}
\end{figure}

Consider for simplicity a hyperspectral image with a ground truth of only two clusters, $A_{1}$ and $A_{2}$. Suppose the randomized initial centroids are chosen such that $c_{1} \approx c_{2} \in A_{1}$; or, that the two random initial pixels are of very similar spectral signatures and belong to the same ground truth cluster. 

Let $x$ be a pixel from $A_{2}$. Then $0\ll \left|g(x)-c_1\right|^2\approx \left|g(x)-c_2\right|^2$.  When  (\ref{e1}) is applied, the fidelity term $ \langle u, f\rangle$ does not change when $u(x)$ moves on the simplex in $\mathbb{R}^{2}$, and thus pixels of $A_{2}$ will be scattered randomly on the simplex. After thresholding, an approximately equal number of pixels from cluster $A_{2}$ will belong to clusters $C_{1}$ and $C_{2}$, so the new centroids $\tilde{c}_{1}$ and $\tilde{c}_{2}$ that are the means of the spectral signatures of the current clusters will once again be approximately equal. 

This situation changes dramatically when (\ref{e2}) is minimized: 
\begin{itemize}
\item Observe that the fidelity term in $E_2$ is minimized for a pixel $x \in A_{2}$ when $u_{1}(x) \approx u_{2}(x) \approx \frac{1}{2}$. Therefore, the pixels of cluster $A_{2}$ will be ``pushed" toward the center of the simplex once $E_2$ is minimized. 
\item With a stable simplex clustering method (explained in Section \ref{sec:simplex}), the clusters are divided such that all of these pixels in the center belong to either $C_{1}$ or $C_{2}$; without loss of generality suppose they belong to $C_{2}$. Then the updated centroid $\tilde{c}_1$ is essentially $c_1$, while the updated centroid $\tilde{c}_2$ is a linear combination of the spectral signature of members belonging to $A_{1}$ and $A_{2}$, and thus quite different from the original $c_2$. 
\item After minimizing the energy $E_2$ again, pixels from $A_1$ will be clustered in $C_1$, and pixels from $A_2$ will be pushed to $C_2$. Therefore, the clustering will be finished in just two steps in theory. See Fig.\ref{pushing} for a graphical illustration.
\end{itemize}

The quadratic model not only reduces the number of iterations needed to find the ``true" clustering because of its capability of anomaly detection, but it allows for random initialization as well, making it a more robust technique.

\subsubsection{Stable Simplex Clustering}
\label{sec:simplex}

As mentioned above, the quadratic model pushes anomalies into the middle of the unit simplex. Therefore it would be ill-conceived to simply classify the pixels based on the largest component of the labeling function $u(x)=(u_1(x),u_2(x),\ldots,u_k(x))$. Instead,  a stable simplex clustering method has to be used.

The concept behind the stable simplex clustering is to choose a division that puts all the data points in the ``middle" of the unit simplex into a single cluster. Fig. \ref{pushing} demonstrates this in the simple two-cluster case. Also refer to section \ref{intuition} for explanation of the ``pushing" process. The idea to accomplish this goal is inspired by \cite{kuang}. We first create a grid on a \textit{k}-dimensional simplex, where $k$ is the number of clusters, and each grid point $\delta$ generates a simplex clustering. Then a  $\delta$ is searched to minimize the energy $g(\delta)$:
\begin{equation*}
g(\delta)=-\log(\prod_{l=1}^k F_l(\delta))+\eta \exp (G(\delta)),
\end{equation*}
where $F_l(\delta)$ is the percentage of data points in cluster $l$, and $G(\delta)$ is the percentage of data points on the edges near the division, i.e. the ``Y-shaped region'' in Figure \ref{pushing}. The first term in $g(\delta)$ rewards keeping clusters approximately of the same size, ensuring no skewed data from clusters far too small. And the second term rewards sparsity of points in the intermediate region. The constant $\eta$ is chosen to be large enough such that stability has a bigger weight in the energy.

\begin{figure}[!t]
  \centering
  \includegraphics[width=3.5in]{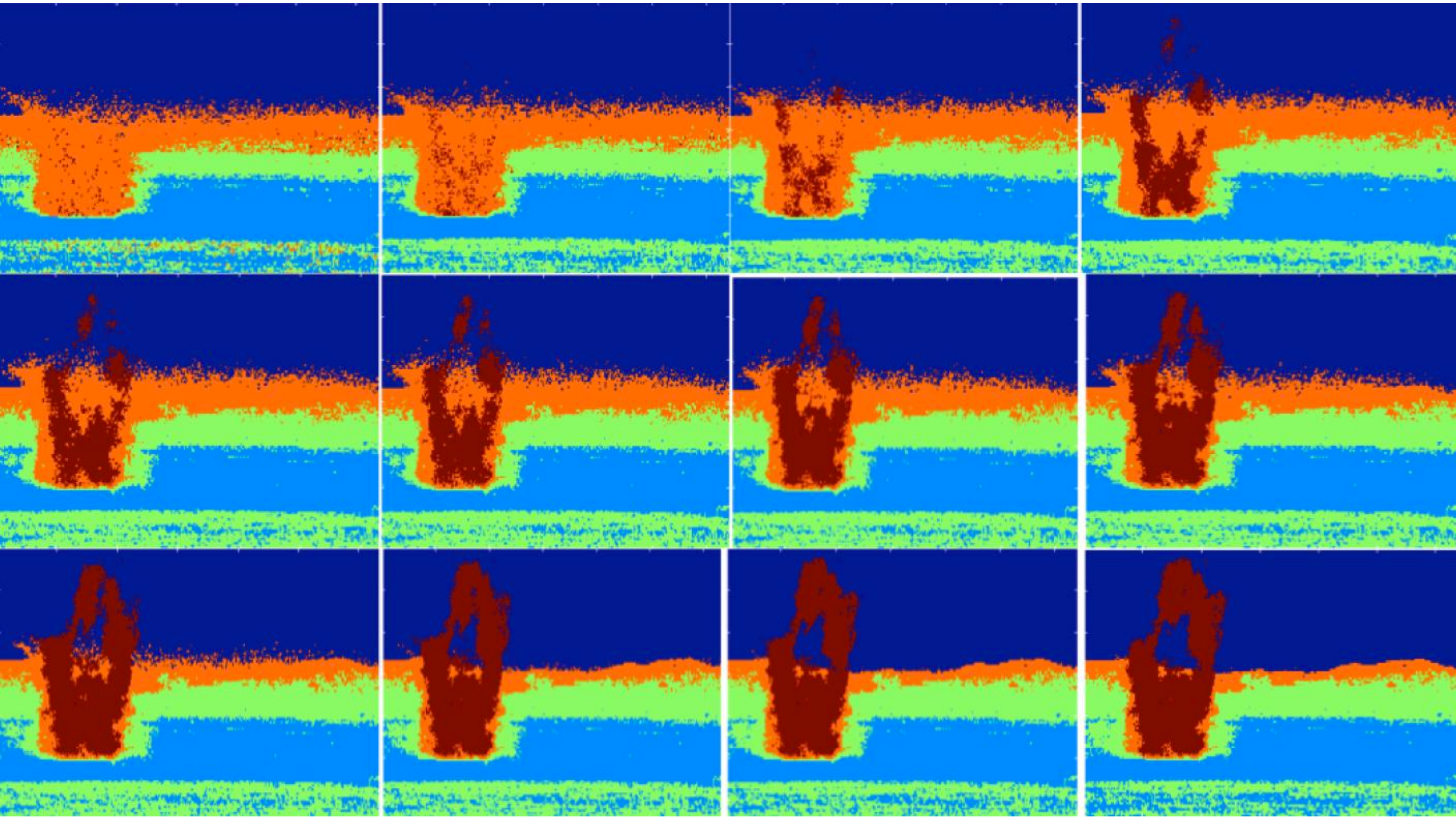}
  \caption{Quadratic model and stable simplex clustering on the plume dataset. The chemical plume (brown) is perfectly detected in 12 iterations.}\label{u2_plume}
\end{figure} 

\begin{figure*}[!t]
  \centering
  \begin{tabular}{cccc}
    Linear, Iteration=1&Linear, Iteration=16&
    Linear, Iteration=32&Linear, Iteration=50\\
    \includegraphics[width=1.3in]{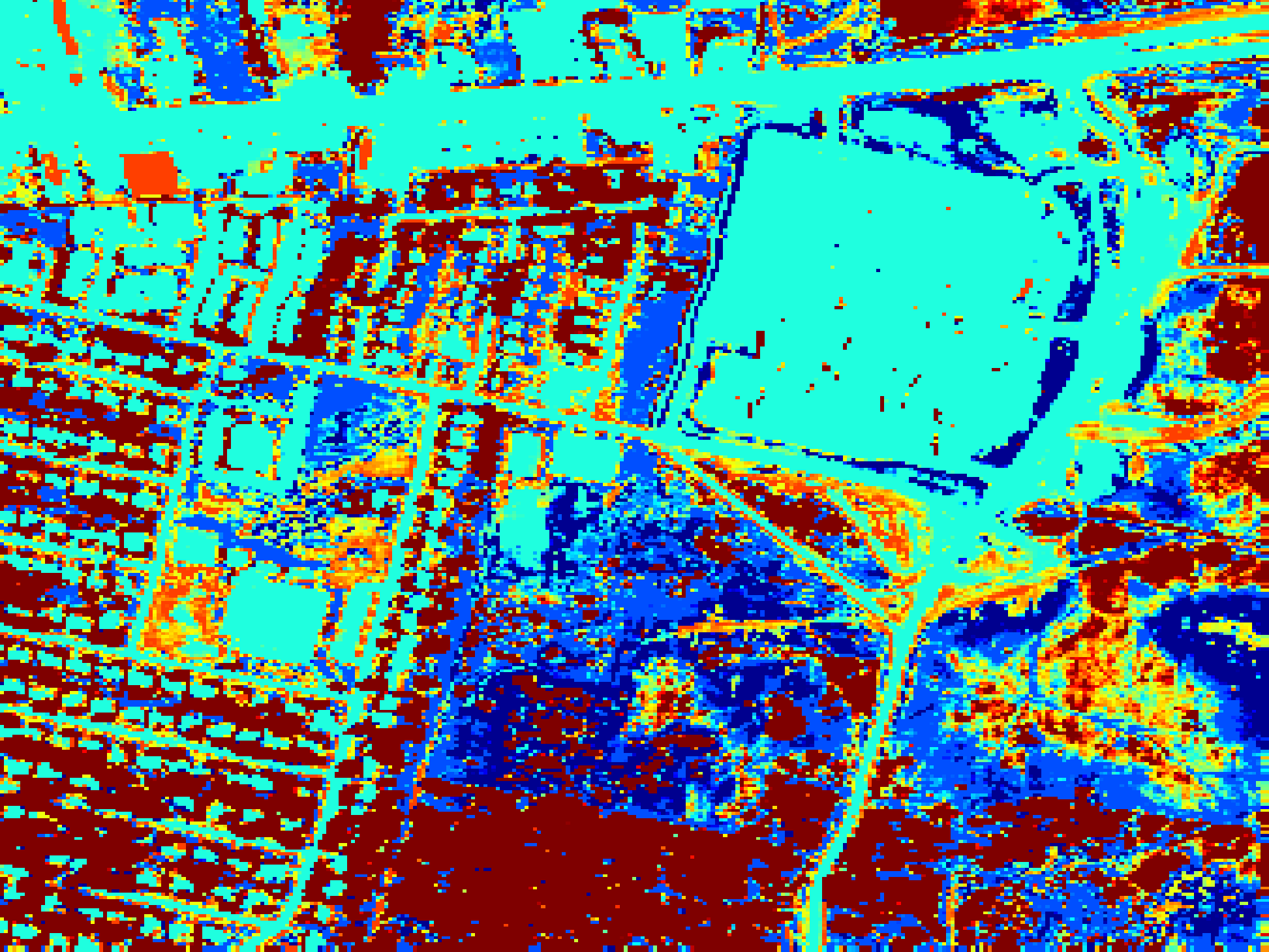}&
    \includegraphics[width=1.3in]{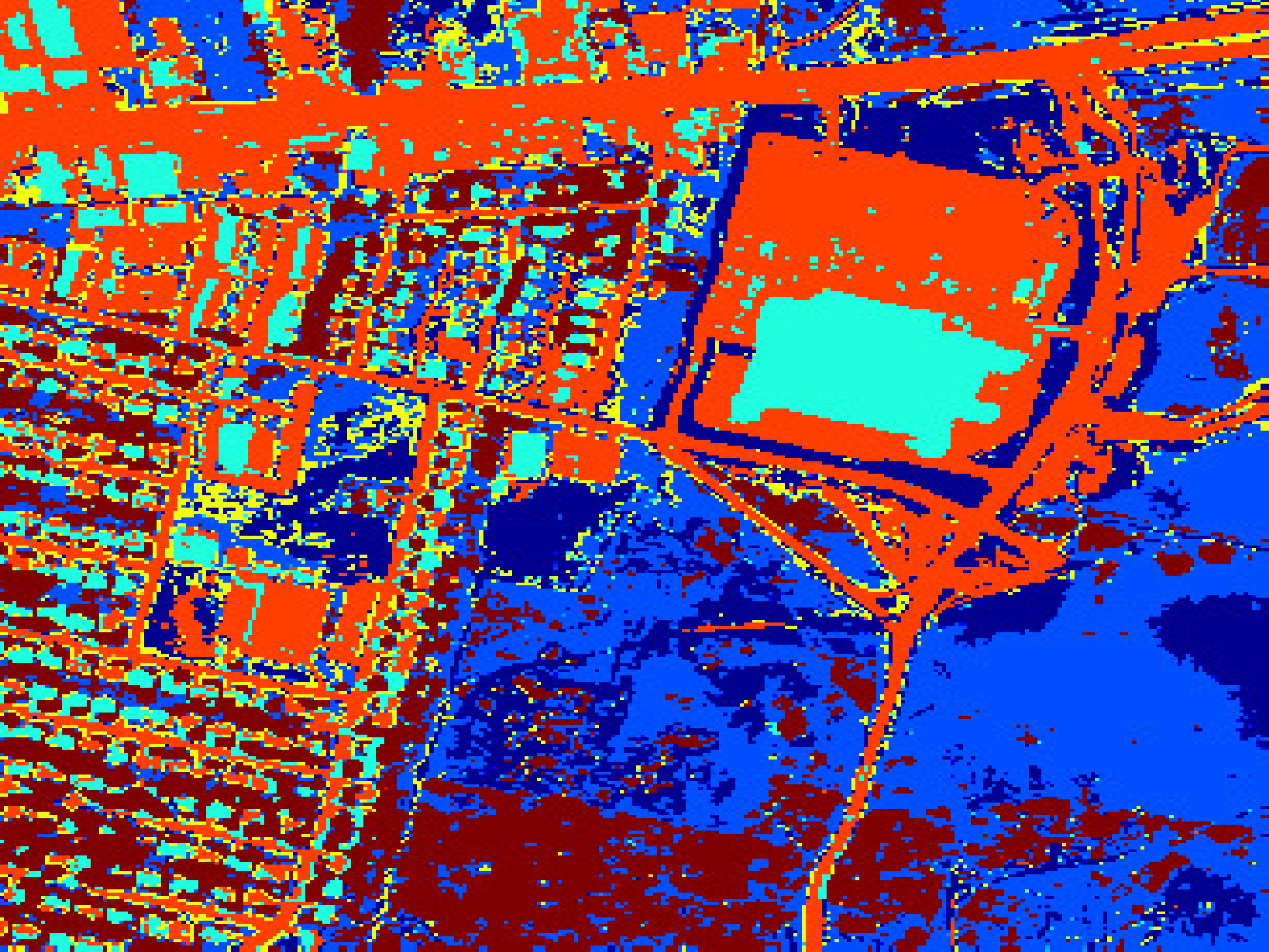}&
    \includegraphics[width=1.3in]{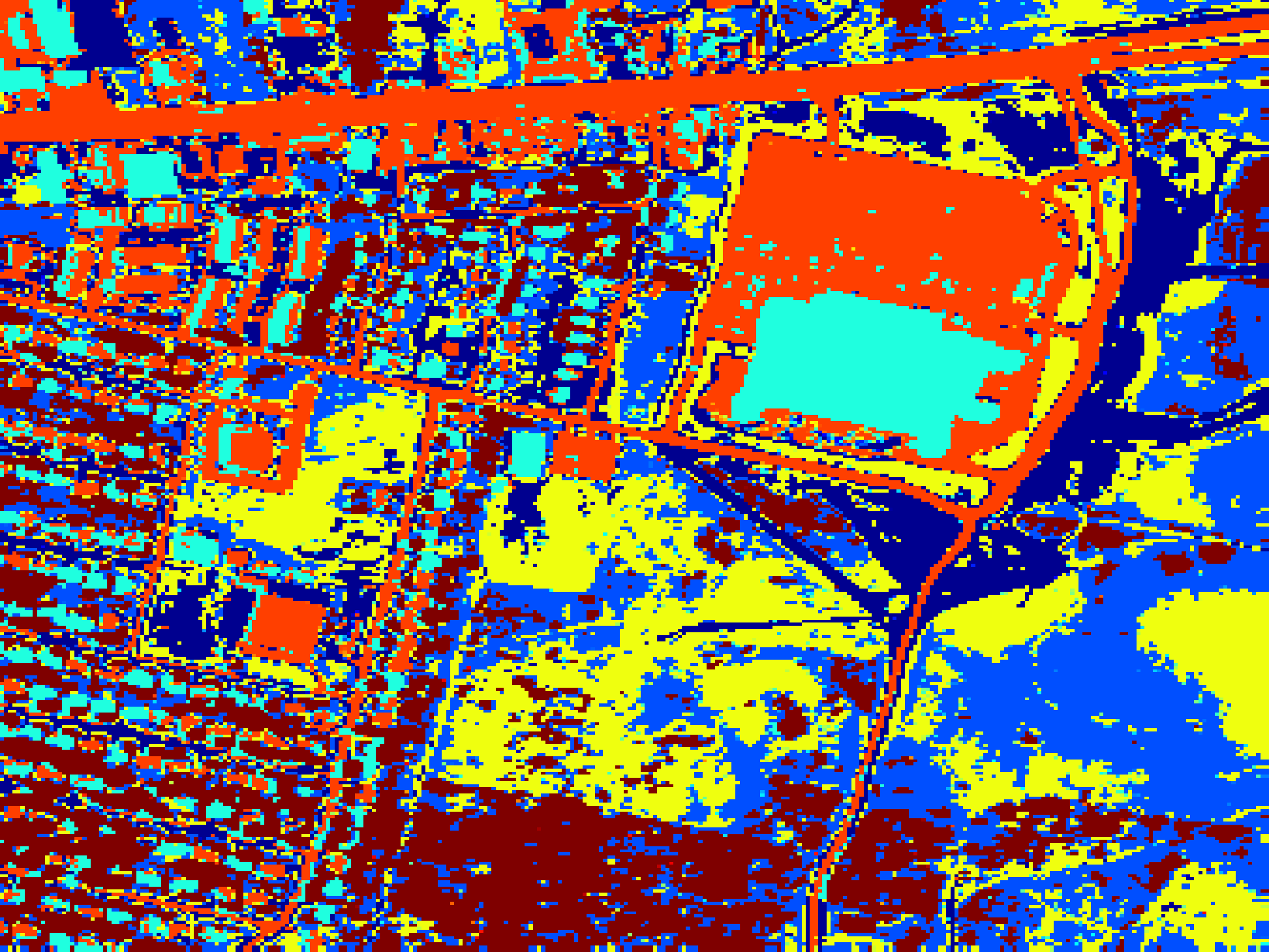}&
    \includegraphics[width=1.3in]{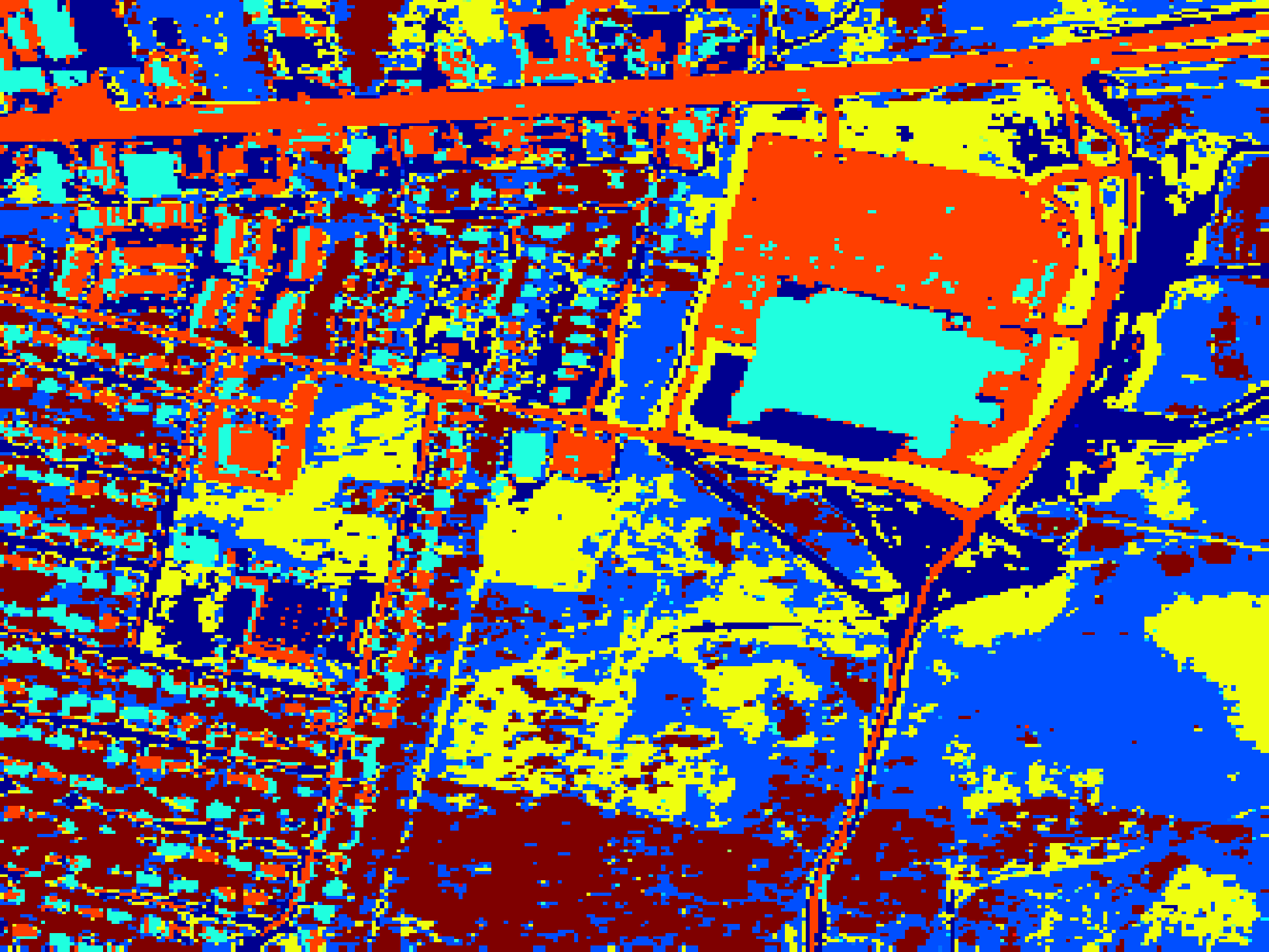}\\
    Quadratic, Iteration=1&Quadratic, Iteration=2&
    Quadratic, Iteration=3&Quadratic, Iteration=4\\
    \includegraphics[width=1.3in]{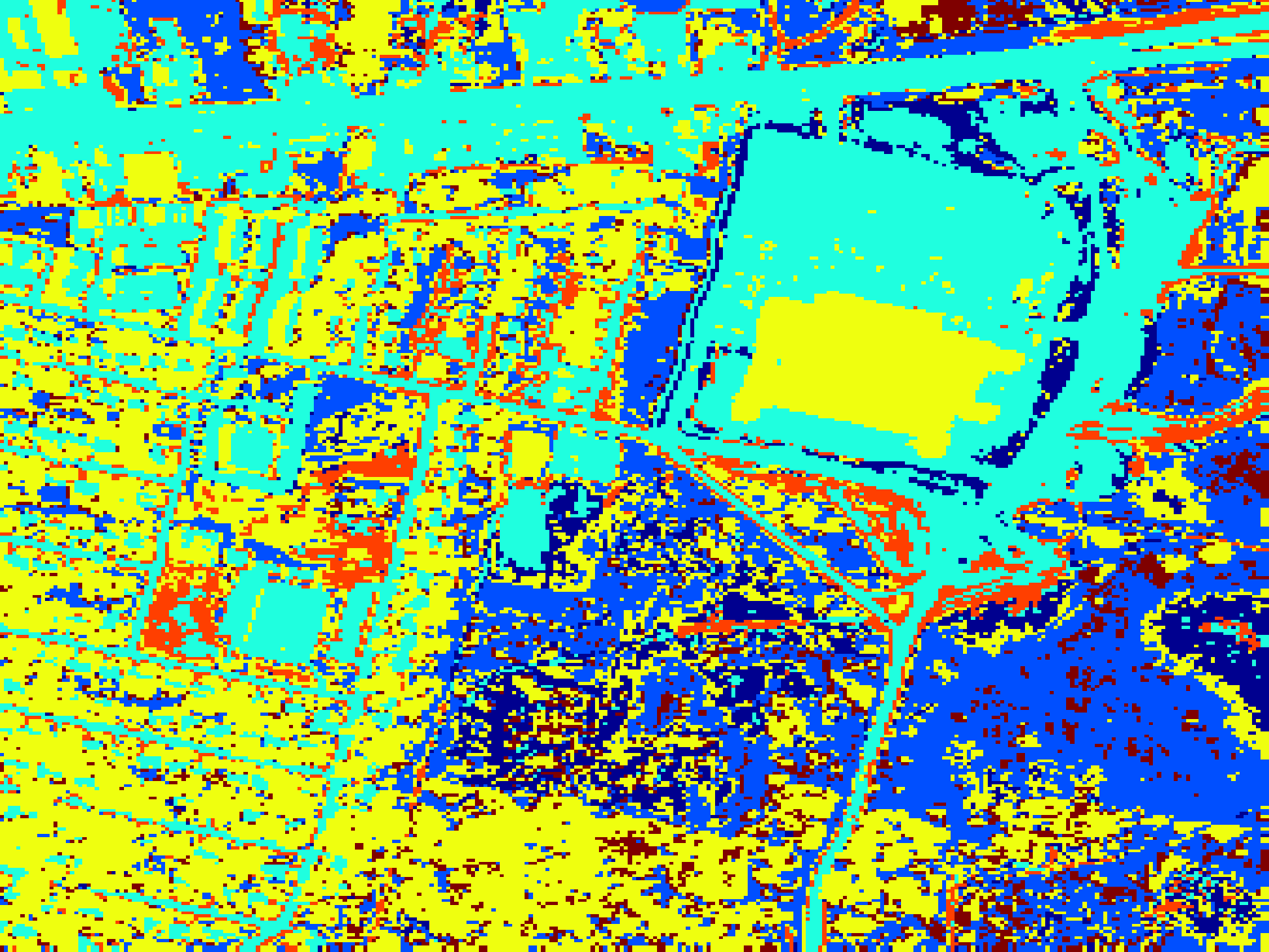}&
    \includegraphics[width=1.3in]{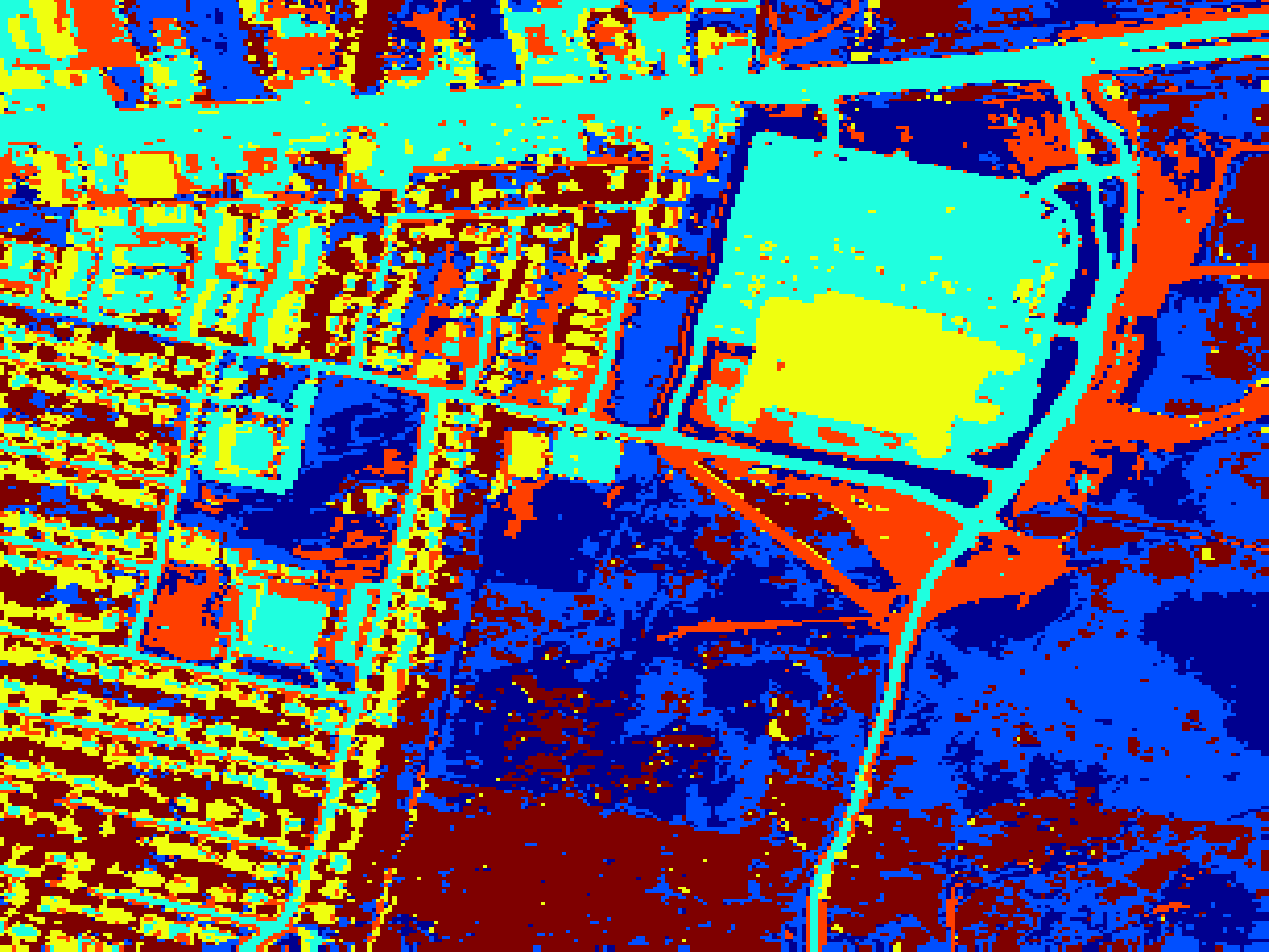}&
    \includegraphics[width=1.3in]{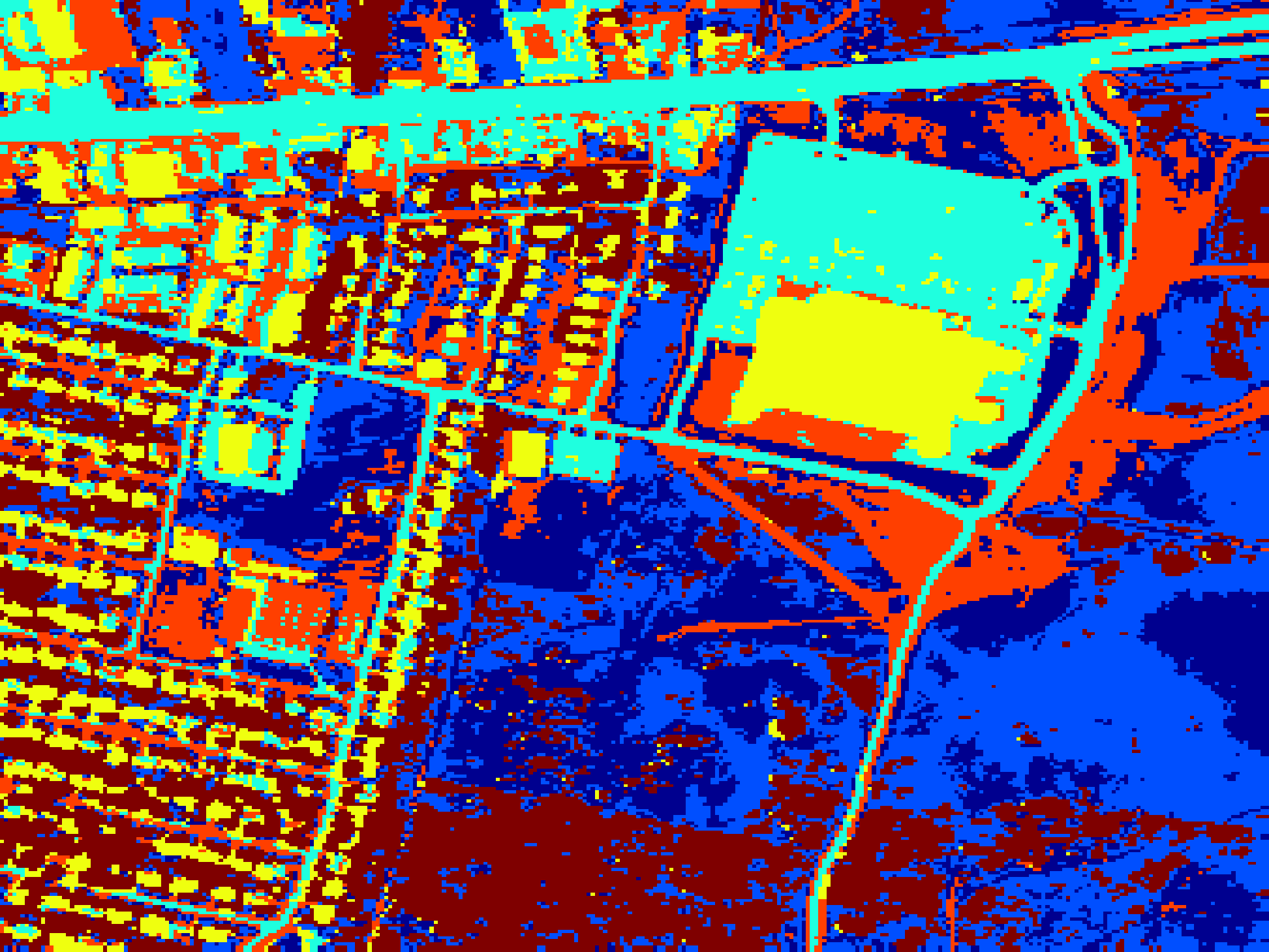}&
    \includegraphics[width=1.3in]{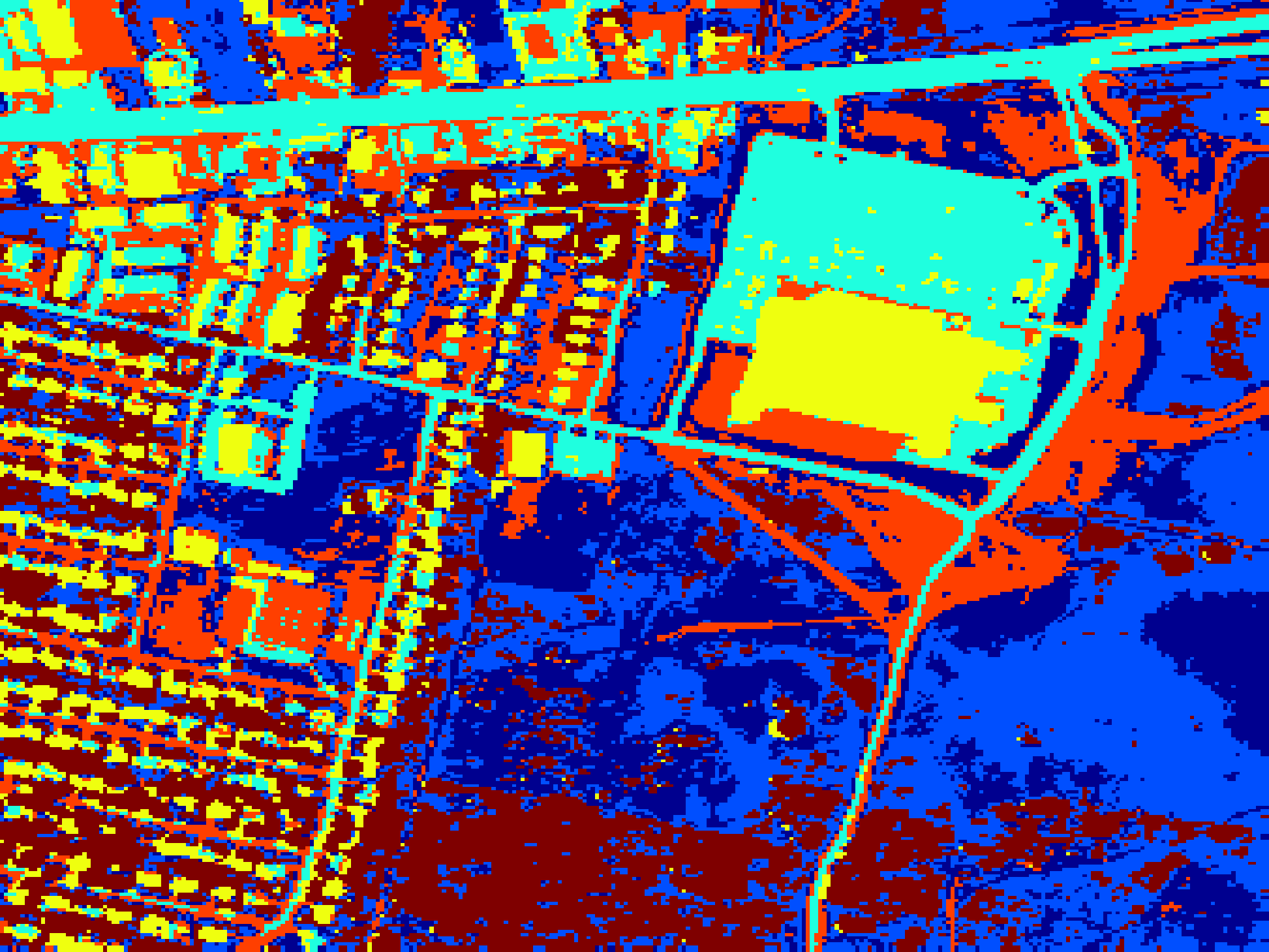}
  \end{tabular}
  \caption{Linear vs Quadratic Model on the Urban dataset with the same centroid initialization. To produce essentially identical results, the Linear model (first row) took 50 iterations of centroid updates, and the Quadratic model (second row) took just 4 iterations.}\label{u2_urban}
  \label{u2_urban}
\end{figure*}

\begin{algorithm}[!t]
  \caption{Quadratic Model with Stable Simplex Clustering}
  \begin{algorithmic}[1]
    \State Initialization of centroids: Choose $(c_l)_{l=1}^k$ (randomized or generated by unsupervised centroid extraction algorithms). 
    \State Initialization of parameters: Choose $\tau,\sigma>0$ satisfying $\sigma \tau \| \nabla_w \|^2 \le 1$, $\theta = 1$
    \State Initial iterate: Set $u^0 \in \mathbb{R}^{r\times k}$ and $p^0 \in \mathbb{R}^{(r\times r)\times k}$ randomly, set $\bar{u}^0=u^0$,
    \While{not converge}
      \State Minimize energy $E_2$ using PDHG algorithm
      \State $u_{hard}=threshold(u)$ with stable simplex clustering
      \State Update $(c_l)_{l=1}^k$
    \EndWhile
  \end{algorithmic}
\label{pdquadratic}
\end{algorithm}

Algorithm \ref{pdquadratic} shows the quadratic model using stable simplex clustering. Fig. \ref{u2_plume} demonstrates how this detected the chemical plumes in a frame with background centroids pre-calculated and random initialization for the final centroid. Notice that no plume is detected in the first iteration. But by the twelfth iteration, the gas plume is nearly perfectly segmented.

Finally, we present the comparison between the results of the linear model and the quadratic model on the Urban dataset with identical random pixel initialization in Figure \ref{u2_urban}. The linear model took about 50 iterations to converge, and the quadratic model only took 4 iterations.

\section{Primal-Dual Hybrid Gradient Algorithm}
\label{sec:pd}
In this section, a detailed explanation is provided on the application of the PDHG algorithm \cite{primaldual, primaldualchan, primaldualchan2, esser} to minimizing $E_1$ (\ref{e1}) and $E_2$ (\ref{e2}) in the previous section. A review of the algorithm is provided in a more general setting to contextualize the extension to nonlocal model for hyperspectral imagery.

\subsection{A Review of PDHG Algorithm}

Consider the following convex optimization problem:
\begin{align}
  \label{eq:generalprimal}
  \min_{x\in X}\{F(Kx)+G(x)\},
\end{align}
where $X$ and $Y$ are finite-dimensional real vector spaces,  $F$ and $G$ are proper convex lower semi-continuous functions $F:Y\rightarrow[0, \infty]$, $G:X\rightarrow[0, \infty]$, and $K:X\rightarrow Y$ is a continuous linear operator with the operator norm
$\| K \| = \sup\{\| Kx\| : x \in X, \| x\|\leq 1\}$. The primal-dual formulation of (\ref{eq:generalprimal}) is the saddle-point problem:
\begin{align}
  \label{generalprimaldual}
  \min_{x\in X}\max_{y\in Y}\{\langle Kx, y \rangle - F^*(y) + G(x)\},
\end{align}
where $F^*$ is the convex conjugate of $F$ defined as $F^*(y)= \sup_x \left<x,y\right>-F(x)$

The saddle-point problem (\ref{generalprimaldual}) is then solved using the iterations of Algorithm \ref{pditer} from \cite{primaldual}.

\begin{algorithm}[!t]
  \caption{Primal-Dual Hybrid Gradient (PDHG) Algorithm}
\begin{algorithmic}[1]
\State Initialization: Choose $\tau,\sigma>0$, $\theta \in [0,1]$, $(x^0,y^0)\in X\times Y$, and set $\bar{x}^0=x^0$
\While {not converge} 
  \State $y^{n+1}=(I+ \sigma \partial F^*)^{-1}(y^n+\sigma K \bar{x}^n)$
  \State $x^{n+1}=(I+\tau \partial G)^{-1}(x^n-\tau K^* y^{n+1})$
  \State $\bar{x}^{n+1}=x^{n+1}+\theta (x^{n+1}-x^n)$
  \State $n=n+1$
\EndWhile
\end{algorithmic}
\label{pditer}
\end{algorithm}

In Algorithm \ref{pditer}, $(I+ \lambda \partial f)^{-1}(x)$ is the proximal operator of $f$, which is defined as:
\begin{equation*}
(I+ \lambda \partial f)^{-1}(x)=\text{prox}_{\lambda f}(x)=\arg \min_y f(y)+\frac{1}{2\lambda}\|y-x\|_2^2.
\end{equation*}

It has been shown in \cite{primaldual} that $O(1/N)$ (where $N$ is the number of iterations) convergence can be achieved as long as $\sigma,\tau$ satisfy $\sigma \tau \| K \|^2 \le 1.$

\subsection{Primal-Dual Iteraions to Minimize $E_1$ and $E_2$}
Recall from Section \ref{sec:model} that the discretized linear and quadratic energy $E_1$ and $E_2$ are:
\begin{align*}
  E_1(u) &=\sum_{l=1}^k\|\nabla _w u_l \|_{L^1}+\sum_{l=1}^{k}\sum_{i=1}^r (u_{l})_i(f_{l})_i,\\
         & = \| \nabla_w u\|_{L^1}+ \langle u, f \rangle,\\
  E_2(u) &=\sum_{l=1}^k\|\nabla _w u_l \|_{L^1}+\sum_{l=1}^{k}\sum_{i=1}^r (u_{l})_i^2(f_{l})_i,\\
         & = \| \nabla_w u\|_{L^1}+ \langle u, f\odot u \rangle,
\end{align*}
where  $u=(u_1,u_2,\ldots,u_k)$ is a nonegative matrix of size $r\times k$, with each row of matrix $u$ summing to one, and $f \odot u$ denotes the pointwise product between two matrices $f$ and $u$. After adding an indicator function $\delta_U$, minimizing $E_1$ and $E_2$ are equivalent to solving (\ref{e1primal}) and (\ref{e2primal}):
\begin{align}
  \label{e1primal}
  &\underset{u}{\text{min}} \| \nabla_wu\|_{L^1}+\langle u,f\rangle+ \delta_U(u),\\
  \label{e2primal}
  &\underset{u}{\text{min}} \| \nabla_wu\|_{L^1}+\langle u,f \odot u\rangle+ \delta_U(u)
\end{align}
where $U=\{u=(u_1,u_2,\ldots,u_k)\in \mathbb{R}^{r\times k}: \sum_{l=1}^k  (u_l)_i=1, \forall i=1,\ldots, r, (u_l)_i \ge 0\}$, and $\delta_U$ is the indicator function on $U$. More specifically:
\begin{equation}
  \delta_U(u)=
  \begin{cases}
    0 \hspace{16pt} \text{if } u\in U,\\
    \infty \hspace{10pt} \text{otherwise}.
  \end{cases}
  \label{eq:indicator}
\end{equation}

By comparing (\ref{e1primal}), (\ref{e2primal}) and (\ref{eq:generalprimal}), we can set $K_1 =K_2 = \nabla_w$,  $F_1(q) = F_2(q) = \|q\|_{L^1}$,  $G_1(u)=\left<u, f\right>+\delta_U(u)$,  and $G_2(u)= \left<u, f\odot u\right>+\delta_U(u)$.
The convex conjugate of $F_1$ (and $F_2$) is $F_1^*(p)=F_2^*(p)=\delta_P(p)$, where the set $P=\{p\in \mathbb{R}^{(r\times r)\times k}: \| p_l \|_\infty \le 1\}$. 

Next, we derive the closed forms of the proximal operators $(I+\sigma\partial F^*_{1,2})^{-1}$ and $(I+\tau \partial G_{1,2})^{-1}$ so that Algorithm \ref{pditer} can be implemented efficiently to minimize $E_1$ and $E_2$.
\begin{align}
  \label{eq:update_p}\nonumber
  &(I+\sigma \partial F_{1,2}^*)^{-1}(\tilde{p})=(I+\sigma \partial \delta_P)^{-1}(\tilde{p})\\ 
& = \arg \min_p \delta_P(p)+\frac{1}{2\sigma}\|p-\tilde{p}\|_2^2 = \text{proj}_P(\tilde{p}),
\end{align}
where $\text{proj}_P(\tilde{p})$ is the projection of $\tilde{p}$ onto the closed convex set $P$.
\begin{align}
  \nonumber
  &(I+\tau \partial G_1)^{-1}(\tilde{u})  = \arg \min_u \left< u,f \right> + \delta_U(u)+ \frac{1}{2\tau}\|u-\tilde{u}\|_2^2\\  \label{eq:update_u_e1}
& = \arg \min_{u \in U} \|u-\tilde{u} + \tau f\|_2^2 = \text{proj}_U(\tilde{u}-\tau f).\\ \nonumber
 & (I+\tau \partial G_2)^{-1}(\tilde{u})= \arg\min_u \left<u,\frac{\tau}{2}\mathcal{A} u\right> +\tau\delta_U(u)+\frac{1}{2}\|u-\tilde{u}\|_2^2\\ \nonumber
& = \arg \min_{u \in U} \frac{1}{2}\left<u,(I+\tau \mathcal{A})u\right>-\left<u,\tilde{u}\right>+\frac{1}{2}\left<\tilde{u},(I+\tau \mathcal{A})^{-1}\tilde{u}\right>\\
\label{eq:update_u_e2}
& = \arg \min_{u \in U} \frac{1}{2}\|(I+\tau \mathcal{A})^{\frac{1}{2}}u-(I+\tau \mathcal{A})^{-\frac{1}{2}}\tilde{u}\|^2_2,
\end{align}
where $\mathcal{A}:\mathbb{R}^{r\times k}\rightarrow \mathbb{R}^{r\times k}$ is a linear operator defined as  $\frac{1}{2}\mathcal{A}u=f\odot u$. Therefore $\mathcal{A}$ is a positive semidefinite diagonal matrix of size $rk \times rk$. It is worth mentioning that the matrix $(I+\tau \mathcal{A})$ is diagonal and positive definite, and hence it is trivial to compute its inverse and square root. Problem (\ref{eq:update_u_e2}) can be solved as a preconditioned projection onto the unit simplex $\mathbb{K}^k$, and the  solution will be explained in Section \ref{sec:precondition}.

Combining (\ref{eq:update_p},\ref{eq:update_u_e1},\ref{eq:update_u_e2}) and Algorithm \ref{pditer}, we have the primal-dual iterations for minimizing $E_1$ (Algorithm \ref{pditerlinear}) and $E_2$ (Algorithm \ref{pditerquadratic}).

\begin{algorithm}[!t]
    \caption{Primal-Dual Iterations for the Linear Model}
    \begin{algorithmic}[1]
      \While{not converge}
        \State $p^{n+1}=\text{proj}_P(p^n+\sigma\nabla_w\bar{u}^n)$
        \State $u^{n+1}=\text{proj}_U(u^n+\tau \text{div}_wp^{n+1}-\tau f)$
        \State $\bar{u}^{n+1}=u^{n+1}+\theta (u^{n+1}-u^n)$
        \State $n=n+1$
      \EndWhile
    \end{algorithmic}
 \label{pditerlinear}
\end{algorithm}

\begin{algorithm}[!t]
    \caption{Primal-Dual Iterations for the Quadratic Model}
    \begin{algorithmic}[1]
      \While{not converge}
        \State $p^{n+1}=\text{proj}_P(p^n+\sigma \nabla_w \bar{u}^n)$        \State Update $u^{n+1}$ as in (\ref{eq:update_u_e2}), where $\tilde{u}=u^n+\tau \text{div}_w p^{n+1}$
        \State $\bar{u}^{n+1}=u^{n+1}+\theta (u^{n+1}-u^n)$
        \State $n=n+1$
      \EndWhile
    \end{algorithmic}
\label{pditerquadratic}
\end{algorithm}

Before moving on to explaining how to solve (\ref{eq:update_u_e2}), we specify the two orthogonal projections $\text{proj}_P$ and $\text{proj}_U$ in Algorithm \ref{pditerlinear}: Let $\tilde{p}=\text{proj}_P(p)$, where $p=(p_l)_{l=1}^k \in \mathbb{R}^{(r\times r)\times k}$. Then for every $i\in \{1,2,\ldots,r\}$ and every $l\in \{1,2,\ldots,k\}$, the $i$-th row of $\tilde{p}_l$ is the projection of the $i$-th row of $p_l$ on to the unit ball in $\mathbb{R}^r$. Similarly, if $\tilde{u}=\text{proj}_U(u)$, then for every $i\in \{1,2,\ldots,r\}$, $((\tilde{u}_1)_i,(\tilde{u}_2)_i,\ldots,(\tilde{u}_k)_i)$ is the projection of $((u_1)_i,(u_2)_i,\ldots,(u_k)_i)$ onto the unit simplex $\mathbb{K}^k$ in $\mathbb{R}^k$.

\subsection{Preconditioned Projection onto the Unit Simplex}
\label{sec:precondition}
This section is dedicated to solving (\ref{eq:update_u_e2}). It is easy to see that the rows of $u$ in (\ref{eq:update_u_e2}) are decoupled, and the only problem that needs to be solved is:
\begin{equation}
  \label{eq:u}
  \underset{u\in \mathbb{R}^k}{\text{min }}\delta_{\mathbb{K}^k}(u)+\frac{1}{2}\| Au-y\|^2,
\end{equation}
where $A=\text{diag}(a_1,a_2,\ldots,a_k)$ is a positive definite diagonal matrix of size $k\times k$, $\mathbb{K}^k$ is the unit simplex in $\mathbb{R}^k$, and $y\in \mathbb{R}^k$ is a given vector.

\newtheorem{theorem}{\textbf{Theorem}}
\begin{theorem}
\label{thm:1}
  The solution $u=(u_1,u_2,\ldots,u_k)$ of (\ref{eq:u}) is:
  \begin{equation}
    \label{eq:solution_u}
    u_i = \max \left(\frac{a_iy_i-\lambda}{a_i^2},0\right),
  \end{equation}
where $\lambda$ is the unique number satisfying:
\begin{equation}
  \label{eq:solution_lambda}
  \sum_{i=1}^k\max\left(\frac{a_iy_i-\lambda}{a_i^2},0\right)=1
\end{equation}
\end{theorem}

The proof of Theorem \ref{thm:1} is shown in the Appendix. The most computationally expensive part of solving (\ref{eq:solution_lambda}) is sorting the sequence $\left( a_iy_i \right)_{1 \le i \le k}$ of length $k$, which is trivial since $k$, the number of clusters, is typically a small number.

\section{Numerical Results}
\label{sec:results}
\subsection{Comparison Methods and Experimental Setup}
\label{sec:setup}
 All experiments were run on a Linux machine with Intel core i5, 3.3Hz with 2GB of DDR3 RAM. The following unsupervised algorithms have been tested:

\begin{enumerate}
	\item \textbf{(Spherical) K-means}: Built in MatLab Code.
	\item \textbf{NMF}: Non-negative Matrix Factorization \cite{nmfcode}.
	\item \textbf{H2NMF}: Hierarchical Rank-2 Non-negative Matrix Factorization \cite{kuang}.
	\item \textbf{MBO}: Graph Merriman-Bence-Osher scheme \cite{ekaterina_plume, huiyi_plume}. The code is run for 10 times on each dataset, and the best result is chosen.
	\item \textbf{NLTV2}: Nonlocal Total Variation, quadratic model with random pixel initialization.
	\item \textbf{NLTV1(H2NMF/K-means)}: Nonlocal Total Variation, linear model with endmembers/centroids extracted from H2NMF/K-means.
\end{enumerate}

Every algorithm can be initialized via the same procedure as that in ``K-means++''\cite{kmeans++}, and the name  ``Algorithm++'' is used if the algorithm is initialized in such a way. For example, ``NLTV2++'' means nonlocal total variation, quadratic model with ``K-means++'' initialization procedure.

The algorithms are compared on the following datasets:
\begin{enumerate}
\item \textbf{Synthetic Dataset:} This dataset\footnote{Available at \url{http://www.math.ucla.edu/~weizhu731/}} contains five endmembers and $162$ spectral bands. The 40,000 abundance vectors were generated as a sum of Gaussian fields. The dataset was generated using a Generalized Bilinear Mixing Model (GBM):
\begin{equation}
\label{synequ}\nonumber
y=\sum_{i=1}^p a_i e_i + \sum_{i=1}^{p-1}\sum_{j=i+1}^p \gamma_{ij}a_ia_je_i\odot e_j + n,
\end{equation}
where $\gamma_{ij}$ are chosen uniformly and randomly in the interval $[0,1]$, $n$ is the Gaussian noise, with an SNR of 30 dB, and $a_i$ satisfies: $a_i \ge 0$, and $\sum_{i=1}^p a_i = 1$. 
\item \textbf{Salinas-A Dataset:} Salinas-A scene\footnote{\label{site}Available at \url{http://www.ehu.eus/ccwintco/index.php?title=Hyperspectral\_Remote\_Sensing\_Scenes}} was a small subscene of Salinas image, which was acquired by the AVIRIS sensor over Salinas Valley. It contains $86 \times 83$ pixels and $204$ bands. The ground truth includes six classes: broccoli, corn, and four types of lettuce.
\item \textbf{Urban Dataset:} The Urban dataset\footnote{Available at \url{http://www.agc.army.mil/.}} is from HYperspectral Digital Imagery Collection Experiment (HYDICE), which has $307\times 307$ pixels and contains $162$ clean spectral bands. This dataset only has six classes of material: road, dirt, house, metal, tree, and grass.
\item \textbf{San Diego Airport Dataset:} The San Diego Airport (SDA) dataset\footnote{Available at \url{http://www.math.ucla.edu/~weizhu731/}} is provided by the HYDICE sensor. It comprises $400 \times 400$ pixels and contains $158$ clean spectral bands. There are seven types of material: trees, grass, three types of road surfaces, and two types of rooftops \cite{kuang}. The RGB image with cluster labels are shown in Fig. \ref{SD_comparepic}.
\item \textbf{Chemical Plume Dataset:} The chemical plume dataset\footnote{Available at \url{http://www.math.ucla.edu/~weizhu731/}} consists of frames taken from a hyperspectral video of the release of chemical plumes provided by the John Hopkins University Applied Physics Laboratory. The image has $128 \times 320$ pixels, with $129$ clean spectral bands. There was no ground truth provided for this data, so a segmentation of four classes is assumed: chemical plume, sky, foreground, and mountain. A fifth cluster is added so that the noise pixels would not interfere with the segmentation \cite{huiyi_plume}.
\item \textbf{Pavia University Dataset:} The Pavia University dataset is collected by the ROSIS sensor. It contains $103$ clean spectral bands and $610 \times 340$ pixels, and comprises $9$ classes of material.
\item \textbf{Indian Pines Dataset:} The Indian Pines dataset was acquired by AVIRIS sensor and consists of $145 \times 145$ pixels, with 200 clean spectral bands. The available ground truth is labeled into 16 classes.
\item \textbf{Kennedy Space Center Dataset:} This dataset was gathered by the NASA AVIRIS sensor over the Kennedy Space Center, Florida. A subscene of the western shore of the center is used in the numerical experiment. $12$ classes of different materials are reported in the datacube of size $512 \times 365 \times 176$.
\end{enumerate}

K-means and NMF are non-parametric, and the parameter setups of H2NMF and the MBO scheme are described in \cite{kuang} and \cite{huiyi_plume,ekaterina_plume}. The key parameters $\lambda$ and $\mu$ in the NLTV models are determined in the following way:
\begin{enumerate}
\item $\lambda$ is chosen such that the data fidelity term is around $10$ times larger than the NLTV regularizing term $\|\nabla_w u\|_{L^1}$.
\item $\mu$ is chosen such that the Euclidean distances between different endmembers are roughly $10$ times smaller than the cosine distances.
\end{enumerate}

\begin{table}[!t]
\renewcommand{\arraystretch}{1.3}
\caption{key parameters used for different datasets}
\label{tab:parameter}
\centering
\begin{tabular}{|c|c|c|c|c|c|c|c|}
\hline
\bfseries Datasets &\bfseries  $\lambda$ & \bfseries $\mu$ & \bfseries Datasets & \bfseries $\lambda$ & \bfseries $\mu$\\
\hline
\bfseries Synthetic & $10^{-1}$ & $10^{-4}$ &\bfseries Plume & $10^7$& $10^{-2}$\\
\hline
\bfseries Urban & $10^6$ & $10^{-5}$ & \bfseries Pavia & $10^6$ & $10^{-8}$\\
\hline
\bfseries Salinas-A & $10^4$ & $10^{-4}$ & \bfseries Pines & $10^6$ & $10^{-9}$\\
\hline
\bfseries SDA & $10^6$ & $10^{-7}$ & \bfseries KSC & $10^6$ & $10^{-8}$\\
\hline
\end{tabular}
\end{table}

Table \ref{tab:parameter} displays the parameters chosen for the numerical experiments. The large variance of the parameter scales results from the variety of image sizes and scales. A  sensitivity analysis over the parameters is presented in Section \ref{sec:sensitivity}.

\subsection{Synthetic Dataset and Salinas-A Dataset}
\begin{figure*}[!t]
  \centering
  \begin{tabular}{c}
    Ground Truth\\
    \includegraphics[width=2.4in]{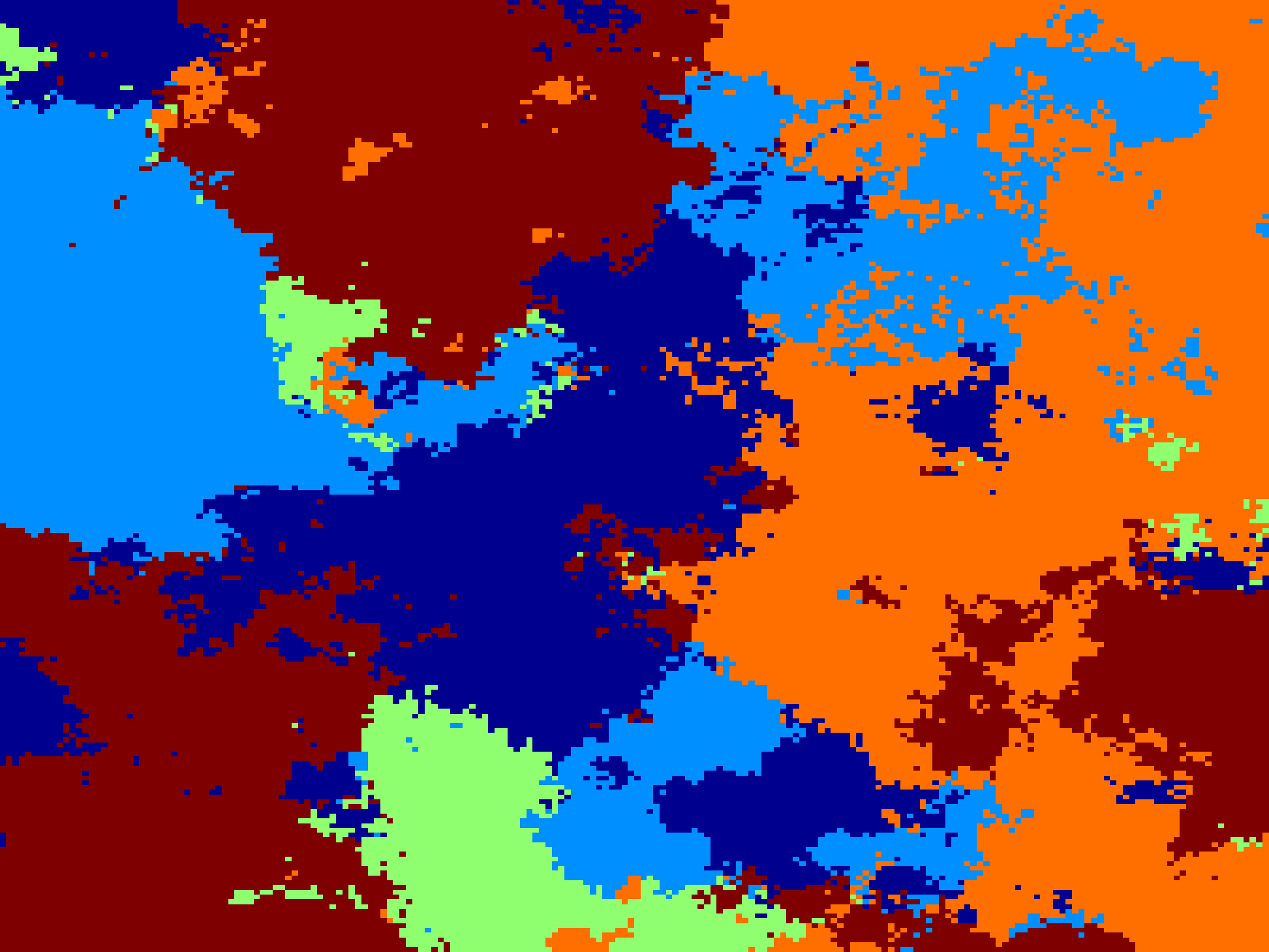}
  \end{tabular}
  \begin{tabular}{ccc}
    K-means++ & NMF++ & H2NMF \\ 
    \includegraphics[width=1.3in]{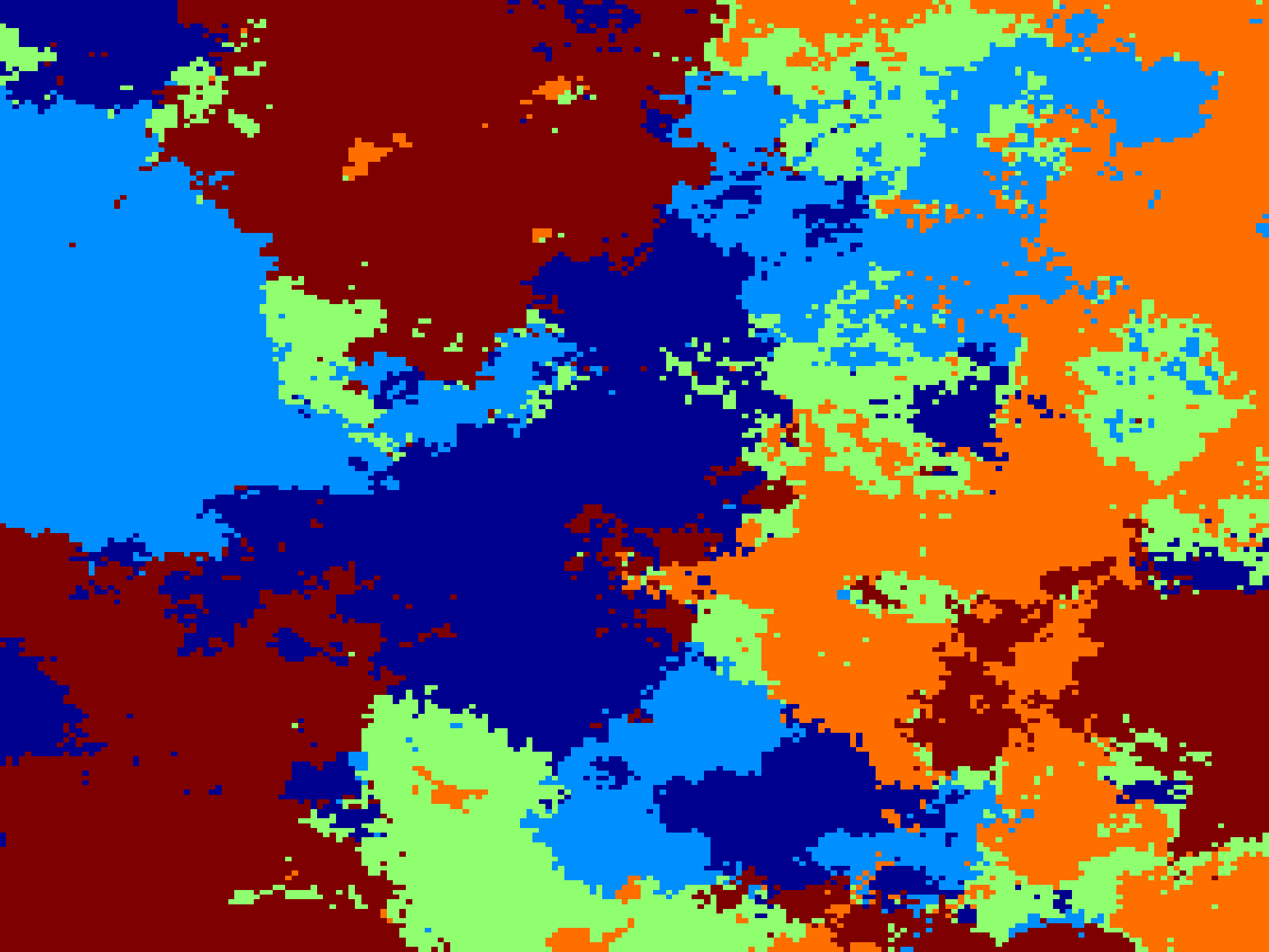}&
    \includegraphics[width=1.3in]{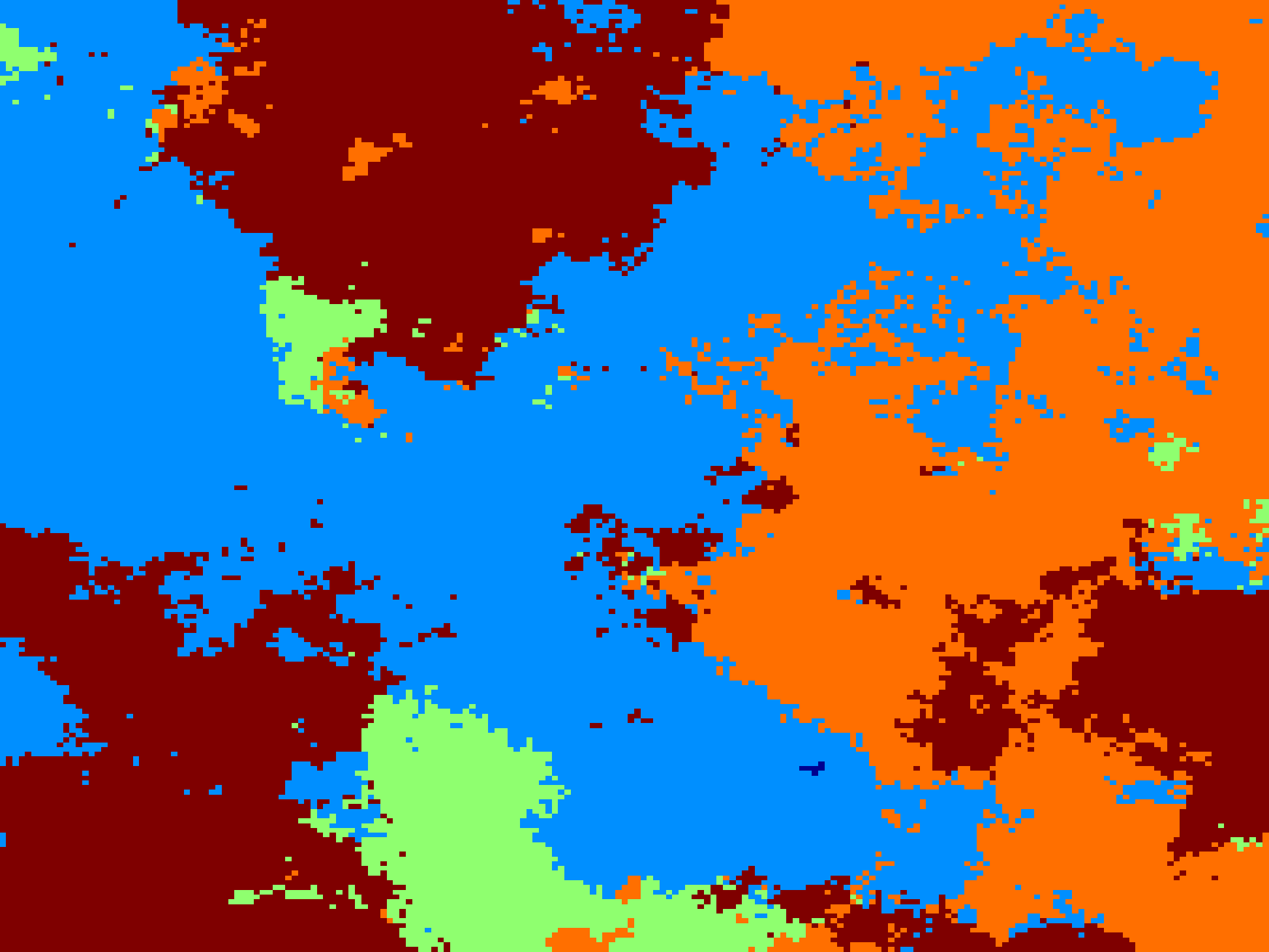}&
    \includegraphics[width=1.3in]{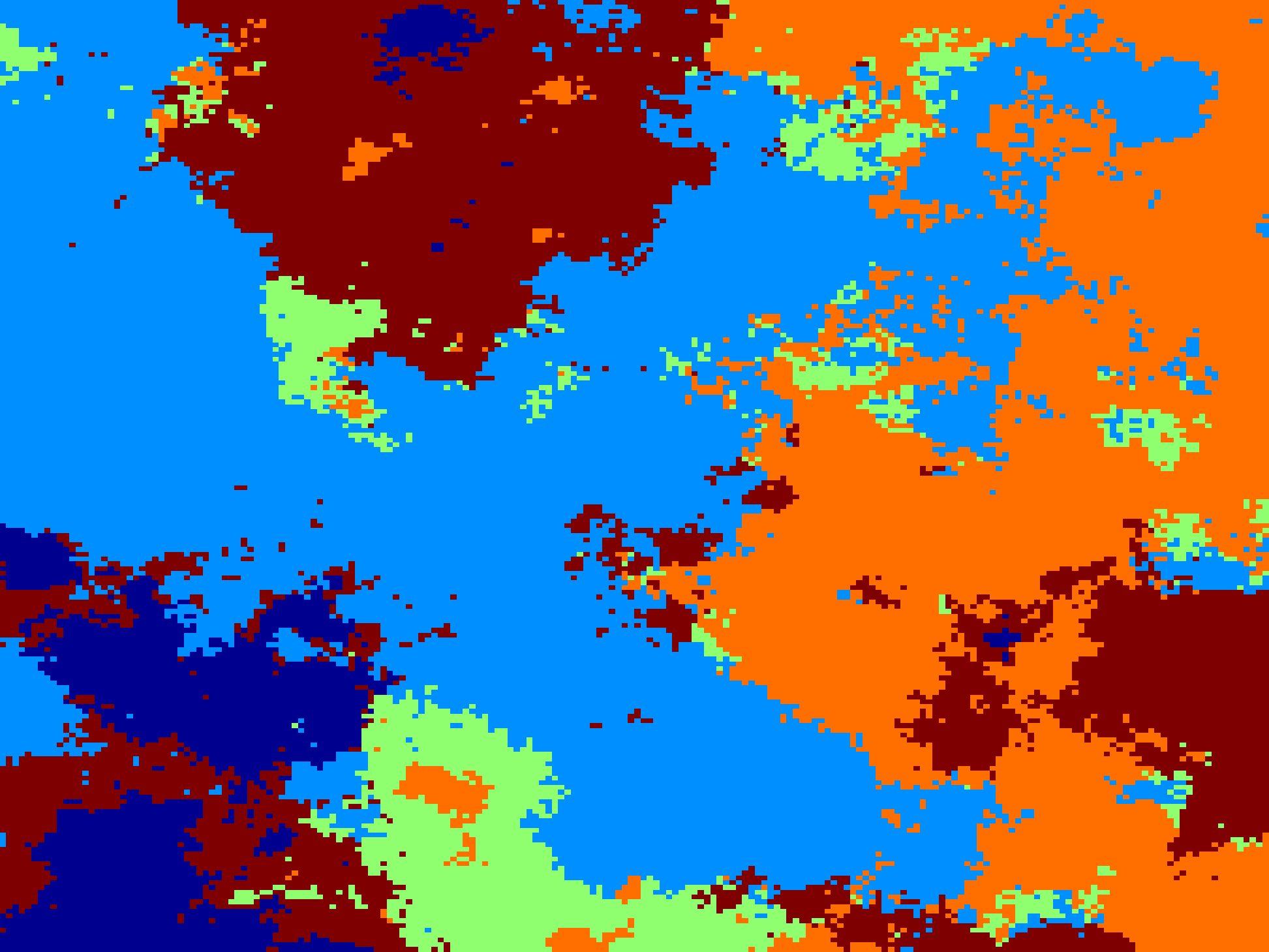}\\
    MBO++ & NLTV2++ & NLTV1(K-means) \\ 
    \includegraphics[width=1.3in]{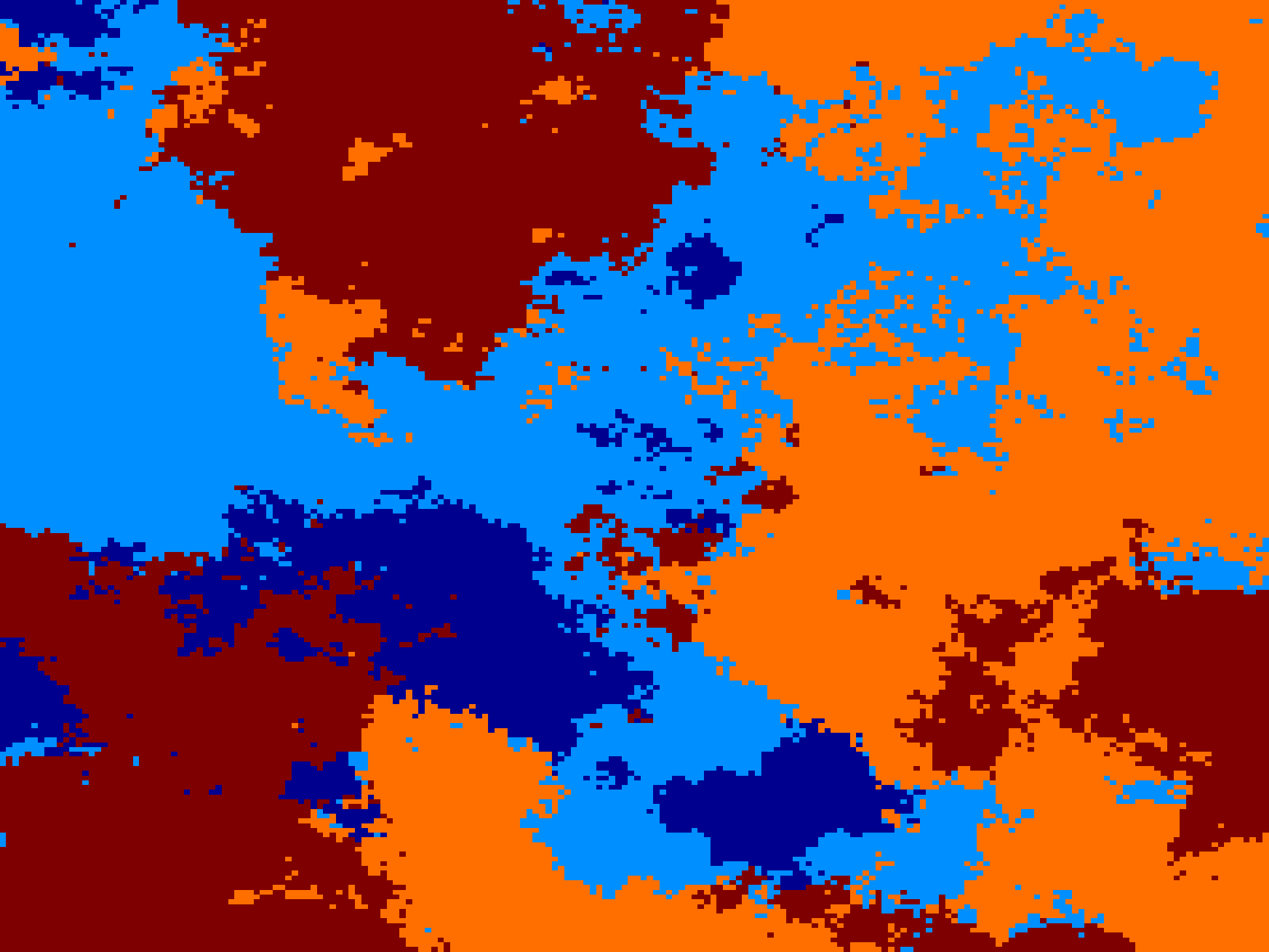}&
    \includegraphics[width=1.3in]{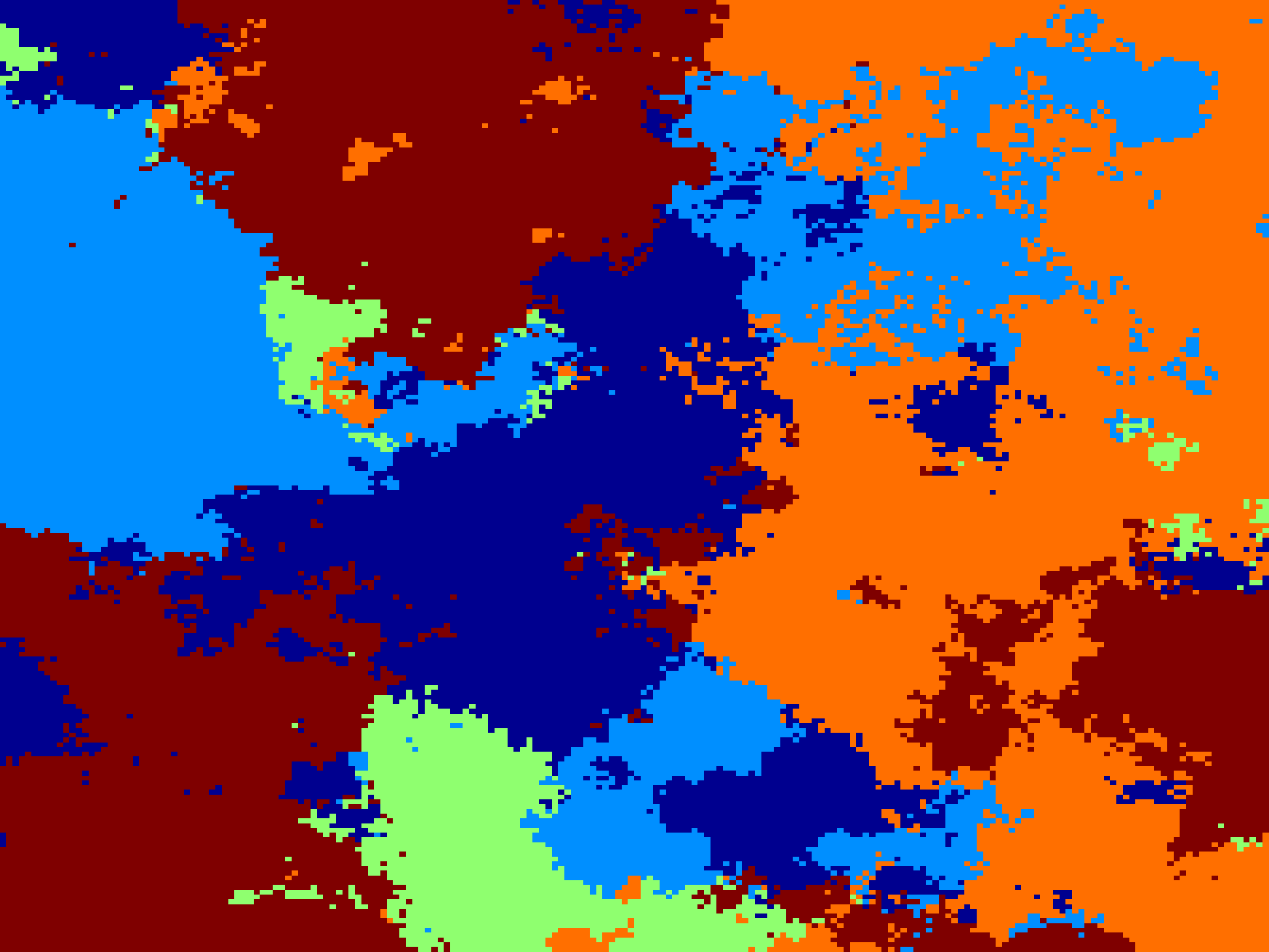}&
    \includegraphics[width=1.3in]{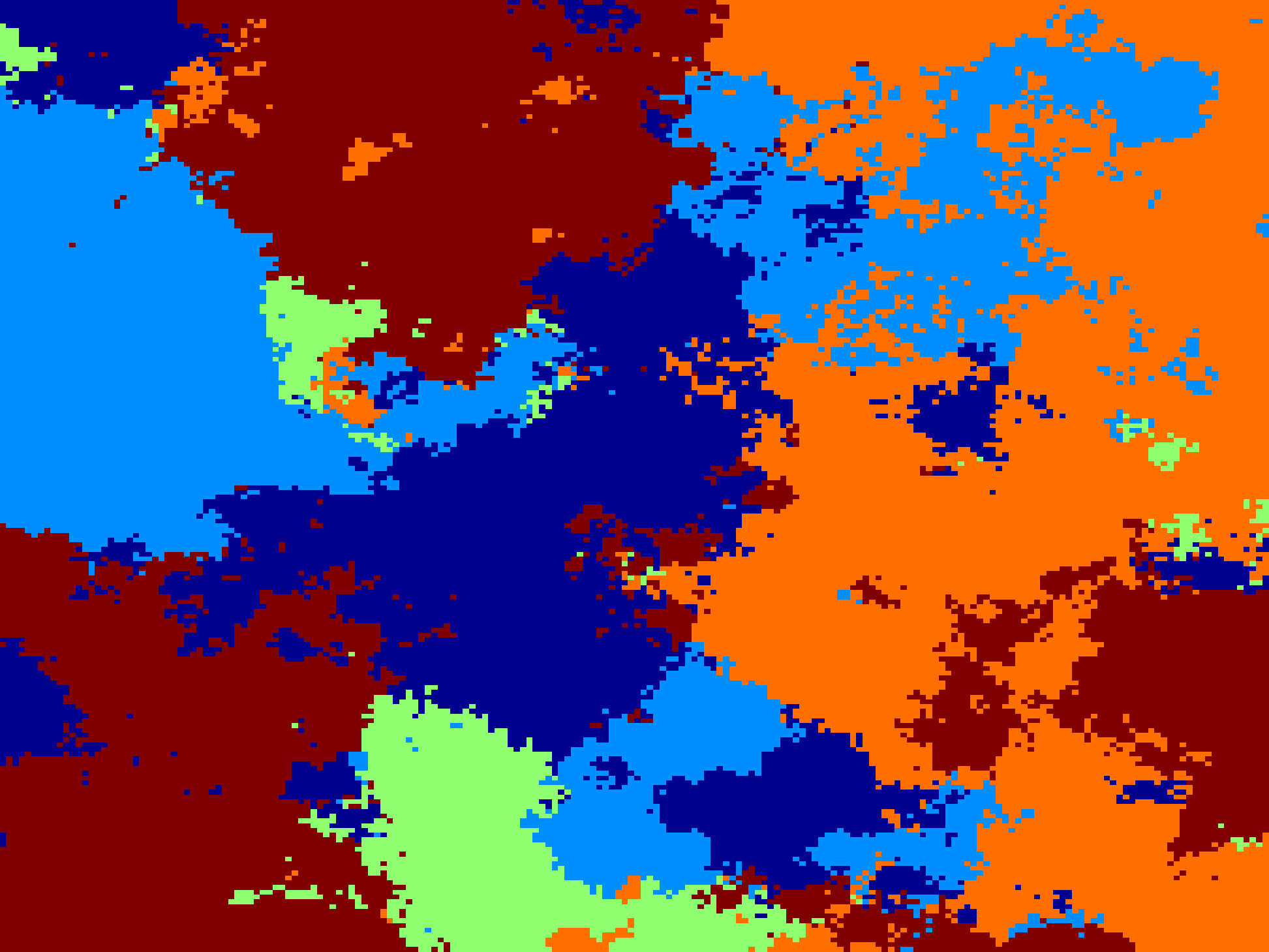}
  \end{tabular}
  \caption{Clustering results for the synthetic dataset generated by 5 endmembers. The first image on the left is the ground truth, and the remaining six images  are the clustering results of the corresponding algorithms.}
  \label{synpic}
\end{figure*}

All the algorithms are first tested on the synthetic dataset. The classification results are shown in Table \ref{syn_salinas_tab} and Fig. \ref{synpic}.  Both  NLTV algorithms have better overall accuracy than all of the other methods, although they took a longer time to converge. The qudratic model classified the image almost perfectly.

\begin{figure*}[!t]
  \centering
  \begin{tabular}{c}
    Ground Truth\\
    \includegraphics[width=2.4in]{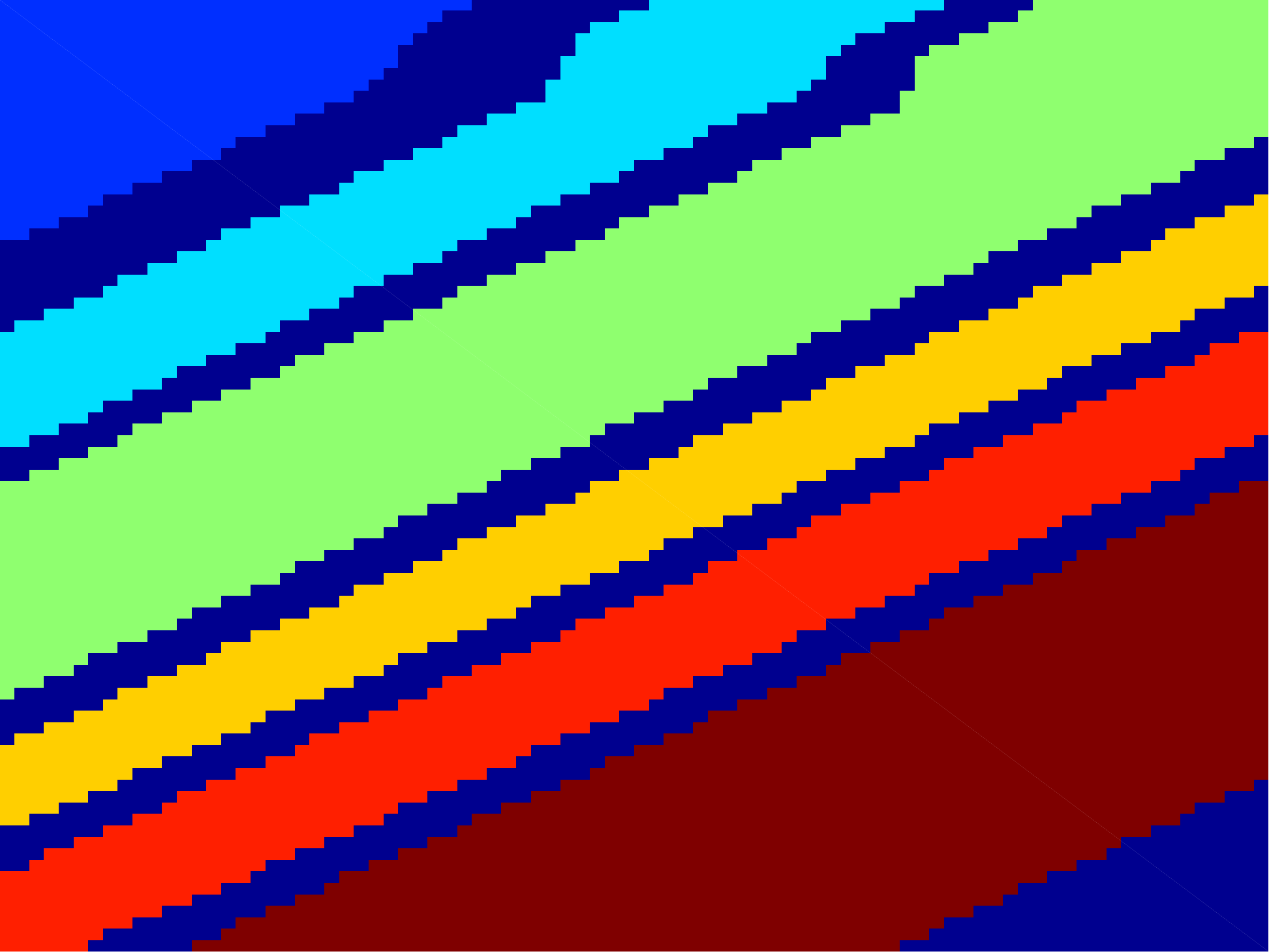}
  \end{tabular}
  \begin{tabular}{ccc}
    K-means++ & NMF++ & H2NMF \\ 
    \includegraphics[width=1.3in]{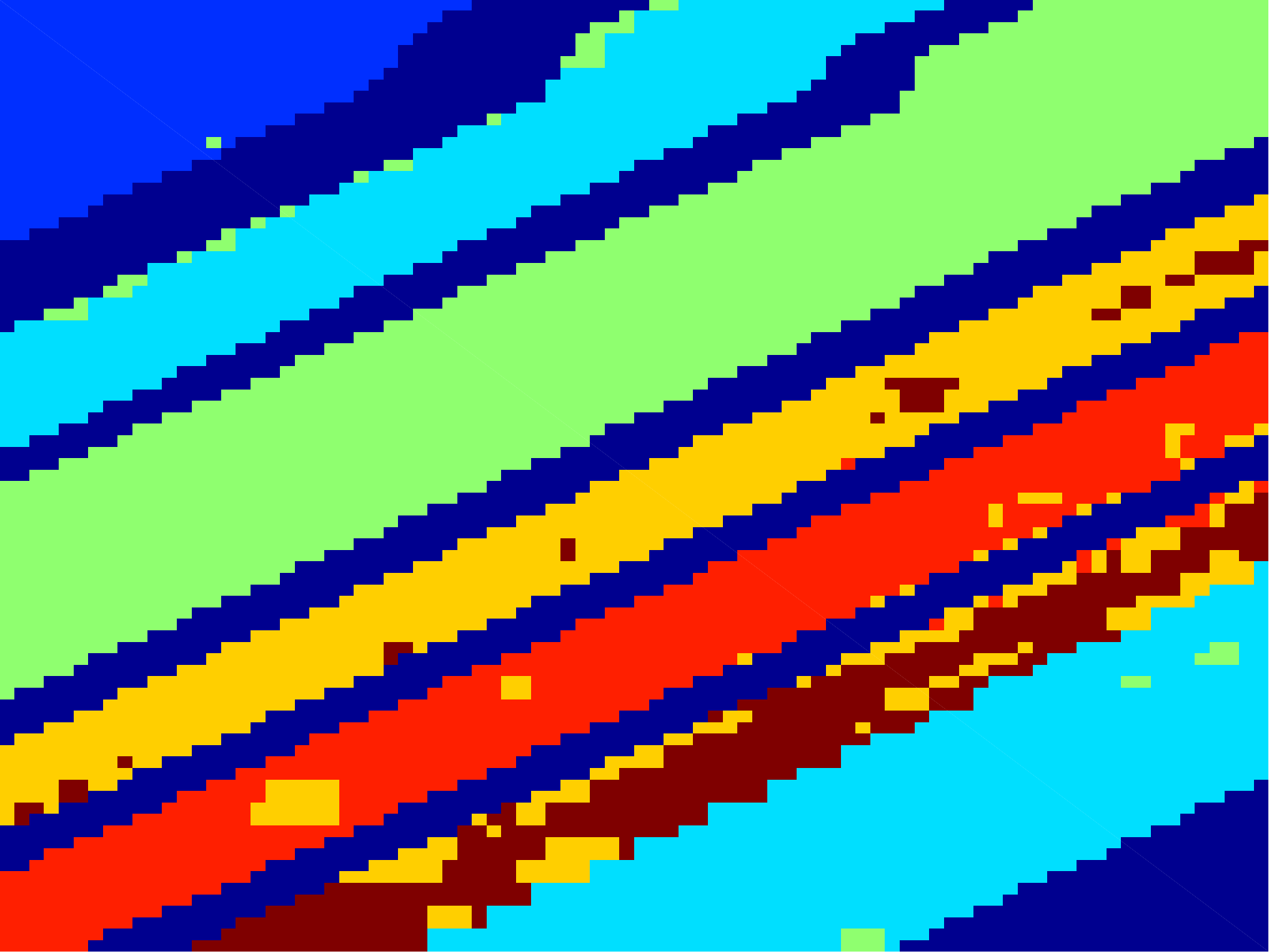}&
    \includegraphics[width=1.3in]{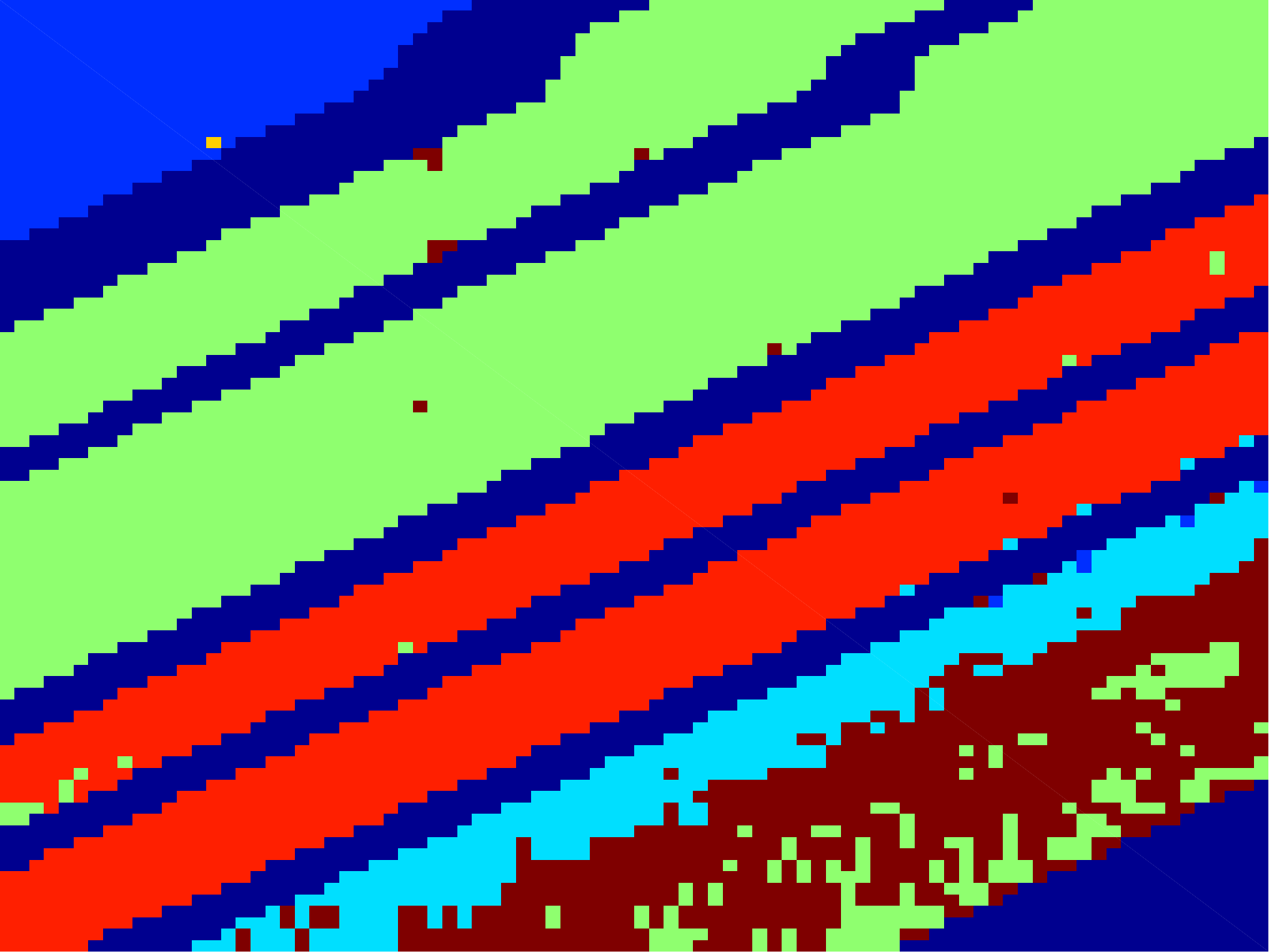}&
    \includegraphics[width=1.3in]{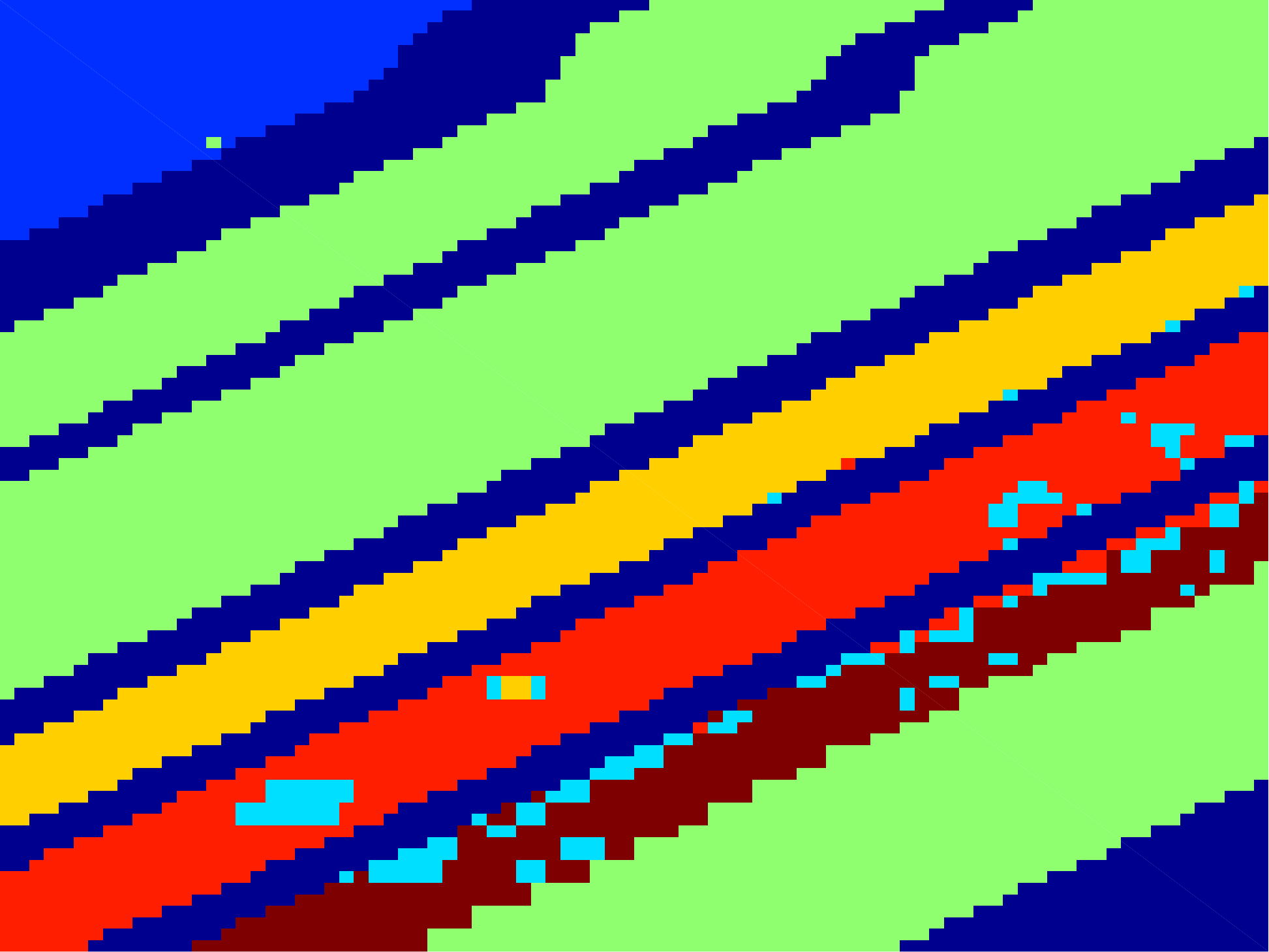}\\
    MBO++ & NLTV2++ & NLTV1(K-means) \\ 
    \includegraphics[width=1.3in]{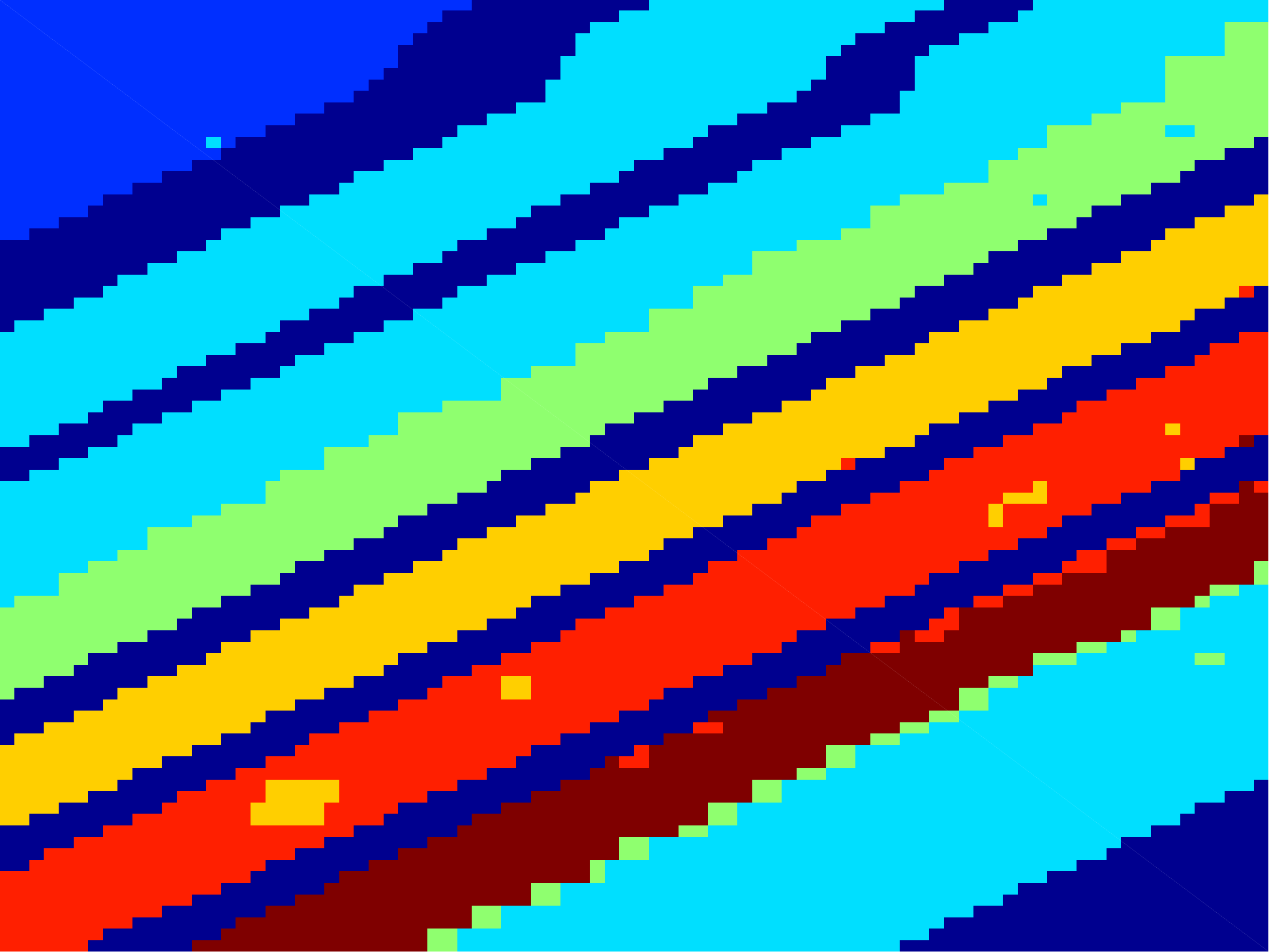}&
    \includegraphics[width=1.3in]{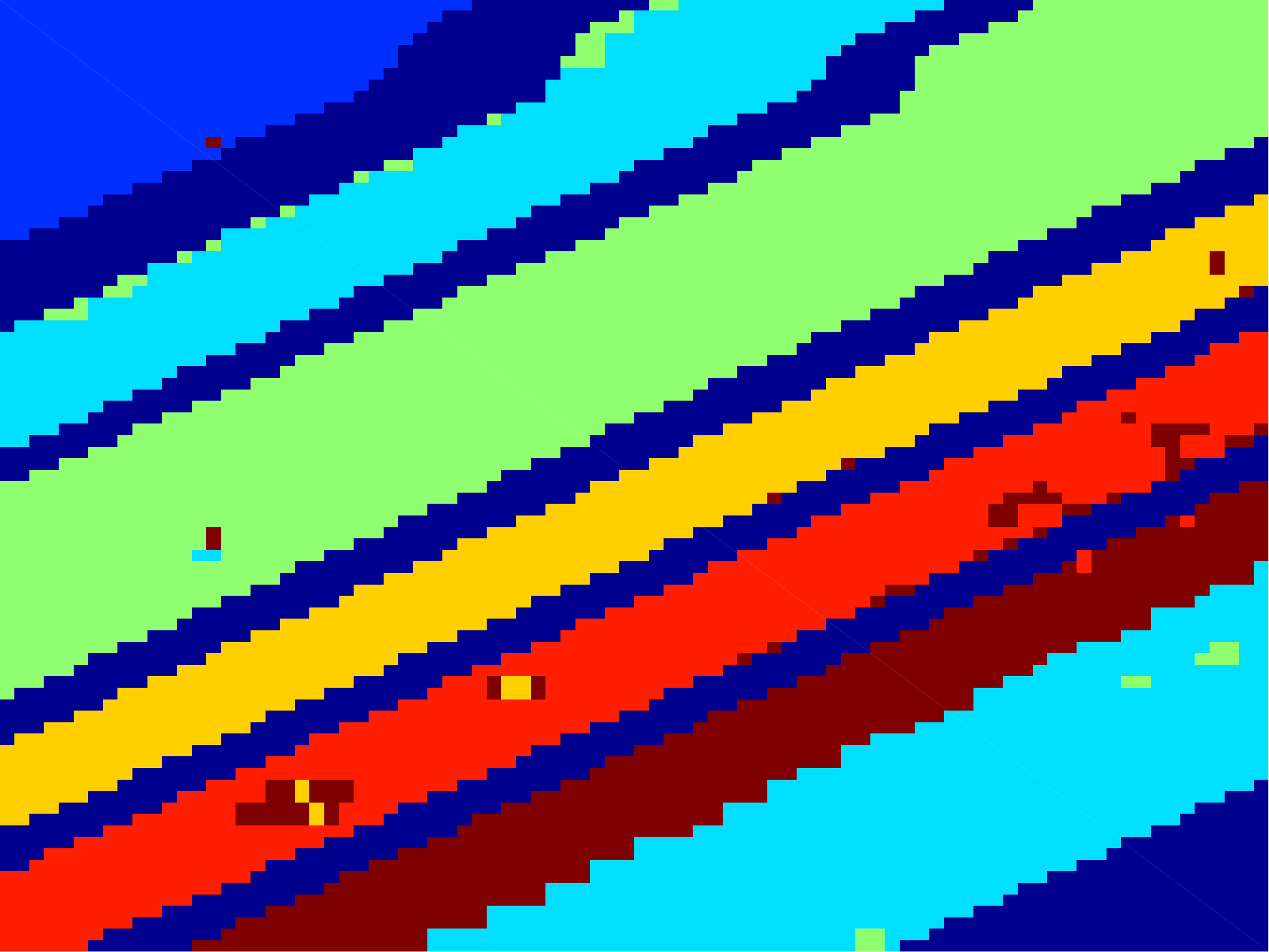}&
    \includegraphics[width=1.3in]{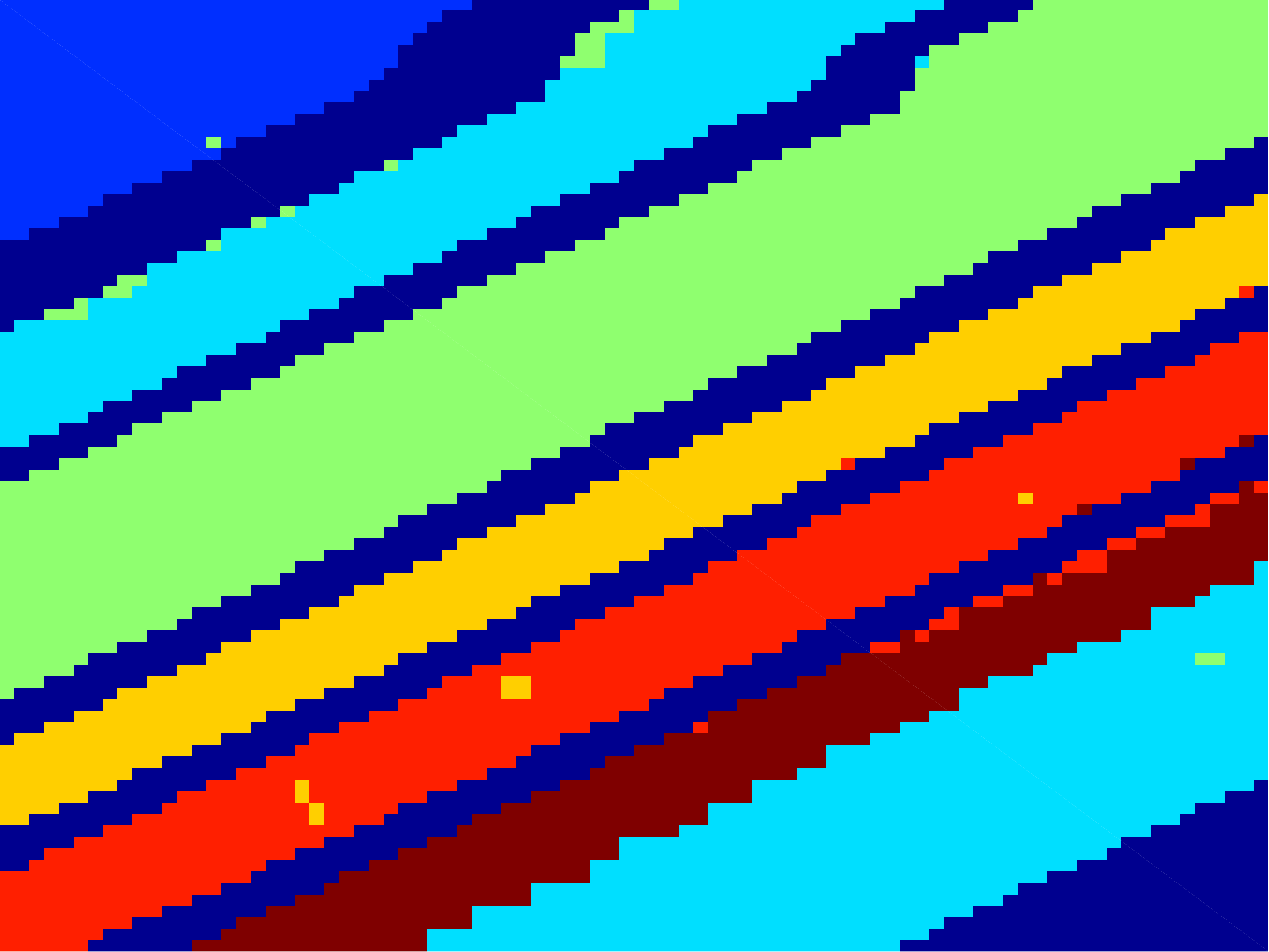}
  \end{tabular}
  \caption{Clustering results for the Salina-A dataset. The first image on the left is the ground truth, and the remaining six images are the clustering results of the corresponding algorithms.}
  \label{salinaspic}
\end{figure*}

The visual classification results and overall accuracies of the Salinas-A dataset are shown in Fig. \ref{salinaspic} and Table \ref{syn_salinas_tab}. Both NLTV methods performed at higher accuracy compared to other methods. The linear model improved the result of K-means by incorporating spatial information of the dataset, and the quadratic model only took 4 iterations to converge.

\begin{table}[!t]
  \renewcommand{\arraystretch}{1.3}
  \caption{Comparison Of Numerical Results on the Synthetic and Salinas-A Datasets}
  \label{syn_salinas_tab}
  \centering
  \begin{tabular}{|c|c|c|c|c|}
    \hline
    \multirow{2}{*}{\bfseries Algorithm}& \multicolumn{2}{c|}{\bfseries Synthetic}&\multicolumn{2}{c|}
                                                                {\bfseries Salinas-A}\\ 
    \cline{2-5}
                              & \bfseries Run-Time & \bfseries Accuracy & \bfseries Run-Time & \bfseries Accuracy\\ \hline
    \bfseries K-means++&2s&90.98\%&0.9s&79.92\%\\ \hline
    \bfseries NMF++&9s&80.99\%&1.0s&64.47\%\\ \hline
    \bfseries H2NMF&2s&72.02\%&1.5s&70.08\%\\ \hline
    \bfseries MBO++ & 21s & 84.49\% & 7.8s & 68.62\% \\ \hline
    \bfseries NLTV2++ & 29s & 99.93\% & 1.6s & 83.69\%\\ \hline
    \bfseries NLTV1(K-means) & 29s & 95.96\% & 3.4s & 83.75\%\\ \hline
  \end{tabular}
\end{table}

\subsection{Urban DataSet}

\begin{table}[!t]
\renewcommand{\arraystretch}{1.3}
\caption{Comparison of Numerical Results on the Urban Dataset}
\label{urban_table}
\centering
\begin{tabular}{|c|c|c|}
\hline
\bfseries Algorithm & \bfseries Run-Time & \bfseries Accuracy\\
\hline
\bfseries K-means & 7s & 75.20\%\\
\hline
\bfseries NMF &87s & 55.70\%\\
\hline
\bfseries H2NMF & 7s & 85.96\%\\
\hline
\bfseries MBO & 92s & 78.86\%\\
\hline
\bfseries NLTV2 & 96s  & 92.14\%\\
\hline
\bfseries NLTV1(H2NMF) & 17s & 91.56\%\\
\hline
\end{tabular}
\end{table}

\begin{figure*}[!t]
  \centering
  \begin{tabular}{c}
    Ground Truth\\
    \includegraphics[width=2.4in]{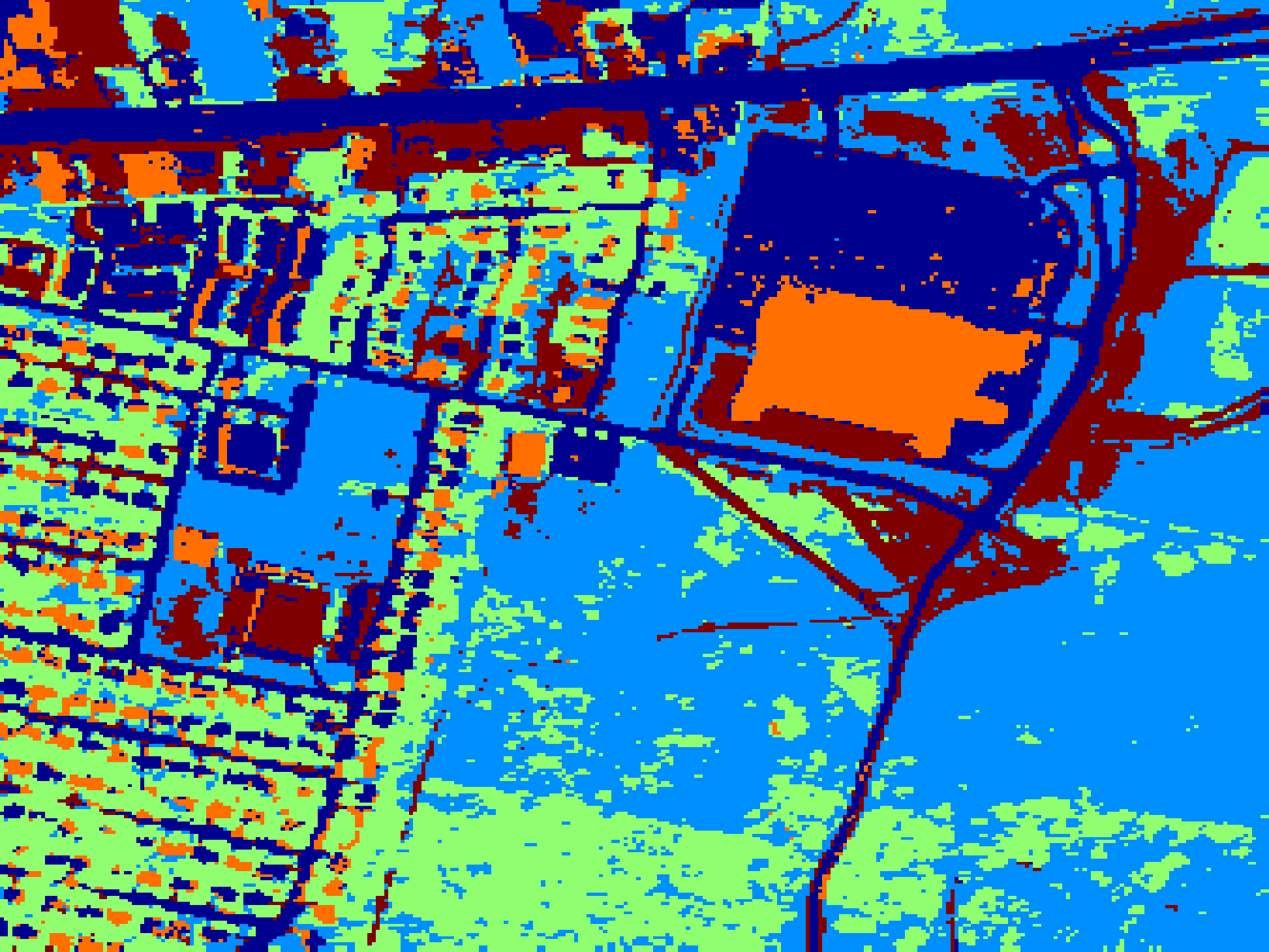}
  \end{tabular}
  \begin{tabular}{ccc}
    K-means & NMF & H2NMF \\ 
    \includegraphics[width=1.3in]{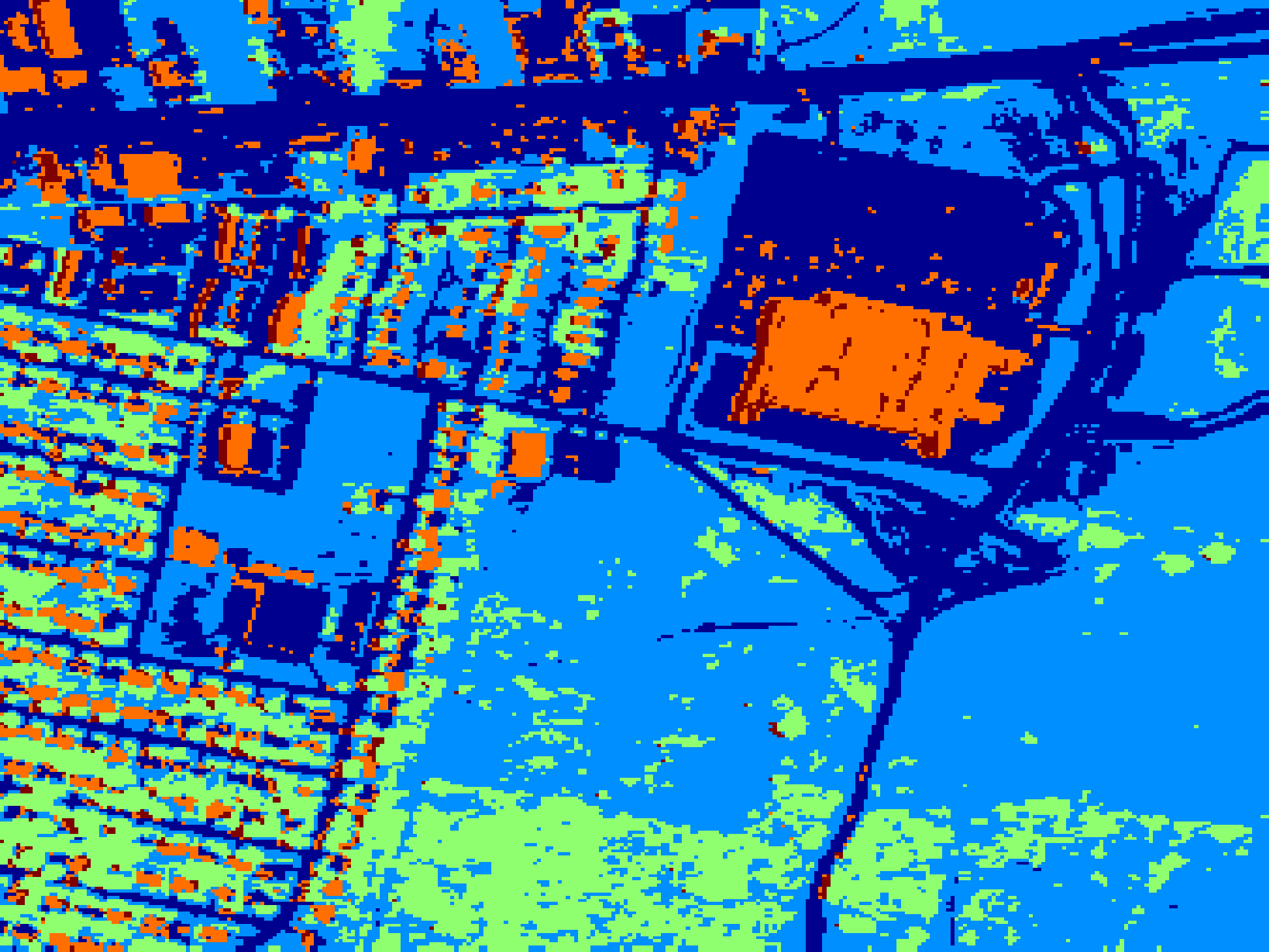}&
    \includegraphics[width=1.3in]{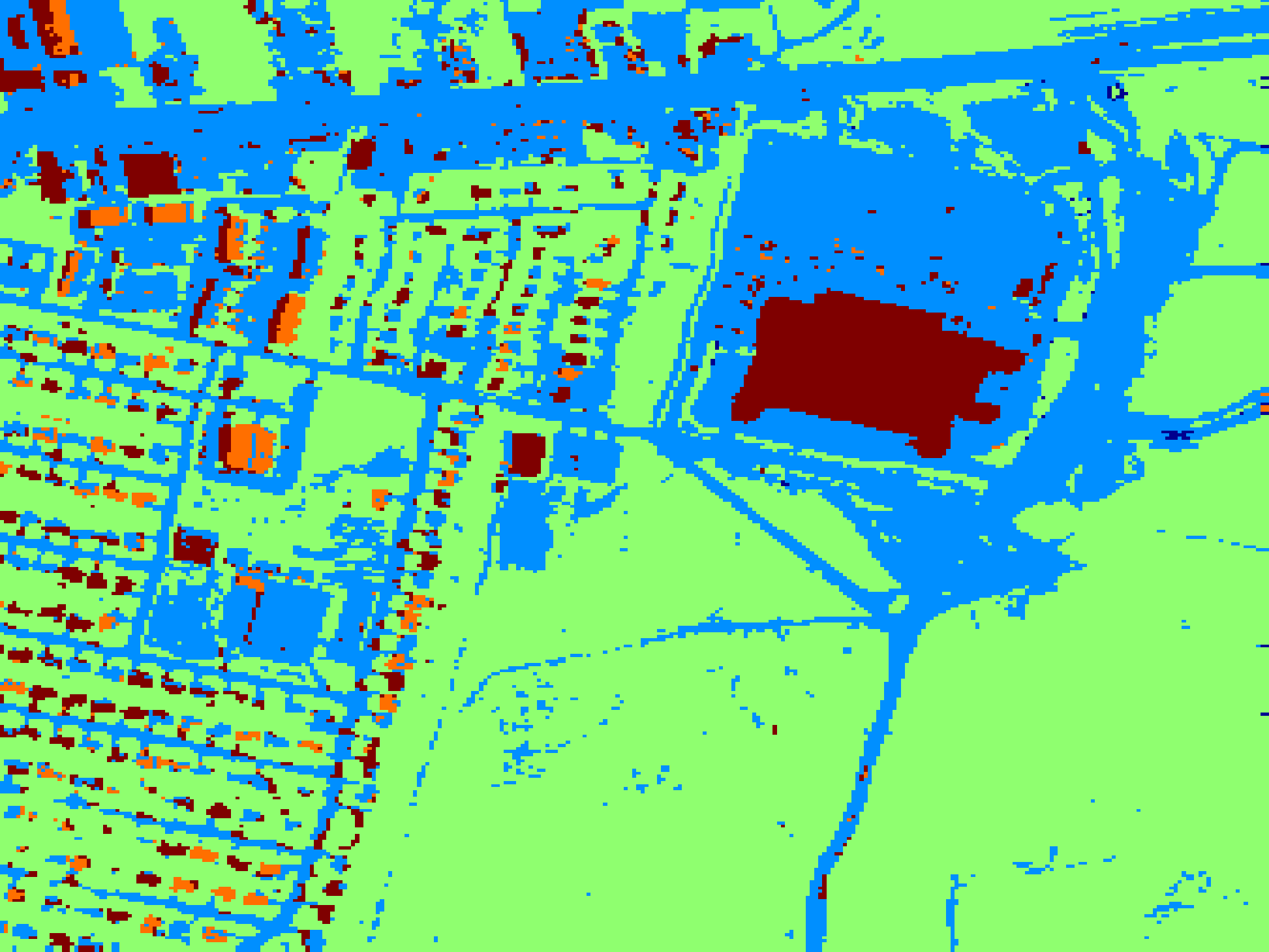}&
    \includegraphics[width=1.3in]{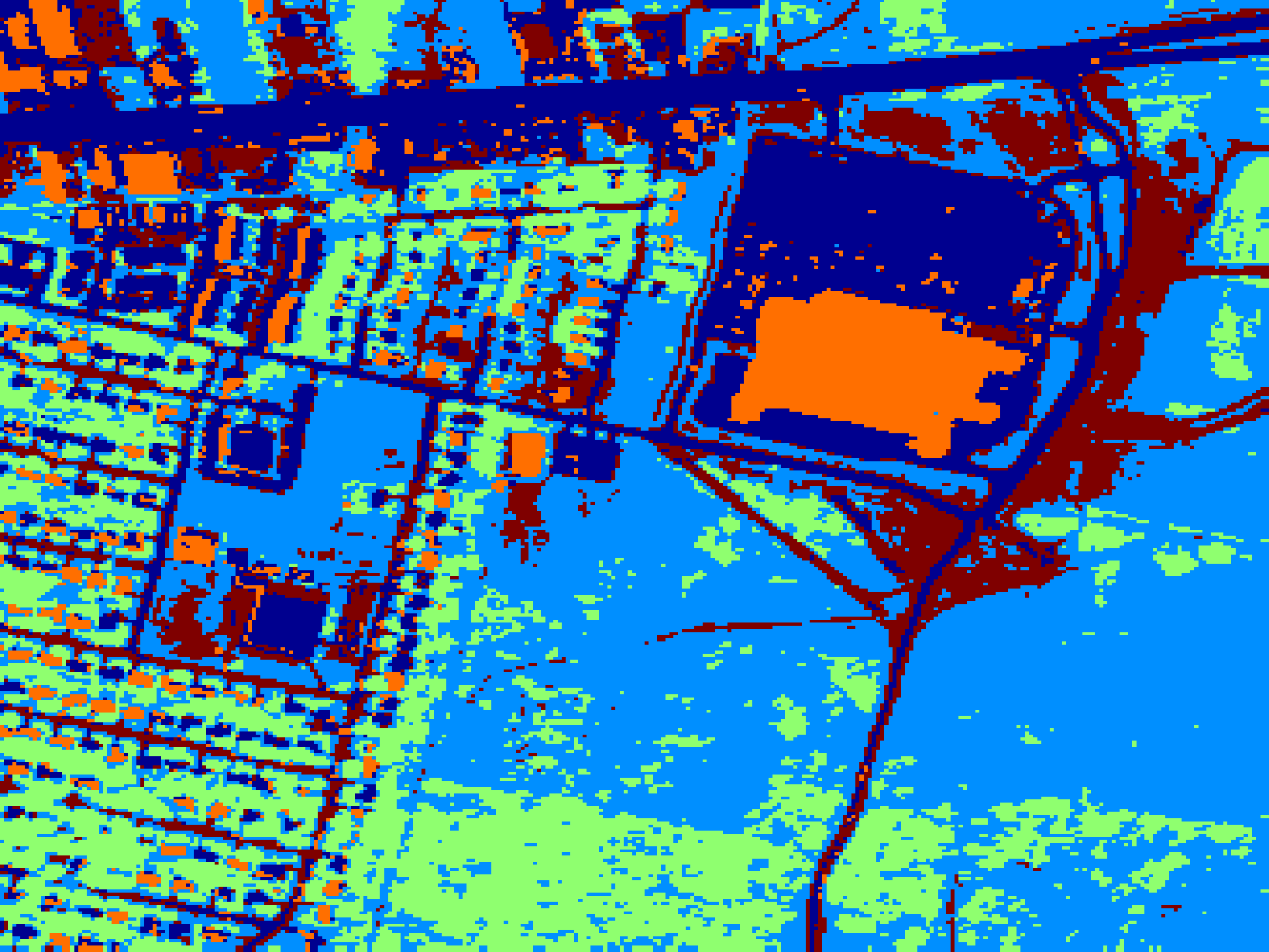}\\
    MBO & NLTV2 & NLTV1(H2NMF) \\ 
    \includegraphics[width=1.3in]{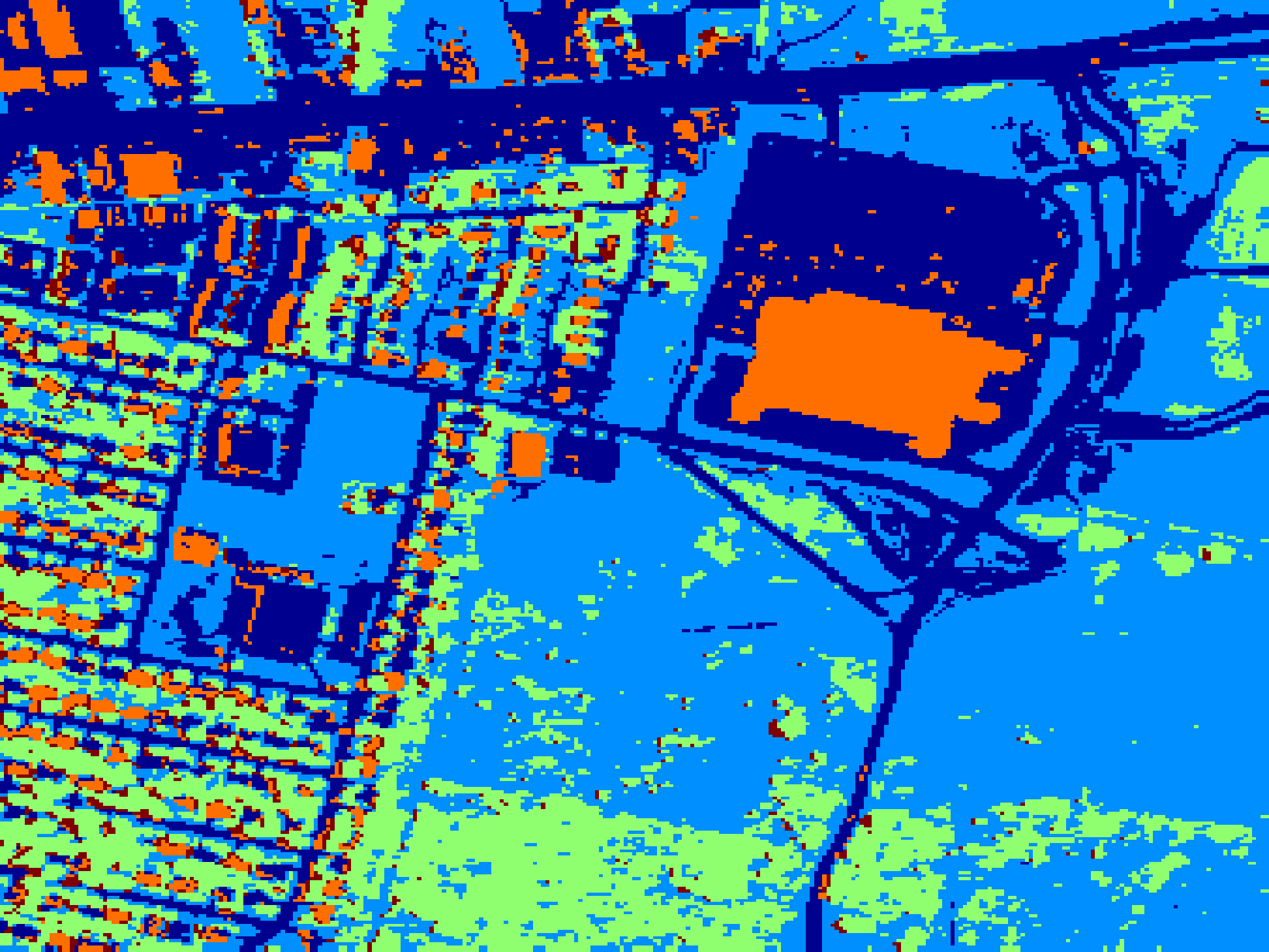}&
    \includegraphics[width=1.3in]{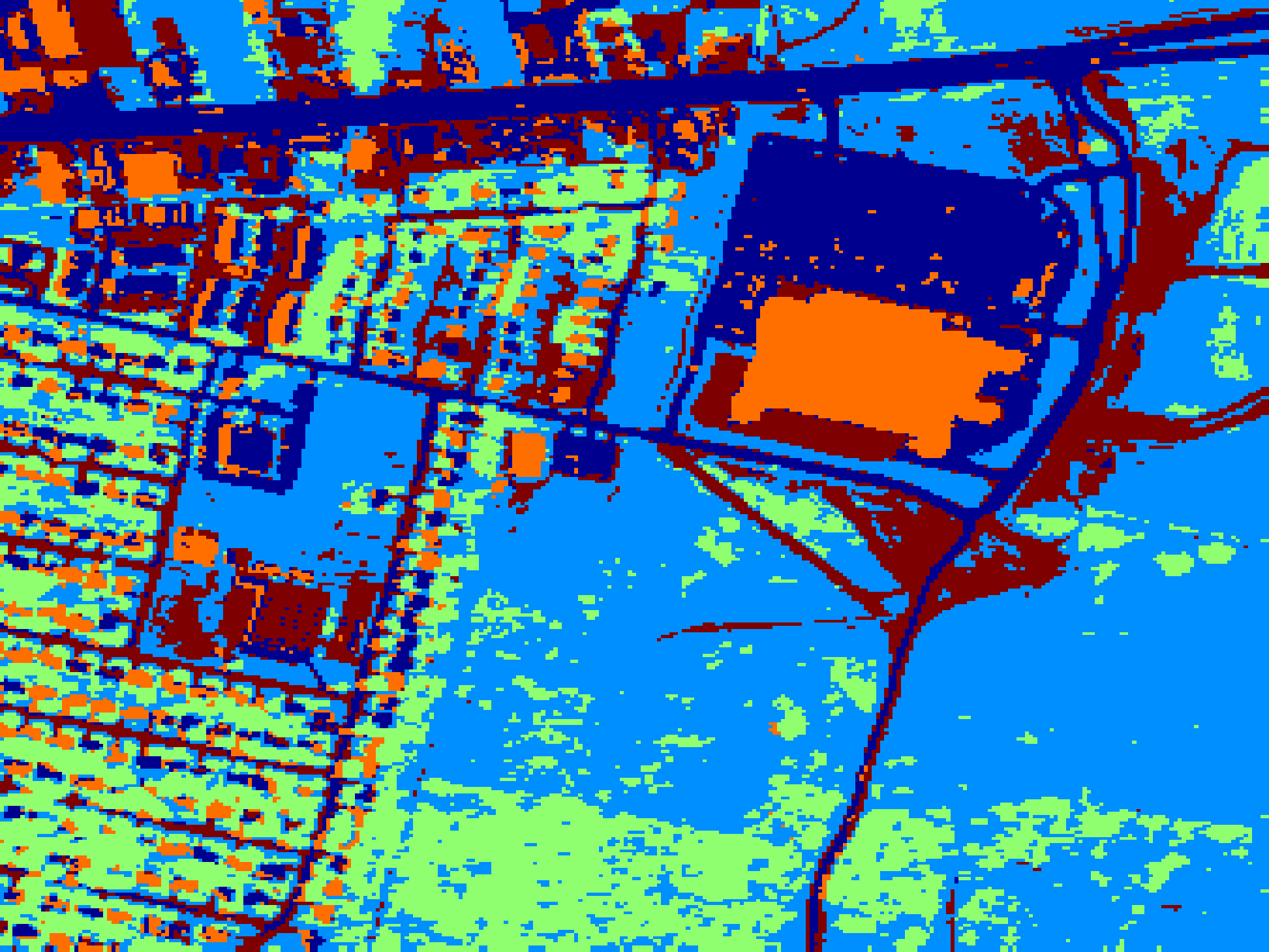}&
    \includegraphics[width=1.3in]{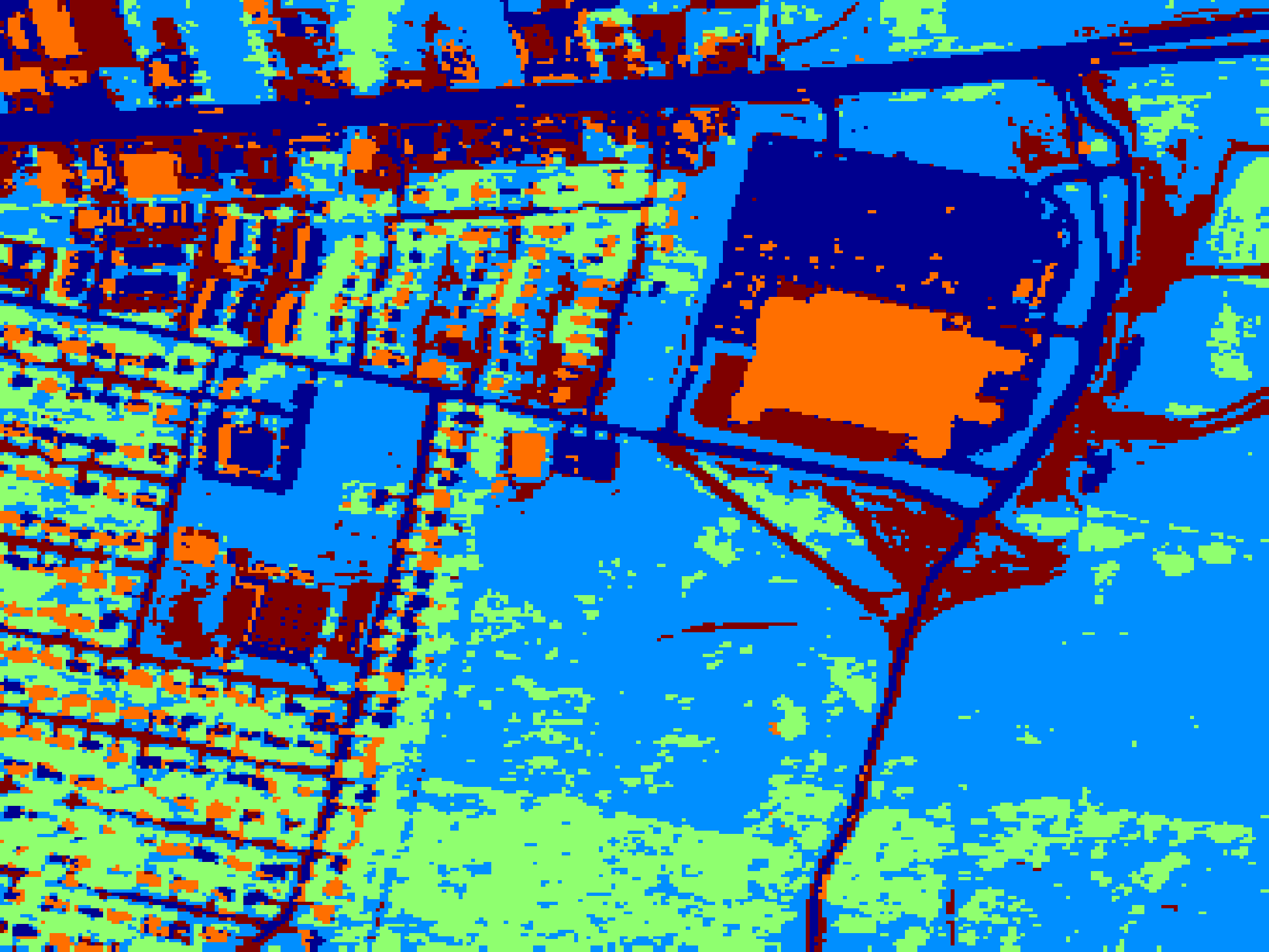}
  \end{tabular}
  \caption{Clustering results for the Urban dataset. Five clusters including rooftops, grass, trees, dirt, and ``road+metal" are generated by the algorithms.}\label{urban_comparepic}
  \label{urban_comparepic}
\end{figure*}

There was no ground-truth provided for the Urban HSI. A structured sparse algorithm \cite{urbangt} (which is different from all of the testing algorithms) has been used to initialize a ground truth, which is then corrected pixel by pixel to provide a framework for numerical analysis of accuracy. As this ``ground truth" was hand-corrected, it does not necessarily represent the most accurate segmentation of the image; however, it provides a basis for quantitative comparison.

After running all the algorithms that are compared to create six clusters, we noticed that they all split ``grass" into two different clusters (one of them corresponds to a mixture of grass and dirt), while treating ``road" and ``metal" as the same. To obtain a reliable overall accuracy of the classification results, the two ``grass" clusters are combined in every algorithm, hence obtaining the classification results for 5 clusters, which are ``grass", ``dirt", ``road+metal", ``roof", and ``tree".

The overall classification accuracies and run-times are displayed in Table \ref{urban_table}. As can be seen, the proposed NLTV algorithms performed consistently better with comparable run-time. It is easier to see visually in Fig. \ref{urban_comparepic} that the NLTV algorithm performed best of the five algorithms tested; specifically, the NLTV algorithm alone distinguished all of the dirt beneath the parking lot and the intricacies of the road around the parking lot. The total variation regularizer also gives the segmented image smoother and more distinct edges, allowing easier human identification of the clusters.
\subsection{San Diego Airport Dataset}
\begin{table}[!t]
\renewcommand{\arraystretch}{1.3}
\caption{Run-Times for the San Diego Airport (SDA), Chemical Plume (Plume),  Pavia University (Pavia), Indian Pines (Pines), and Kennedy Space Center (KSC) Datasets}
\label{runtime}
\centering
\begin{tabular}{|c|d|d|d|d|d|}
\hline
\bfseries Algorithm & \bfseries SDA  & \bfseries Plume & \bfseries Pavia & \bfseries Pines & \bfseries KSC  \\
\hline
\bfseries K-means &9s  & 2s & 26s & 10s& 47s\\
\hline
\bfseries NMF & 4s  & 2s & 120s & 19s& 135s\\
\hline
\bfseries H2NMF & 13s  & 2s & 12s & 4s& 24s\\
\hline
\bfseries MBO & 329s  & 18s & 1020s & 198s& 754s\\
\hline
\bfseries NLTV2 & 43s  & 23s & 299s & 64s& 561s\\
\hline
\bfseries NLTV1(H2NMF) & 17s  & 18s & 132s & 21s & 188s\\
\hline
\end{tabular}
\end{table}

\begin{figure*}[!t]
  \centering
  \begin{tabular}{c}
    RGB Image\\
    \includegraphics[width=2.4in]{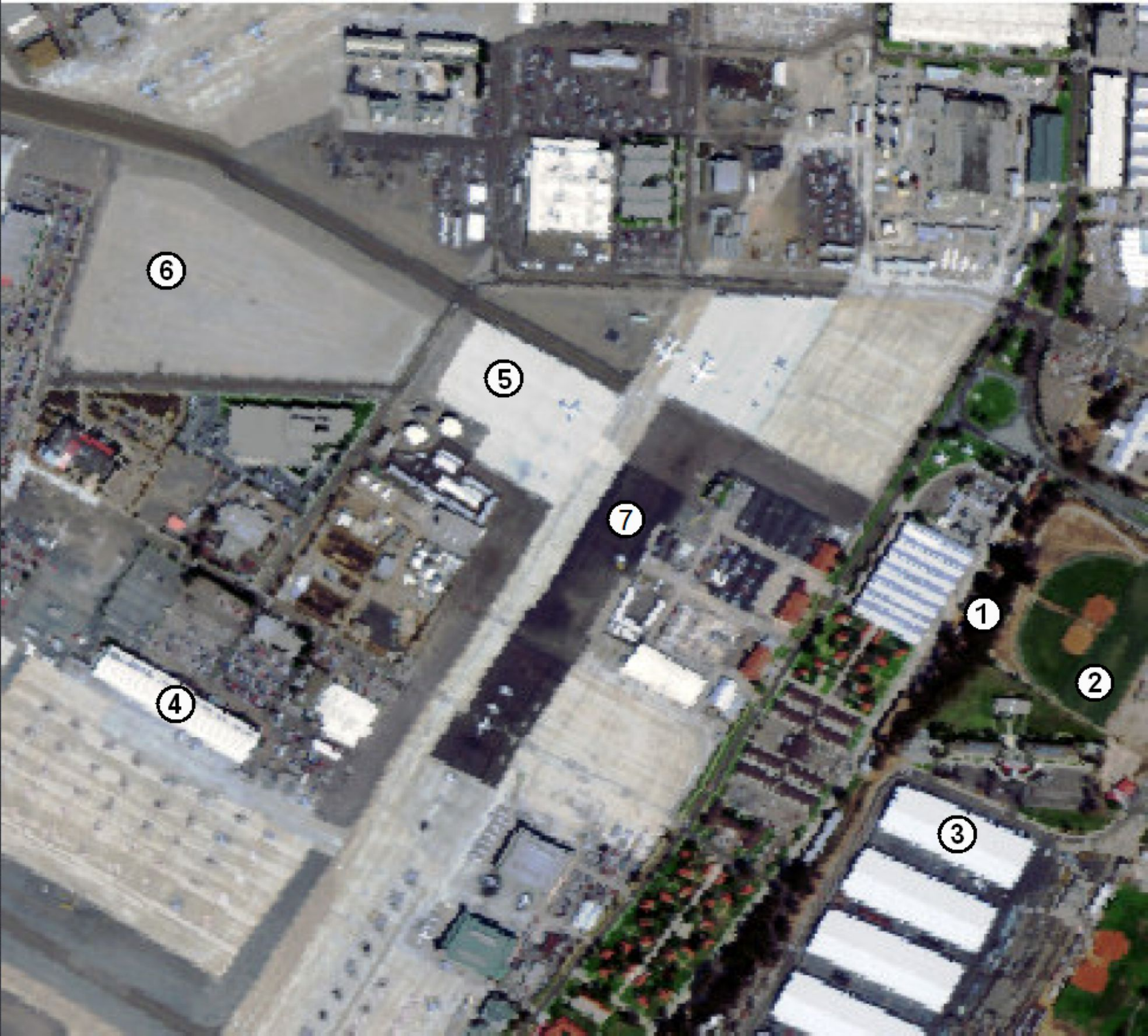}
  \end{tabular}
  \begin{tabular}{ccc}
    K-means & NMF & H2NMF \\ 
    \includegraphics[width=1.3in]{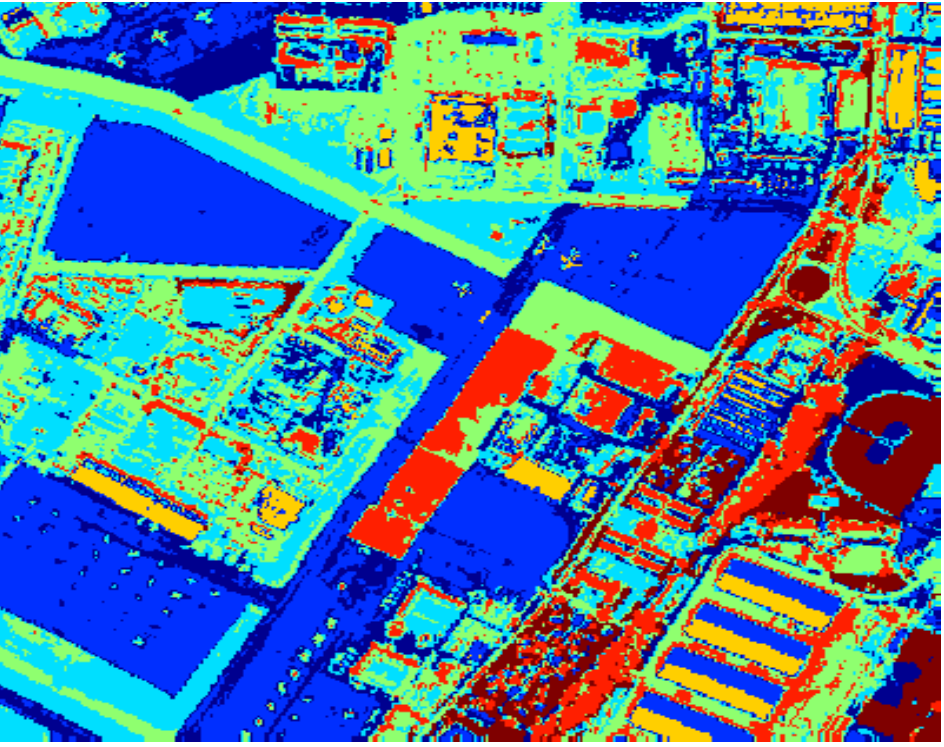}&
    \includegraphics[width=1.3in]{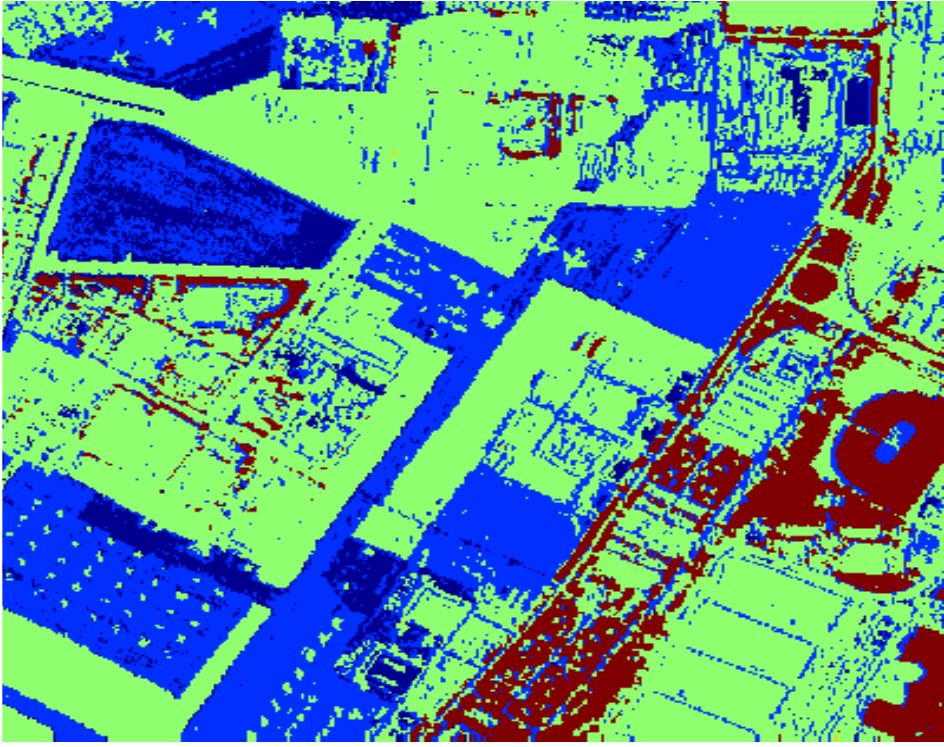}&
    \includegraphics[width=1.3in]{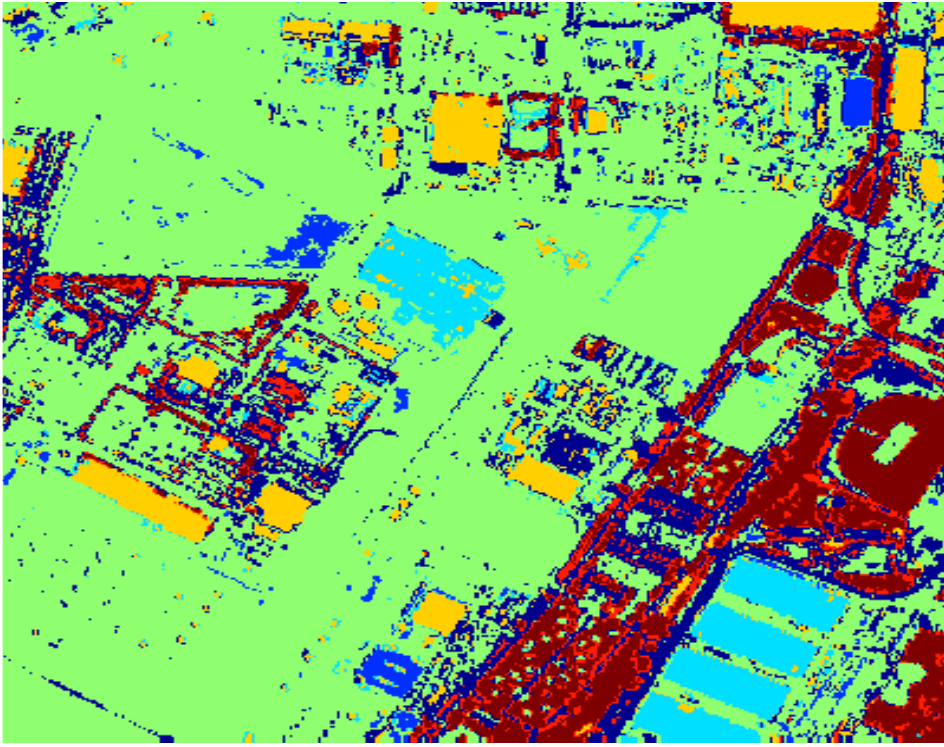}\\
    MBO & NLTV2 & NLTV1(H2NMF) \\ 
    \includegraphics[width=1.3in]{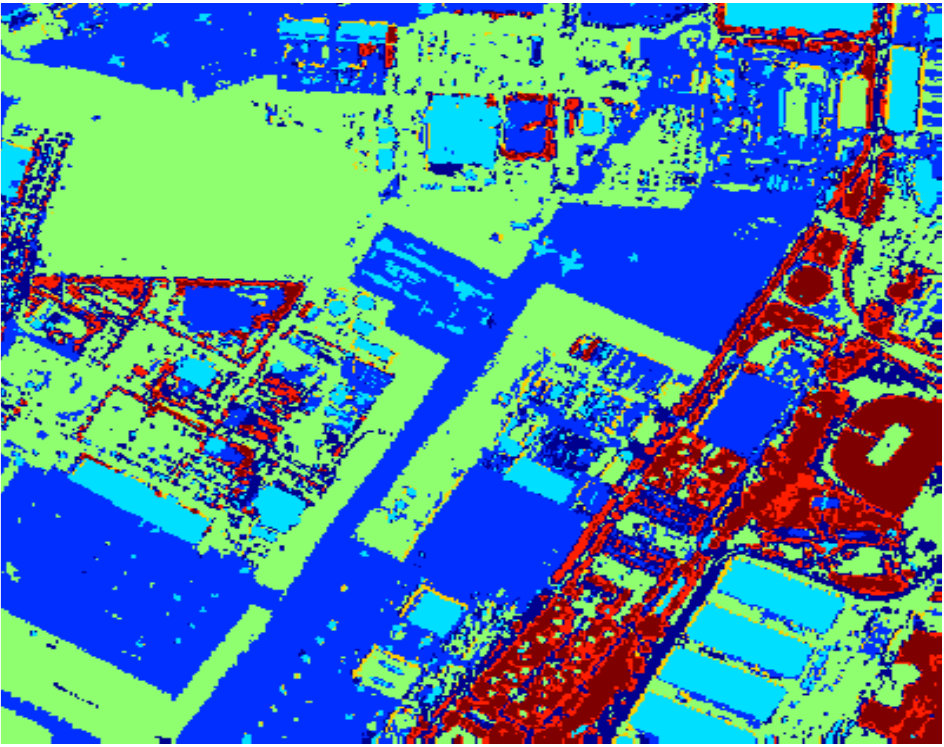}&
    \includegraphics[width=1.3in]{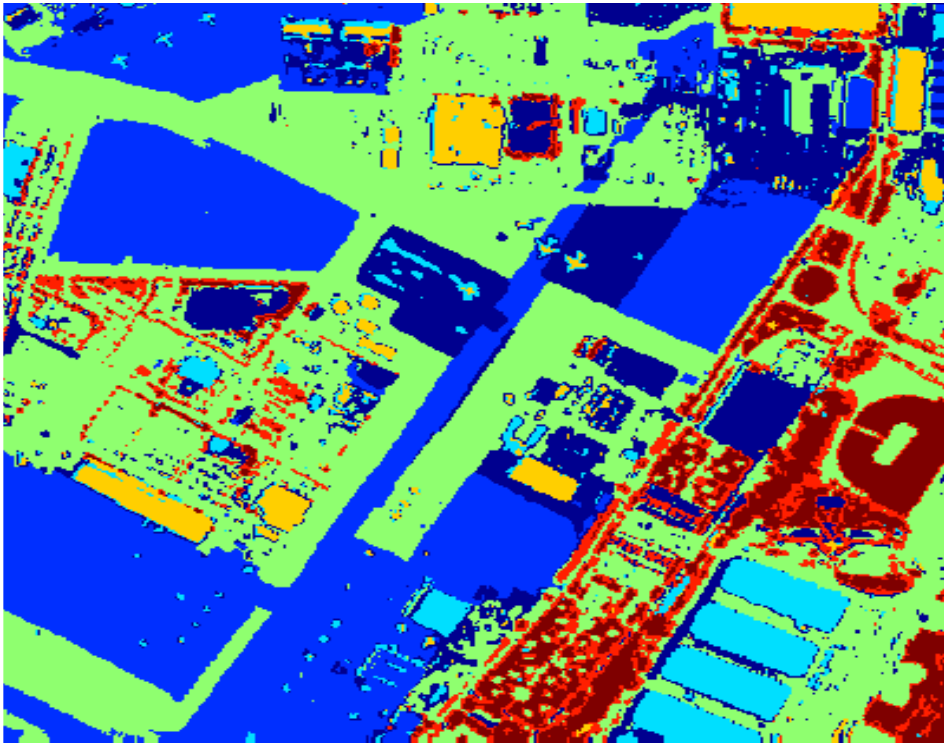}&
    \includegraphics[width=1.3in]{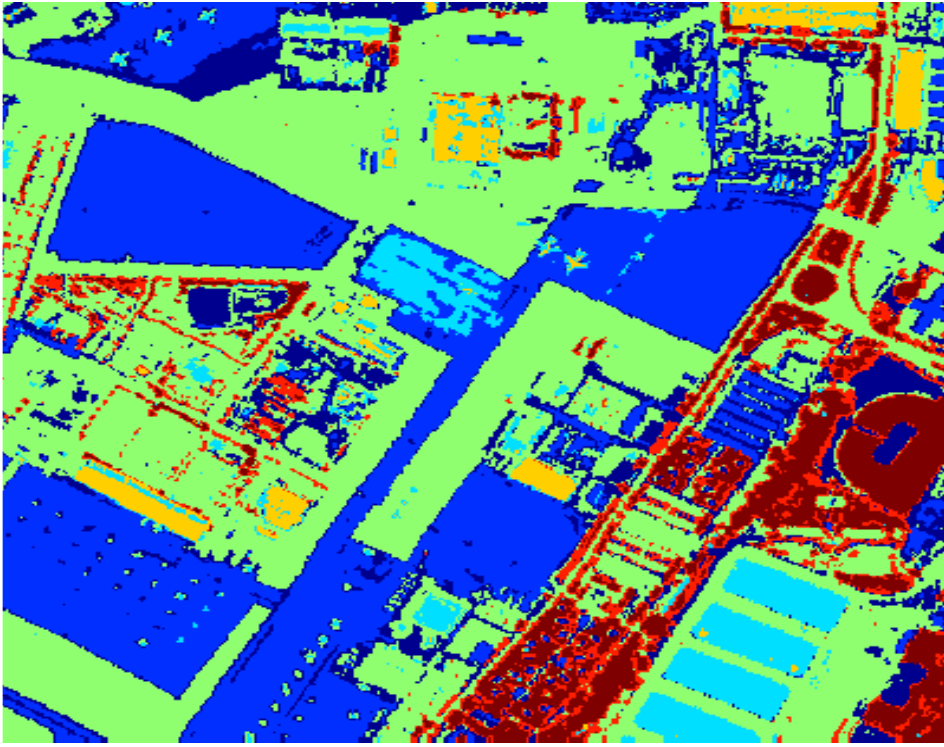}
  \end{tabular}
  \caption{Clustering results for the San Diego Airport dataset. The first image on the left is the RGB image, and the remaining six images are the clustering results of the corresponding algorithms.}
  \label{SD_comparepic}
\end{figure*}

The classification results and computational run-times are shown in Fig. \ref{SD_comparepic} and Table \ref{runtime}. No ground truth classification is available for this HSI, but after examining the spectral signatures of various pixels in the scene, we managed to pinpoint some errors that were common for each algorithm. We will not go into detail about the NMF and H2NMF algorithms, which clearly do not perform well on this dataset. K-means obtained some decent results, but splitted the rooftops of the four buildings on the bottom right of the image into two distinct clusters, and failed to separate two different road types (cluster 5 and 6). The MBO scheme failed on two accounts: it did not properly segment two different road surfaces (cluster 6 and 7), and did not account for the different rooftop types (cluster 3 and 4). The linear NLTV  model with H2NMF initialization is significantly more accurate than H2NMF and MBO. It successfully picked out two different types of roof (cluster 3 and 4), two different types of road (cluster 6 and 7), although the other type of road (cluster 5) is mixed with one type of roof (cluster 3). The best result was obtained by using the NLTV quadratic model with random initialization, with the only problem that tree and grass (clusters 1 and 2) are mixed together. However, the mixing of grass and tree is actually the case for all the other algorithms. This means that NLTV quadratic model alone was able to identify six of the seven clusters correctly. 

\subsection{Chemical Plume Dataset}
\begin{figure}[!t]
  \centering
  \begin{tabular}{ccc}
    K-means++ & NMF++ & H2NMF\\
    \includegraphics[width=1in]{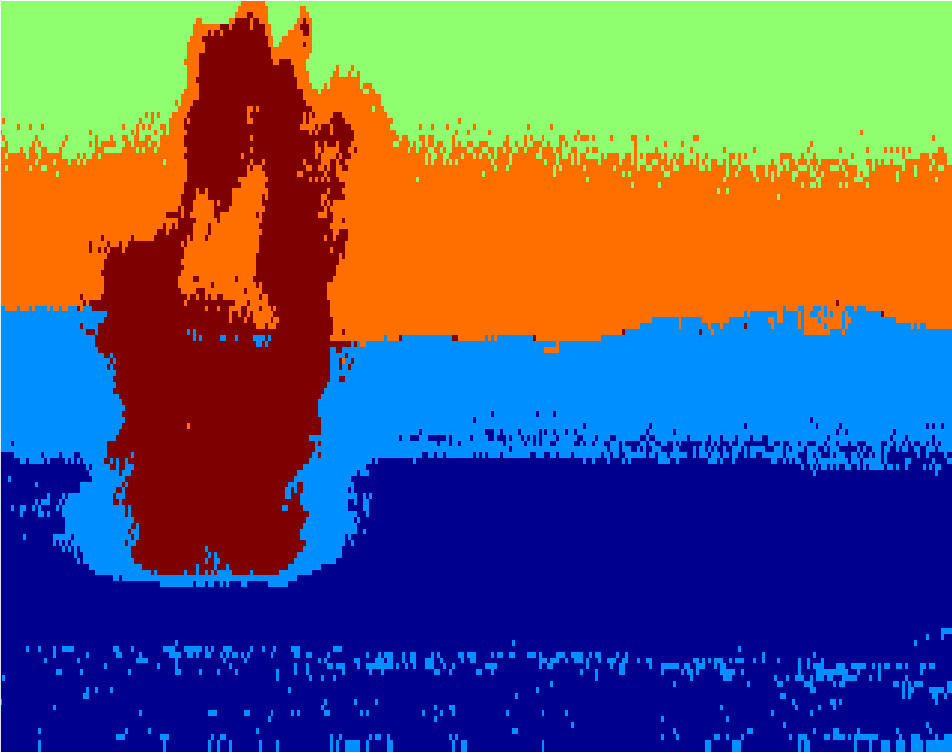}&
    \includegraphics[width=1in]{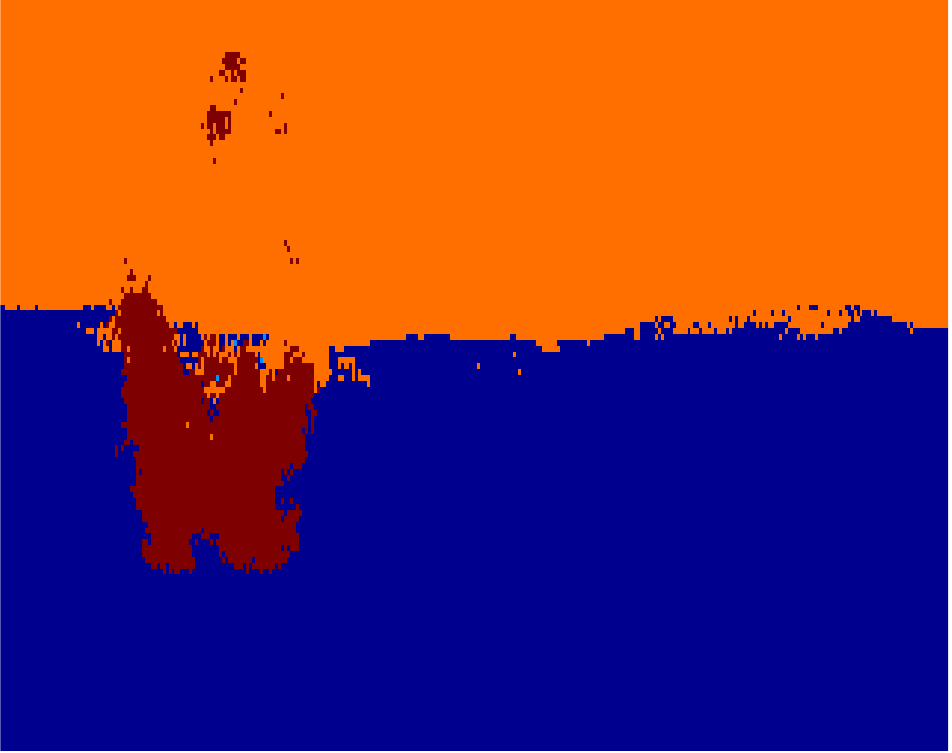}&
    \includegraphics[width=1in]{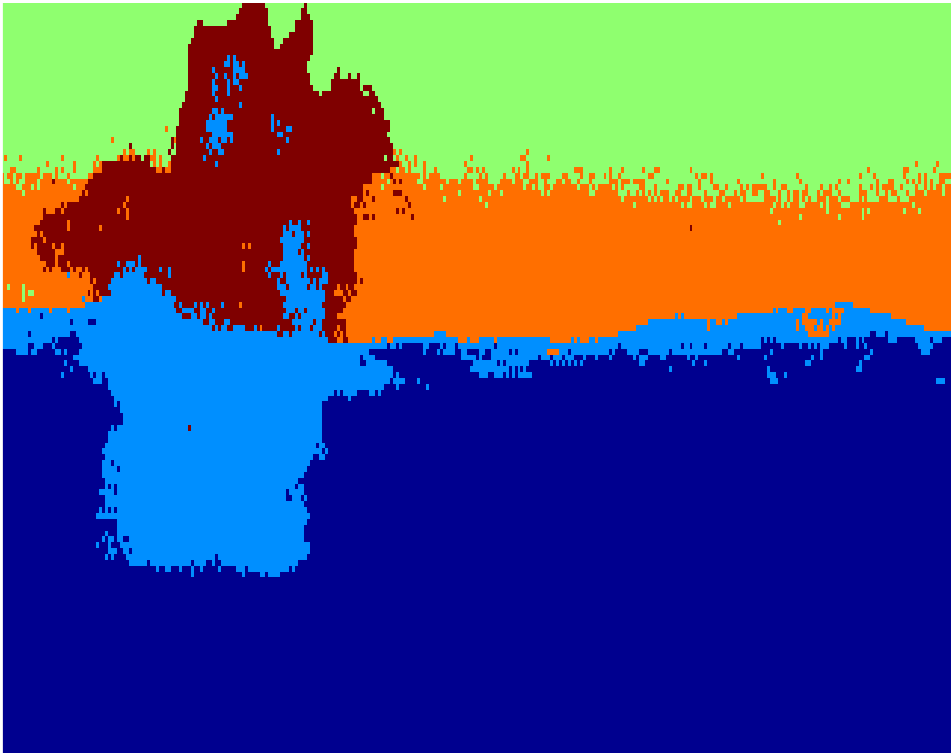}\\
    MBO++ & NLTV2++ & NLTV1(H2NMF)\\
    \includegraphics[width=1in]{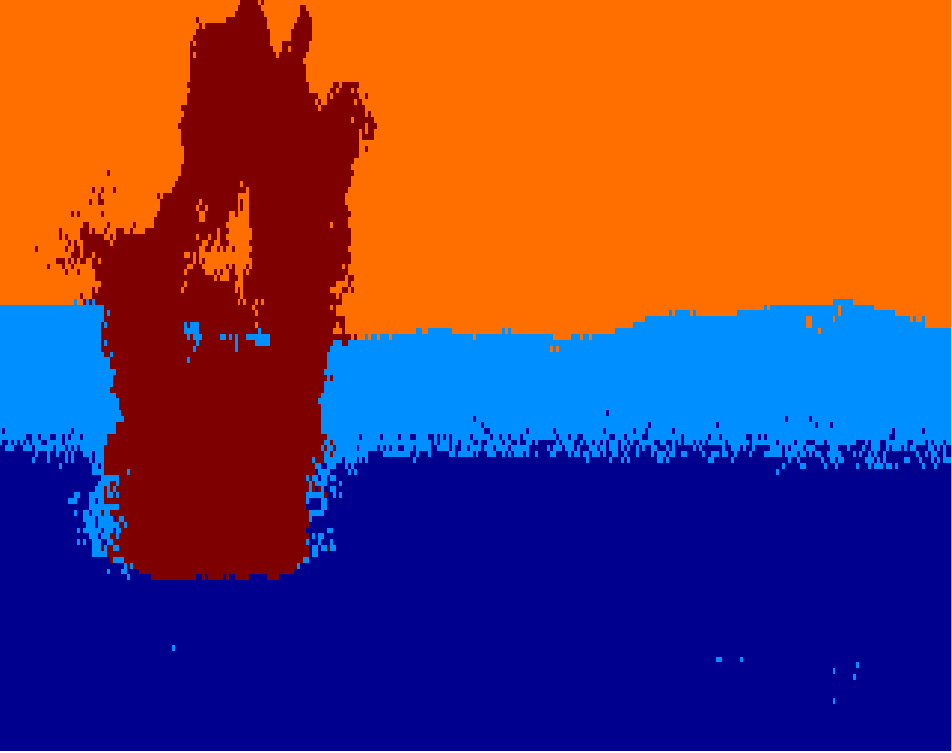}&
    \includegraphics[width=1in]{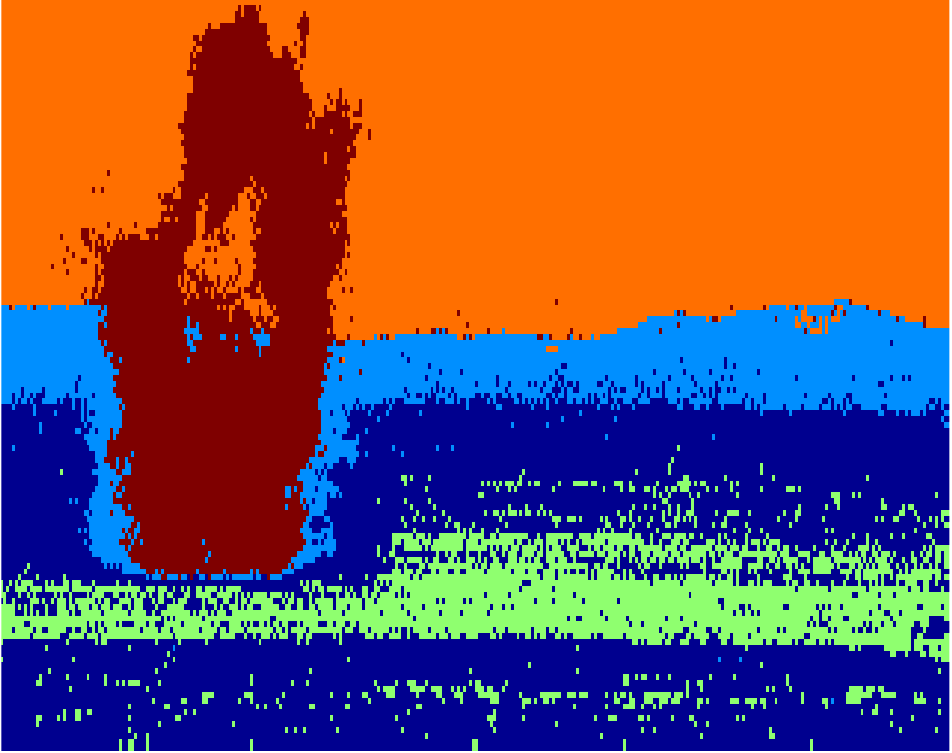}&
    \includegraphics[width=1in]{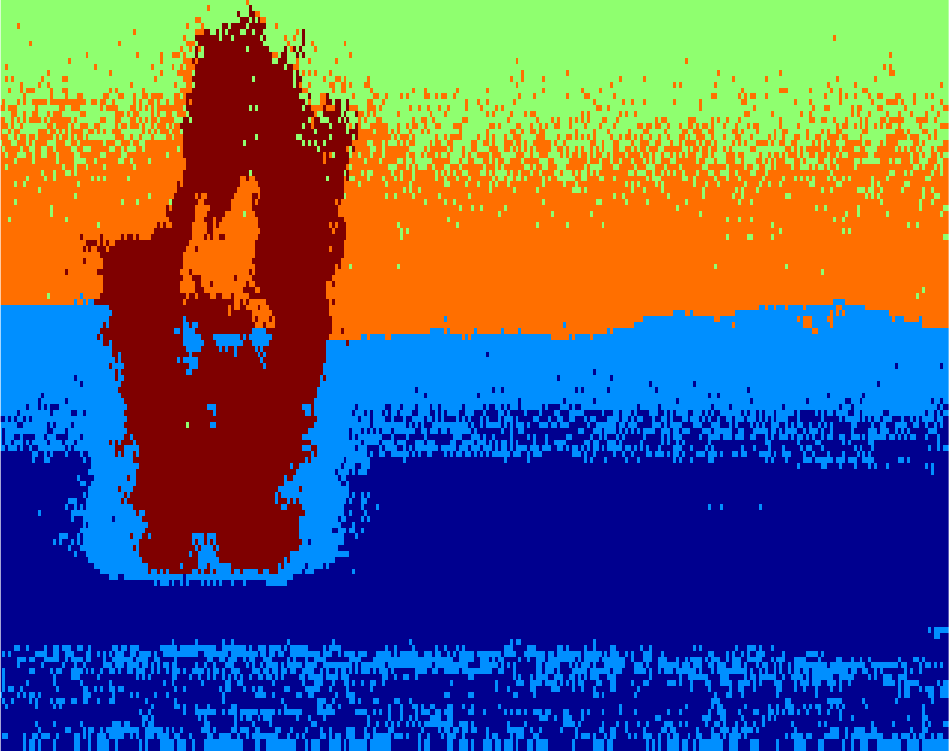}
  \end{tabular}
  \caption{Clustering results for the Chemical Plume dataset.}
  \label{plume_compare}
\end{figure}

Analyzing images for chemical plumes is  more difficult because of its diffusive nature. All the algorithms are run  on the image before it was denoised and the results are shown in Figure \ref{plume_compare}. The unmixing methods such as NMF and H2NMF do not perform satisfactorily on this dataset. MBO++,  K-means++, and  NLTV2++ can all properly identify the chemical plume. Note that NLTV with H2NMF as centroid initialization outperforms H2NMF as a classification method. We have to point out that the NLTV quadratic model is not so robust with respect to the centroid initialization even with a ``K-means++'' type procedure on this dataset. But this is also the case for all the other testing algorithms. The MBO scheme, which was specifically designed for this dataset \cite{huiyi_plume}, does seem to have the highest robustness among all the algorithms.

\subsection{Pavia University, Indian Pines, and Kennedy Space Center Dataset}

\begin{table}[!t]
\renewcommand{\arraystretch}{1.3}
\caption{Comparison of Overall Accuracies on the Pavia University, Indian Pines, and Kennedy Space Center Datasets}
\label{badtab}
\centering
\begin{tabular}{|c|C|C|C|}
\hline
\bfseries Algorithm & \bfseries Pavia & \bfseries Pines & \bfseries KSC\\
\hline
\bfseries K-means++ & 42.31\% & 38.99\% &41.73\% \\
\hline
\bfseries NMF++ & 54.97\%& 38.84\% &37.07\%\\
\hline
\bfseries H2NMF & 43.75\% & 36.78\% &37.07\%\\
\hline
\bfseries MBO++ & 50.04\% & 36.49\% &41.85\%\\
\hline
\bfseries NLTV, H2NMF init & 42.83\% &36.22\% &41.41\%\\
\hline
\bfseries NLTV++ & 44.01\% & 42.35\% &41.48\%\\
\hline
\end{tabular}
\end{table}

The Pavia University (9 clusters), Indian Pines (16 clusters), and Kennedy Space Center (12 clusters) datasets are frequently used to test supervised classification algorithms. To save space, we only report the numerical overall accuracies in Table \ref{badtab}. As can be seen, all the competing unsupervised algorithms performed poorly on these three datasets. Different clusters were merged and same clusters were splitted in various fashions by all the algorithms, which rendered the numerical accuracies no longer reliable.

The computational run-times of these three datasets are listed in Table \ref{runtime}. Unfortunately, when the number of clusters is increasing, the computational complexity of the quadratic model grows exponentially. The reason is that the number of grid points ($\delta$ in Fig. \ref{pushing}) on the unit simplex grows exponentially as the dimension of the simplex increases. Therefore, when the number of clusters is large enough (greater than 10), the stable simplex clustering will become the most time-consuming part of the quadratic model. On these three datasets, we sacrificed the accuracy of the quadratic model by creating a coarser mesh on the unit simplex.

The reason why  NLTV, as well as all the other competing unsupervised algorithms, performed poorly on these three datasets is two-fold. First, when the number of classes is too large in a HSI covering a large geographic location, the variation of spectral signatures within the same class cannot be neglected when compared to the difference between the constitutive materials, especially when the endmembers themselves are similar. As a result, the unsupervised algorithms tend to split a ground-truth cluster with large variation in spectral signatures and merge clusters with similar centroids or endmembers. Second, there might exist more distinct materials in the image than reported in the ground truth. Therefore the algorithms might detect those unreported materials because no labeling has been used in these unsupervised algorithms. Thus we can conclude that NLTV, as well as other unsupervised methods reported in this paper, is not suitable for such  images at current stage. Modifying the NLTV algorithm to work for such datasets would be the direction of future work.

\subsection{Sensitivity Analysis over Key Model Parameters}
\label{sec:sensitivity}

\begin{figure}[!t]
\centering
\begin{tabular}{c}
  \includegraphics[width=3.5in]{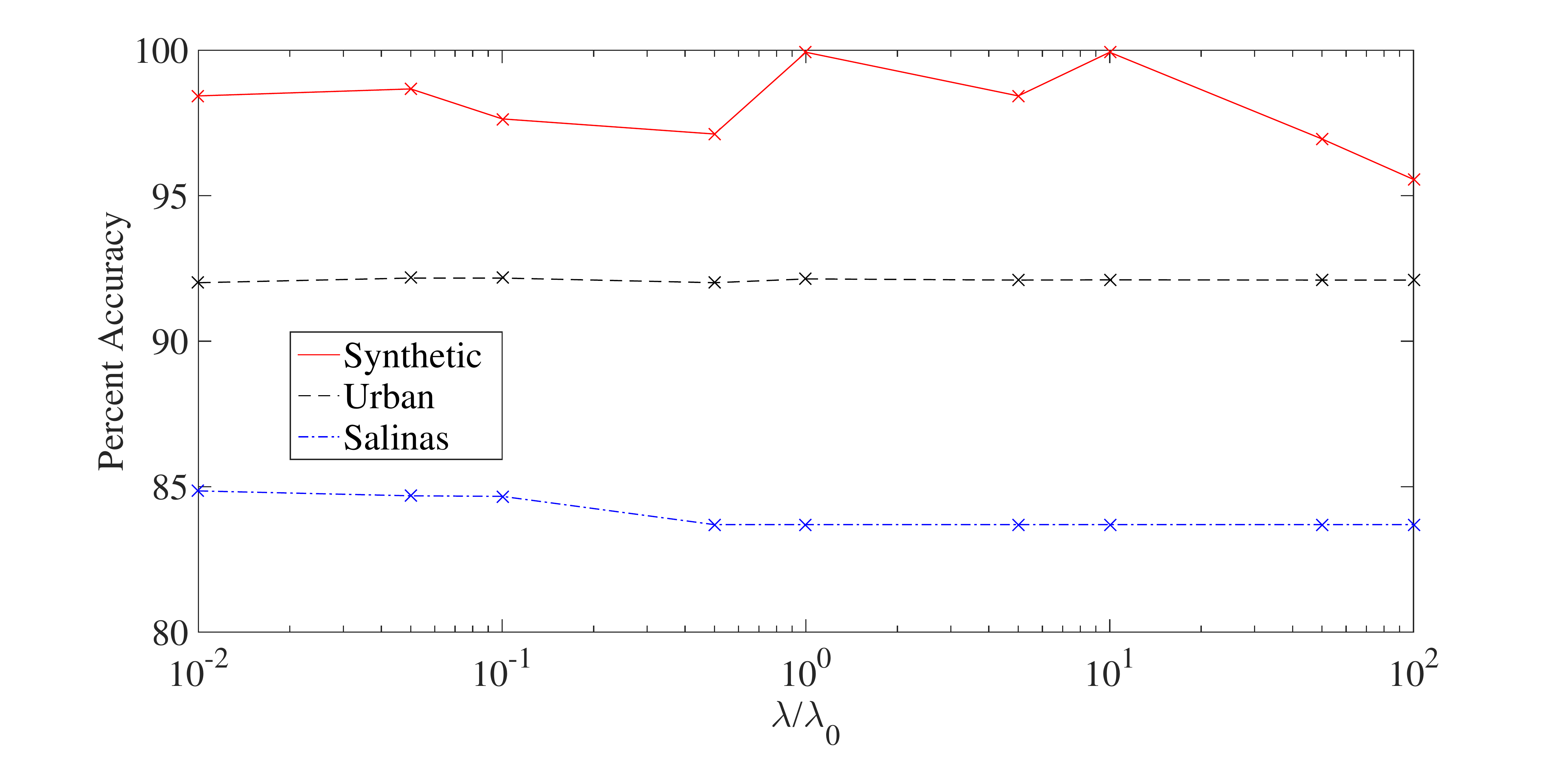}\\
  \includegraphics[width=3.5in]{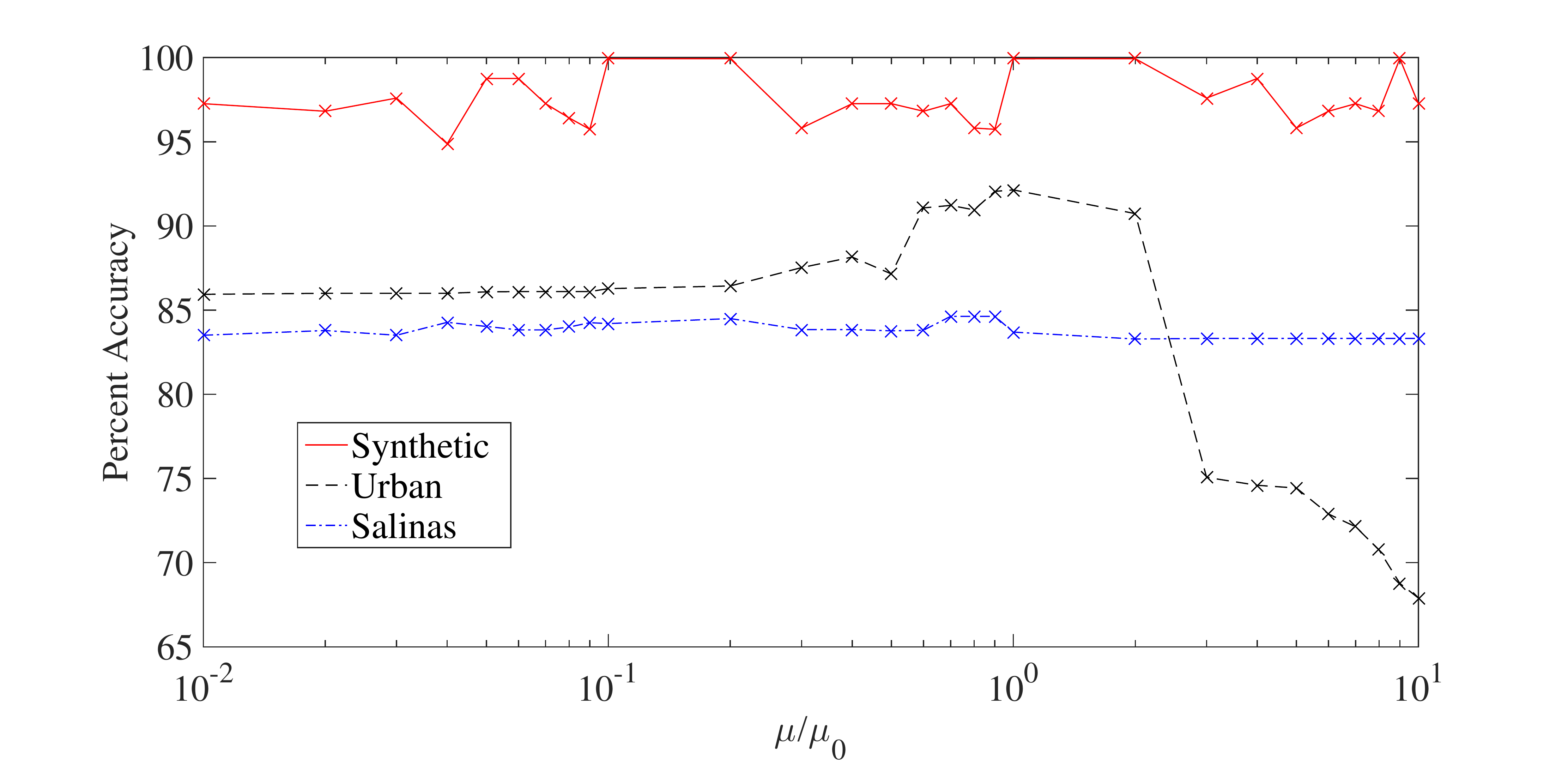}
\end{tabular}
\caption{This figure shows the robustness of the NLTV algorithm with respect to $\lambda$ and $\mu$. Centroid initialization remains identical as  $\lambda$ and $\mu$ are changing. $\lambda_0$ and $\mu_0$ are the optimal values specified in Section \ref{sec:setup}. The overall accuracies of the Synthetic, Urban, and Salinas-A datasets are displayed. }
\label{fig:sensitivity}
\end{figure}

\begin{figure}[!t]
\centering
\begin{tabular}{cccc}
  \hspace{-.3cm}$\mu=10^{-1}$&\hspace{-.5cm}$\mu=10^{-2}$&\hspace{-.5cm}$\mu=10^{-3}$&\hspace{-.5cm}$\mu=10^{-4}$\\
  \hspace{-.3cm}\includegraphics[width=.85in]{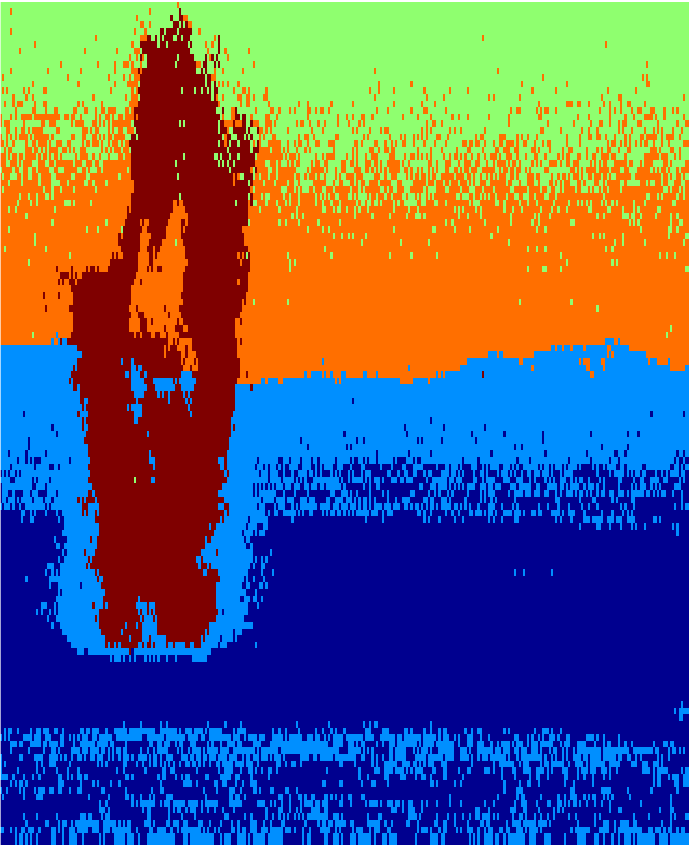}&\hspace{-.5cm}
  \includegraphics[width=.85in]{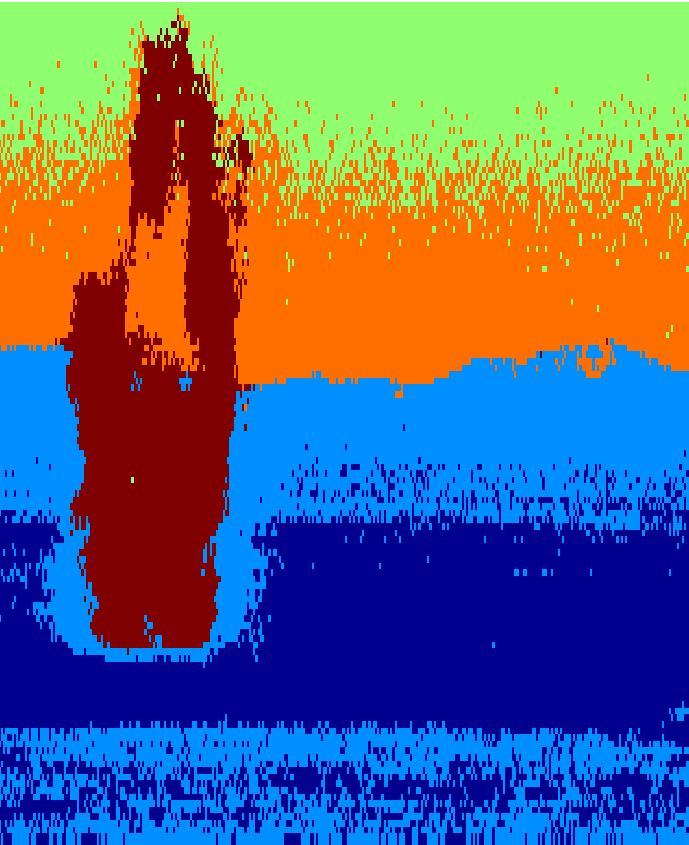}&\hspace{-.5cm}
  \includegraphics[width=.85in]{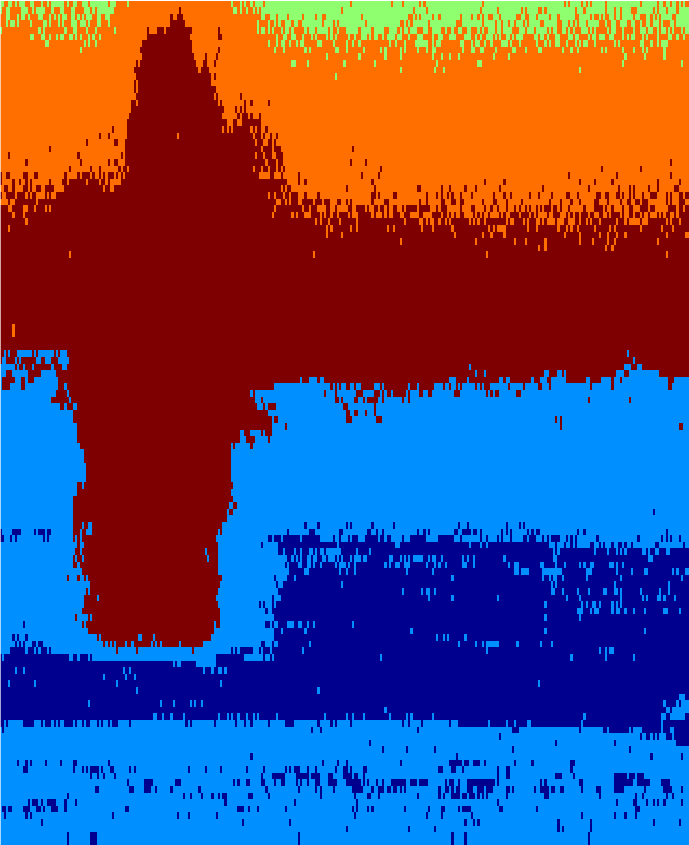}&\hspace{-.5cm}
  \includegraphics[width=.85in]{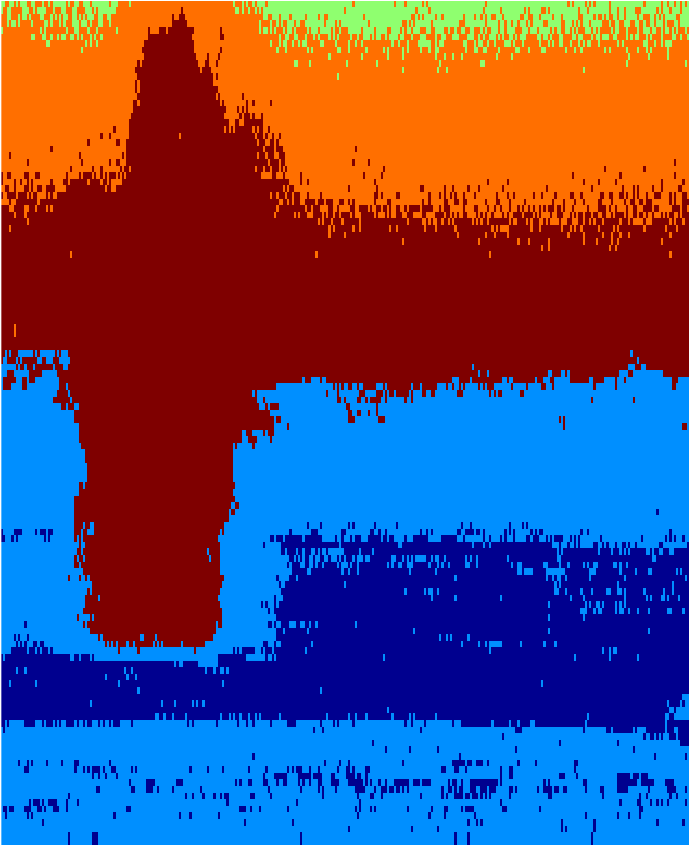}\\\hspace{-.5cm}
\end{tabular}
\caption{The sensitivity of the NLTV algorithm with respect to $\mu$ in the plume dataset. All the tests used the same centroid initialization (H2NMF).}
\label{fig:plume_sensitivity}
\end{figure}

At last,  a sensitivity analysis is provided over the parameters $\lambda$ and $\mu$ in the NLTV models. As mentioned in Section \ref{sec:setup}, $\lambda$ and $\mu$ are chosen to balance the scale of the regularizing and fidelity terms or the cosine and Euclidean distances. Fig. \ref{fig:sensitivity} displays the robustness of the NLTV algorithm on the Synthetic, Urban, and Salinas-A datasets with respect to $\lambda$ and $\mu$ within the variance of two magnitudes. Centroid initialization remains identical as $\lambda$ and $\mu$ are changing. It is clear that the NLTV algorithm is fairly robust with respect to $\lambda$ on all three datasets. The algorithm is also relatively robust with respect to $\mu$ on the Synthetic and Salinas-A datasets. As for the Urban dataset, a significant decay in accuracy can be observed as $\mu$ increases. This phenomenum is due to the fact that larger $\mu$ causes Euclidean distance to be the dominant one, which is not ideal with the presence of atmospheric interference in the Urban dataset. Smaller $\mu$ also leads to lower accuracy in the Urban dataset, which results from the similarity of ``road'' and ``dirt'' clusters measured in cosine distance. Overall, a reasonable robustness with respect to the key parameters $\lambda$ and $\mu$ can be concluded on these three tests.

Similar robustness can be observed on other datasets except for the Chemical Plume. Fig. \ref{fig:plume_sensitivity} shows the sensitivity of the result with respect to $\mu$. All the centroids are initialized using H2NMF, and vastly different results occurred as $\mu$ changes. This could be due to the presence of significant noise.

\section{Conclusion}
\label{sec:conclusion}
In this paper we present the framework for a nonlocal total variation method for unsupervised HSI classification, which is solved with the primal-dual hybrid gradient algorithm. A  linear and a quadratic version of this model are developed; the linear version updates more quickly and can refine results produced by a centroid extraction algorithm, and the quadratic model with stable simplex clustering method provides a robust means of classifying HSI with randomized pixel initialization.

The algorithm is tested on both a synthetic and seven real-world datasets, with promising results. The proposed NLTV algorithm consistently performed with highest accuracy on synthetic and urbanized datasets such as Urban, Salinas-A, and the San Diego Airport, both producing smoother results with easier visual identification of segmentation, and distinguishing classes of material that other algorithms failed to differentiate. The NLTV algorithm also performed well on anomaly detection scenarios like the Chemical Plume datasets; with proper initialization, it performed on par with the Merriman-Bence-Osher scheme developed specifically for this dataset. However, NLTV, as well as other unsupervised algorithms, failed to achieve satisfactory results on  datasets with a relatively large number of clusters. The run-times of the NLTV algorithms are generally comparable to the other methods, and the consistent higher accuracy on different types of datasets suggests that this technique is a more robust and precise means of classifying hyperspectral images with a moderate number of clusters.

\appendix{Proof of Theorem \ref{thm:1}:}
\label{app1}
Problem (\ref{eq:u}) is equivalent to:
\begin{equation}
  \label{eq:u_semiconstrained}
  \min_{\sum_{i=1}^ku_i=1}\delta_{\mathbb{R}_+^{k}(u)}+\frac{1}{2}\|Au-y\|^2_2,
\end{equation}
where $\mathbb{R}_+^{k}=\{ u \in \mathbb{R}^k: u_i \ge 0\}$ is the nonnegative quadrant of $\mathbb{R}^k$. The Lagrangian of (\ref{eq:u_semiconstrained}) is:
\begin{align*}
  \mathcal{L}(u,\lambda) = \sum_{i=1}^k \left(\frac{1}{2}\left|a_iu_i-y_i\right|^2+\delta_{\mathbb{R}_+}(u_i)+\lambda u_i\right)-\lambda.
\end{align*}
If $u^*$ is a soluton of (\ref{eq:u_semiconstrained}), KKT conditions \cite{vand} imply that there exists a $\lambda$ such that:
\begin{align*}
  u^* = \arg \min_u \mathcal{L}(u,\lambda) = \arg \min_{u_i \ge 0} \sum_{i=1}^k \frac{1}{2}a_i^2 \left(u_i+\frac{\lambda-a_iy_i}{a_i^2}\right)^2.
\end{align*}
Therefore $u_i^*= \max\left(\frac{a_iy_i-\lambda}{a_i^2},0\right)$. Meanwhile, the primal feasibility requires:
\begin{align*}
  \sum_{i=1}^ku^*_i = \sum_{i=1}^k \max\left(\frac{a_iy_i-\lambda}{a_i^2},0\right)=1.
\end{align*}
And this proves Theorem \ref{thm:1}.

\section*{Acknowledgment}
The authors would like to thank Zhaoyi Meng and Justin Sunu, for providing and helping with the MBO code.

\ifCLASSOPTIONcaptionsoff
  \newpage
\fi

\bibliographystyle{./IEEEtran}
\bibliography{./IEEEabrv,./HSI}

%




\end{document}